\documentclass[journal]{IEEEtran}

\usepackage{booktabs}       
\usepackage{amsfonts}       
\usepackage{nicefrac}       
\usepackage{microtype}      
\usepackage{color}
\usepackage{amsmath}
\usepackage{multirow}
\usepackage{diagbox}
\usepackage{soul}

\usepackage{cite}
\usepackage{amsmath,amssymb,amsfonts}
\usepackage{graphicx}
\usepackage{textcomp}
\usepackage{xcolor}
\usepackage{url}
\usepackage{enumerate}
\usepackage[font=footnotesize,labelfont=bf]{caption}

\usepackage{cite}
\usepackage{amsmath,amssymb,amsfonts}

\usepackage{graphicx}
\usepackage{booktabs}

\ifCLASSINFOpdf
\else
\fi
\begin{document}
%
\title{GAAF: Searching Activation Functions for Binary Neural Networks through Genetic Algorithm}


%
%
%
\author{\IEEEauthorblockN{Yanfei~Li$^{\dag}$, Tong~Geng$^{\ddag}$, Samuel~Stein$^{\ddag}$, Ang~Li$^{\ddag}$, and~Huimin~Yu$^{\dag}$}\\
\IEEEauthorblockA{$^\dag$Zhejiang University, Hangzhou, Zhejiang, China\\ $^\ddag$ Pacific Northwest National Laboratory, Richland, WA, USA  \\
aoxue18@zju.edu.cn, tong.geng@pnnl.gov, ang.li@pnnl.gov, yhm2005@zju.edu.cn}
}
\maketitle

\begin{abstract}

Binary neural networks (BNNs) show promising utilization in cost and power-restricted domains such as edge devices and mobile systems. This is due to its significantly less computation and storage demand, but at the cost of degraded performance. To close the accuracy gap, in this paper we propose to add a complementary activation function (AF) ahead of the sign based binarization, and rely on the genetic algorithm (GA) to automatically search for the ideal AFs. These AFs can help extract extra information from the input data in the forward pass, while allowing improved gradient approximation in the backward pass. Fifteen novel AFs are identified through our GA-based search, while most of them show improved performance (up to 2.54\% on ImageNet) when testing on different datasets and network models. Our method offers a novel approach for designing general and application-specific BNN architecture. Our code is available at http://github.com/flying-Yan/GAAF.

\end{abstract}


%
\IEEEpeerreviewmaketitle

\section{Introduction}
\label{s:introduction}
\noindent
Binary Neural Networks (BNNs) \cite{courbariaux2016binarized, hubara2016binarized} binarize the full precision inputs and weights of deep neural networks into binary values: $\{+1,-1\}$. Consequently, the costly floating-point dot-product (or multiply accumulation) computation can be replaced by lighgtweight bit-wise exclusive-nor (\texttt{xnor}) and population-count (\texttt{popc}) operations. BNNs show great potential in cost- \& power-restricted domains \cite{fasfous2021binarycop, chen2020gpu, huang2021fpga, ma2019efficient, ma2018binary}, such as Internet-of-Things (IoT), embedded systems, mobile devices, etc. This is primarily due to the significantly reduced computation requirements, lower power cost, lower storage demand, and improved robustness to input noise \cite{courbariaux2016binarized, li2019bstc, galloway2017attacking}. Existing research has shown that, through BNN, significant speedups and memory reduction can be harvested on CPUs \cite{hu2018bitflow}, GPUs \cite{li2019bstc, li2020accelerating}, and FPGAs \cite{geng2019o3bnn, geng2019lp, geng2020o3bnn}, compared with their DNN counterparts. 
As a result, BNNs have been adopted in a variety of practical applications such as COVID-19 face-cover detection \cite{fasfous2021binarycop}, auto-driving \cite{chen2020gpu}, smart agriculture \cite{huang2021fpga}, image enhancement \cite{ma2019efficient}, 3D objection detection \cite{ma2018binary}, etc. 

Though BNNs significantly alleviate the computation cost, they suffer from non-trivial performance degradation largely due to the imperfection of the binarization function and its derivative approximation \cite{anderson2017high, bethge2021meliusnet}. Most existing BNN works binarize the inputs and weights into binary (+1/-1) tensors through a deterministic \textbf{sign} function. This sign function also serves as the non-linear activation function of the neural networks. However, the gradient of the sign function is always zero and is non-differentiable at zero.
To allow standard back-propagation still able to work, the \emph{Straight Through Estimator} (STE) \cite{bengio2013estimating} is adopted as an approximated gradient of the sign function  \cite{courbariaux2016binarized, hubara2016binarized}. STE approximates the gradient as an identity function. To avoid the saturation effect where certain neurons can never get a chance to toggle its binary value, a clip function is placed prior to the sign function. It zeros out the gradient in the backward pass when real-value is outside the range (usually [-1,1]) but making no impact in the forward pass. 

The accuracy degradation with the sign activation in training arises from the information loss due to the rigid binarization through sign in the forward pass \cite{bethge2021meliusnet}, and the approximation imposed by the STE gradient. Consequently, significant efforts have been devoted in designing better activation functions for BNNs \cite{liu2018bi, darabi2018bnn+, liu2019circulant, qin2020forward}. Bi-Real Net \cite{liu2018bi} presented a customized BNN sign function -- \emph{ApproxSign}, which is a differentiable piecewise polynomial function that has a triangle-shaped derivative, acting as a magnitude-aware gradient. \emph{BNN+} \cite{darabi2018bnn+} showed a switch-like activation function for a better approximation of the sign derivative function. \emph{Self-Binarizing network} \cite{lahoud2019self} even adopted an evolving activation function to progressively reshape the scaled hyperbolic tangent function towards the sign function. Finally, \emph{ReActNet} \cite{liu2020reactnet} generalized the sign and PRelu functions as \emph{RSign} and \emph{RPReLU}, allowing shifting and reshaping the distribution of the real-valued feature maps prior to binarization, mitigating the information loss from to the rigid sign activation. 

Despite harvesting improved performance, most of these designs are crafted based on human expertise \cite{liu2018bi, darabi2018bnn+, liu2019circulant, lahoud2019self, qin2020forward, liu2020reactnet}. Whether they have already been optimal or if better activation functions can be found still remain an open question.

Recently, using machine-learning based search techniques for automatically discovering  traditionally human-designed components of deep neural networks exhibits great effectiveness. For example, enhanced accuracy has been showcased with the new activation functions discovered by reinforcement learning \cite{ramachandran2017searching}, with respect to the widely used ReLU activation in DNNs. In \emph{EvoNorms}, a set of new normalization-activation layers have been found by evolution-based search \cite{liu2020evolving}. \emph{DARTS} \cite{liu2018darts} modeled network architecture search (NAS) in a differentiable manner, thus can be efficiently handled by gradient descent.

Motivated by these search approaches, in this paper, we propose to leverage the \emph{genetic algorithm} (GA) based search algorithms for discovering novel activation functions for BNNs, and perform associated design space exploration. Observing that the complementary activation function is one-to-one mapping (i.e., transforming a tensor under the same shape), we define the structure of the first-order (Type-I) and second-order (Type-II) activation functions, acting as the templates describing how unary and binary functional operators can be composed to form an activation function. Each of the operators is selected from a list of labeled candidates, and the goal is to find the optimal combination. We encode the choice of each combination as a string of fixed length. We then rely on GA to auto-search the optimal combination. The newly discovered activation functions are evaluated on CIFAR10 and ImageNet with several BNN models, demonstrating improved validation accuracy (up to $2.54\%$ on ImageNet). In total, seven lightweight activation functions are identified showing low computation cost and few parameters. Similar to the clip function, these complementary activation functions only show-up in the training, and do not impact BNN inference efficiency. This paper thus makes the following contributions:

\begin{itemize}
    \item Our work is the first effort to automatically search complementary activation functions for BNNs, to the best of our knowledge.
    \item Through GA-based effective searching, we have identified seven activation functions that can show enhanced accuracy on different datasets and network models, with up to 2.54\% performance improvement on ImageNet.    
\end{itemize}

This paper is organized as follows. In Section~\ref{s:Background and Motivation}, we discuss the background of BNNs and GA-based search, highlighting our motivation. In Section~\ref{s:Methodology}, we present the proposed searching methodology. In Section~\ref{s:EXperiments}, we show the experiment and evaluation results. We discuss our result in Section~\ref{s:Discussion}, summarize the related work in Section~\ref{s:Related Work}, and conclude in Section~\ref{s:Conclusion}.

\section{Background and Motivation}
\label{s:Background and Motivation}
\subsection{Binary Neural Networks}
\noindent
BNNs \cite{courbariaux2016binarized, hubara2016binarized} are seen as an alternative approach of neural network compression, in addition to network pruning \cite{han2015deep, blalock2020state}. Instead of reducing the total number of weights in the network, BNN reduces the bit-width of the inputs and weights to the extreme of a single bit only. That is to say, both weights and inputs are expressed as 1-bit with two possible values: +1 or -1. Compared to traditional DNNs, the 32-bit floating-point multiplication and addition can be replaced by efficient bit operations (i.e., \emph{exclusive-nor} and \emph{population-count}), which is extremely hardware-friendly \cite{geng2020o3bnn, li2020accelerating}, given the one-to-one mapping from a neuron to a logic-bit.

\begin{figure}[ht]
\centering
\includegraphics[width=0.9\columnwidth]{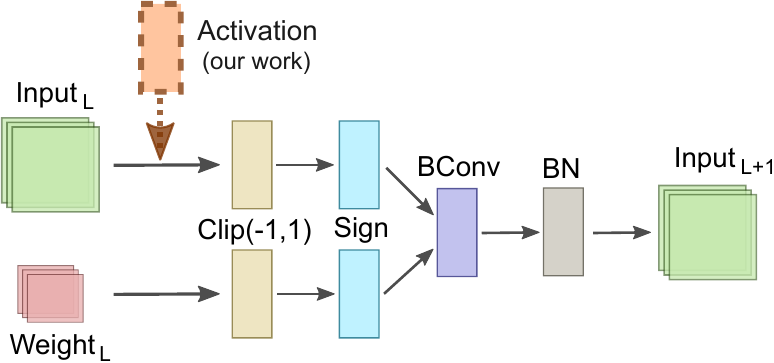} 
\caption{Structure of a typical binary convoluation layer.}
\label{fig:BNN}
\end{figure}

A typical binary convolution layer in BNNs is depicted in Fig.~\ref{fig:BNN}, First, the real-valued $input_L$ and weight $w_L$ of layer $L$ are binarized. Then, a binarized convolution (BConv) is applied on the binary inputs and weights. The outputs of BConv are real-valued, and serve as the inputs for the consecutive layer $L+1$ (i.e., $input_{L+1}$) after batch-normalization (BN) \cite{ioffe2015batch}.

\begin{figure}[!htb]
\minipage{0.49\columnwidth}
\includegraphics[width=1\columnwidth]{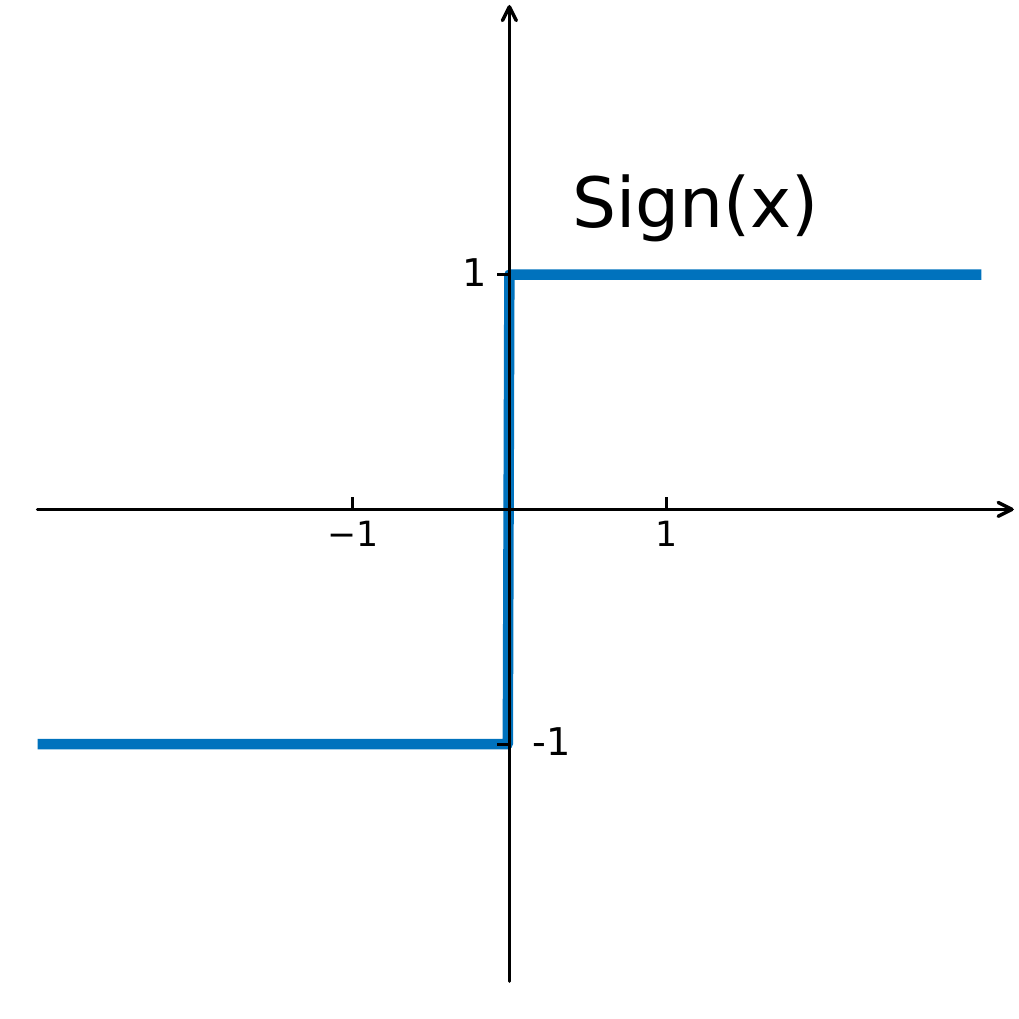} 
\caption{\emph{sign} function.}
\label{fig:sign}
\endminipage\hfill
\minipage{0.49\columnwidth}
\includegraphics[width=1\columnwidth]{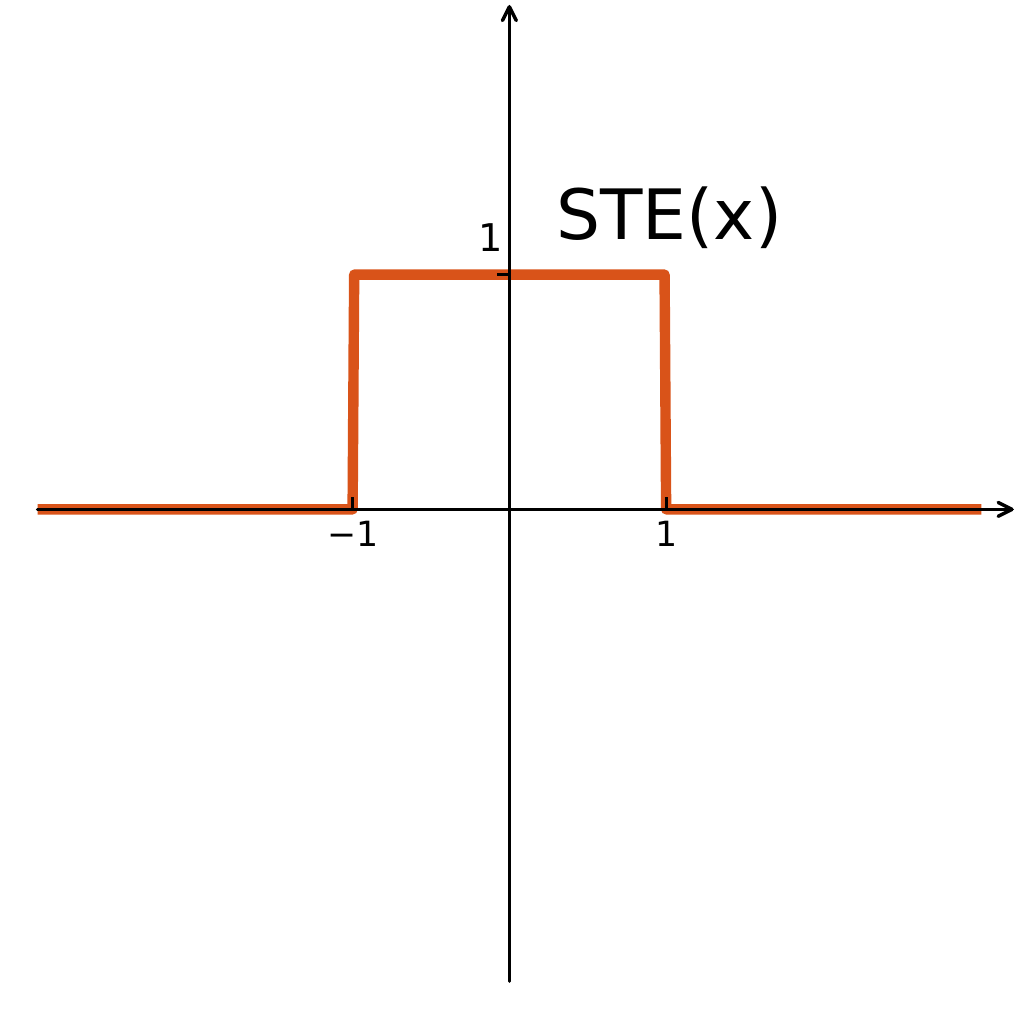} 
\caption{STE gradient of \emph{sign}.}
\label{fig:ste}
\endminipage\hfill
\end{figure}

BNN uses the sign function in Eq.~(\ref{eq:bin}) for binarization, which also serves as the non-linear activation function. Its plot is shown in Fig.~\ref{fig:sign}.

\begin{equation}
  \text{Forward:} \quad 
  x^b = sign(x) = \begin{cases}
    +1 & x\geq 0 \\
    -1 &\text{otherwise}  \\
  \end{cases}       \\
  \label{eq:bin}
\end{equation}
\vskip 2mm

Since the gradient of sign function is always zero and is non-differentiable at 0, the \emph{Straight-Through-Estimator} (STE) \cite{bengio2013estimating} is used for approximating the gradient of the sign function, as shown in Eq.~(\ref{eq:dbin}). 

\begin{equation}
\text{Backward:} \quad \frac{\partial Loss}{\partial x} = \frac{\partial Loss}{\partial x^b} \mathbf{1}_{|x|<t_{clip}}
\label{eq:dbin}
\end{equation}
\vskip 2mm

where $t_{clip}$ is typically 1. This implies that for values within the range of [-1,+1], the gradient of full-precision values ($\frac{\partial Loss}{\partial x}$) is equal to the gradient of the binarized values ($\frac{\partial Loss}{\partial x^b}$). For values outside the range, the gradient keeps zero. The plot is shown in Fig.~\ref{fig:ste}.  

Eq.~(\ref{eq:bin}) and (\ref{eq:dbin}) are achieved in BNN through the combination of \emph{Clip(-1,1)} (also known as \emph{hard tanh}) and \emph{Sign} in Fig.~\ref{fig:BNN}. Being followed by \emph{Sign}, \emph{Clip(-1,1)} does not impact the forward pass. However, for backward pass, it constraints the range of [-1,1] over the gradient and cuts-off values beyond the range. 

As seen in Fig.~\ref{fig:sign}, the sign function is very rigid by fixing the binarization boundary at value 0. In \cite{liu2020reactnet}, Liu et al. observed that by involving small variations to distributions of feature maps, better performance can be achieved. This is because the output only has two options: +1/-1, a small shift or reshape to the distribution of the input feature maps could result in a completely distinct binary output \cite{liu2020reactnet}. 

\begin{figure}[!htb]
\minipage{0.45\columnwidth}
\includegraphics[width=1\columnwidth]{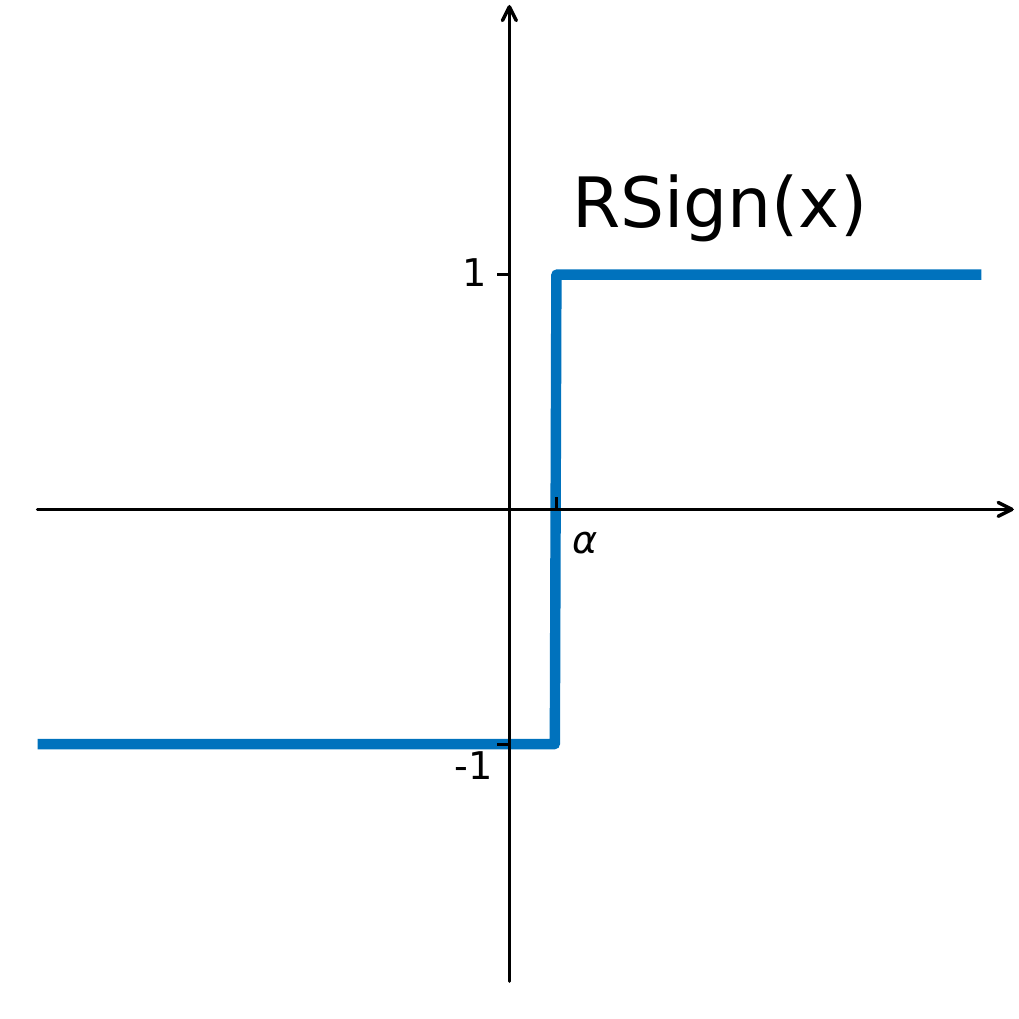} 
\caption{\emph{RSign} from \cite{liu2020reactnet}.}
\label{fig:rsign}
\endminipage\hfill
\minipage{0.45\columnwidth}
\includegraphics[width=1\columnwidth]{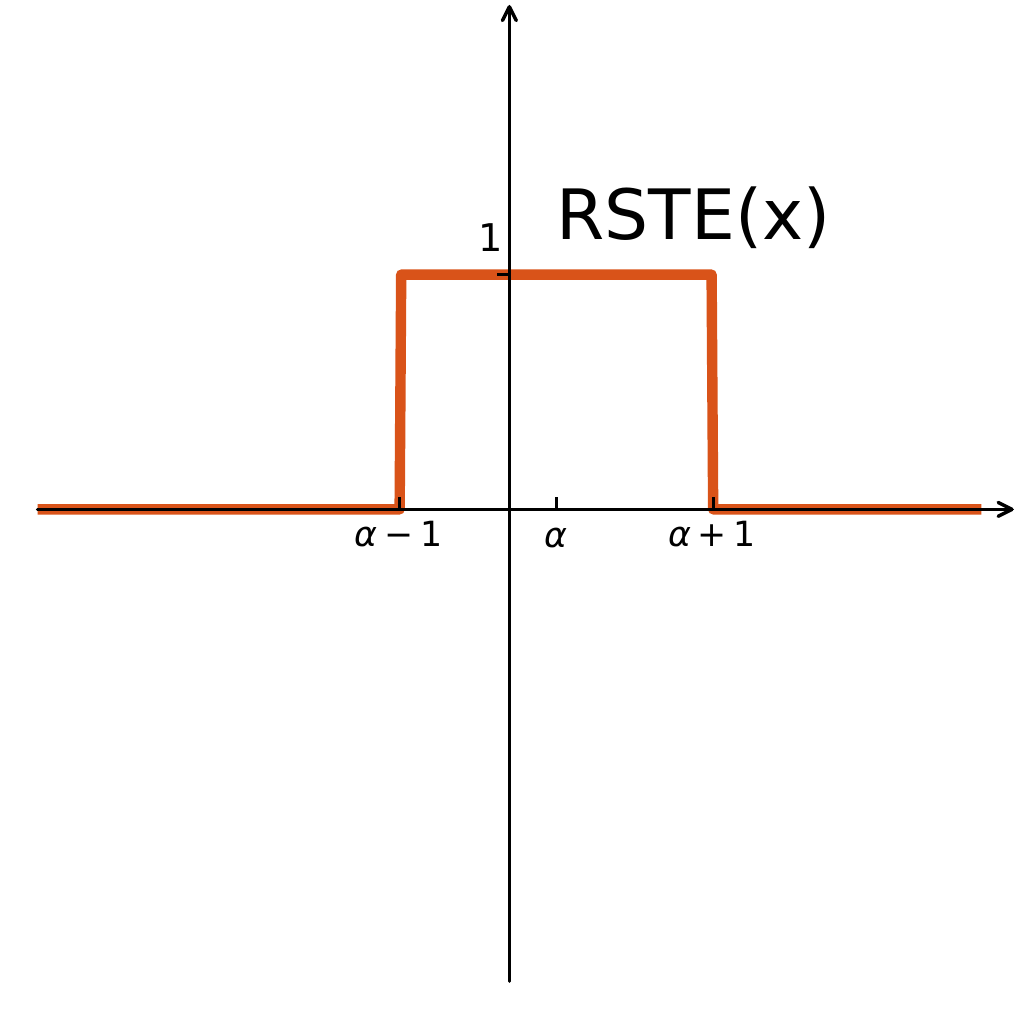} 
\caption{Gradient of \emph{RSign}.}
\label{fig:drsign}
\endminipage\hfill
\end{figure}

With such an observation, they proposed two improving approaches: (i) \textbf{\emph{RSign}}, where a channel-wisely learnable bias is imposed to the original sign function, see Eq.~(\ref{eq:rsign}). 

\begin{equation}
  x_i^b = RSign(x_i) = \begin{cases}
    +1 & x_i \ge \alpha_{i} \\
    -1 & x_i < \alpha_{i}  \\
  \end{cases}       \\
  \label{eq:rsign}
\end{equation}
\vskip 2mm

where $\alpha_{i}$ is a learnable channel-wise parameter. Fig.~\ref{fig:rsign} and \ref{fig:drsign} show the plot of the \emph{RSign} activation function and its gradient.

\begin{figure}[!htb]
\minipage{0.45\columnwidth}
\includegraphics[width=1\columnwidth]{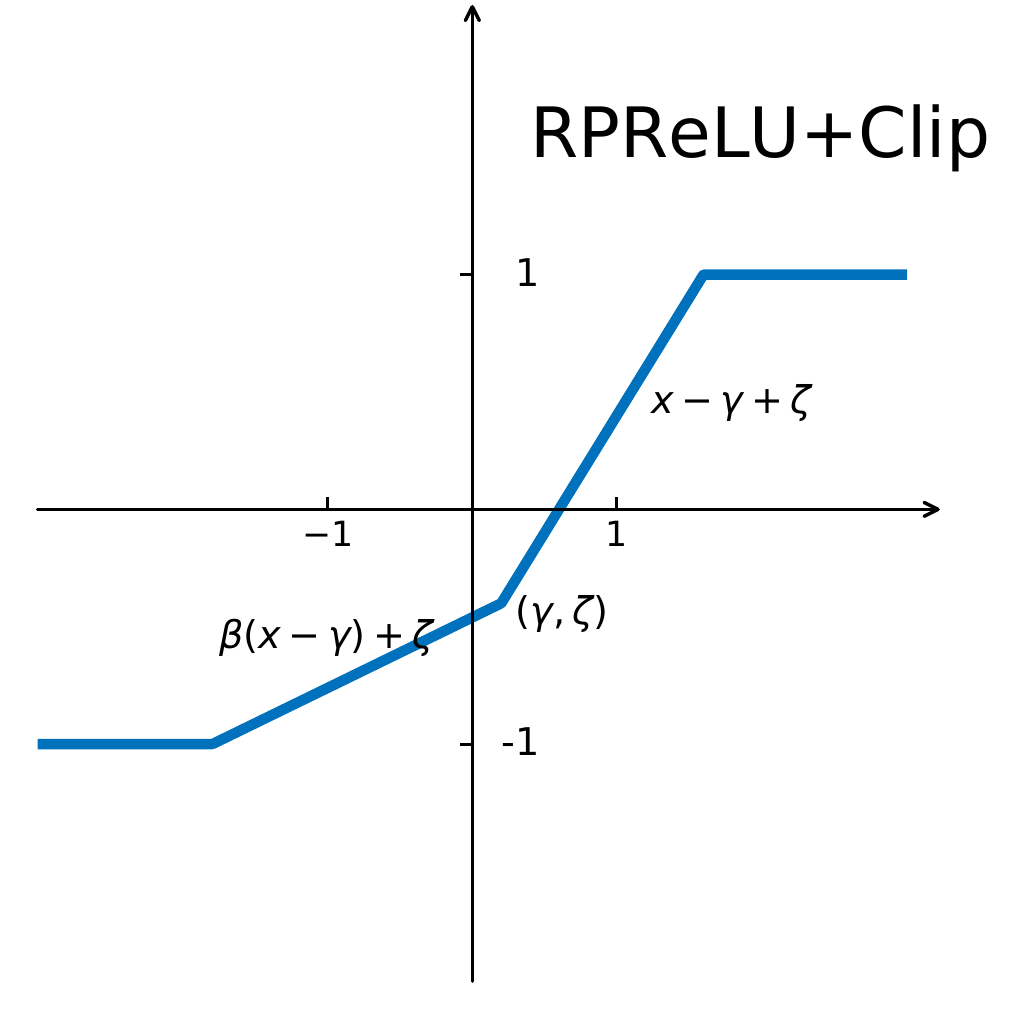} 
\caption{\emph{RPReLU}+\emph{Clip(-1,1)} in \cite{liu2020reactnet}.}
\label{fig:rprelu}
\endminipage\hfill
\minipage{0.45\columnwidth}
\includegraphics[width=1\columnwidth]{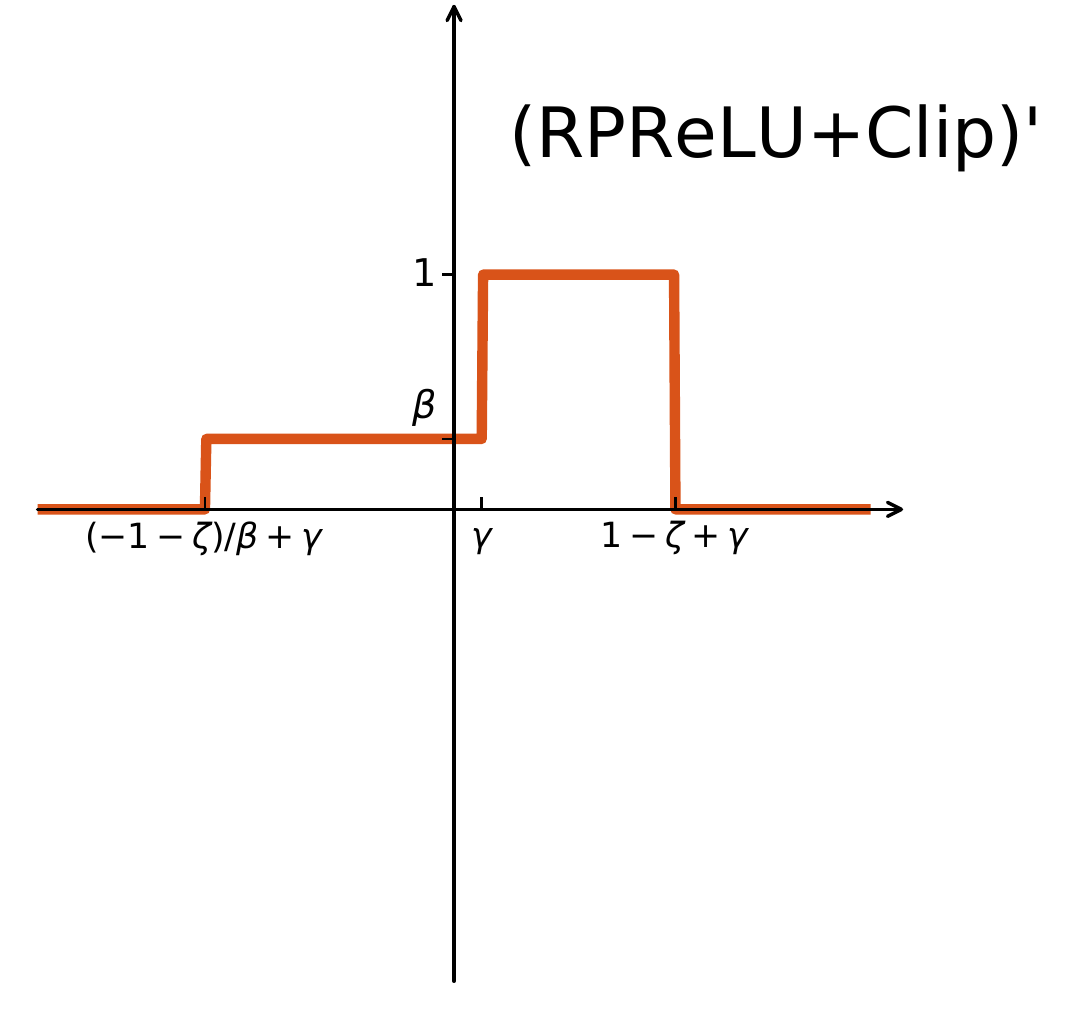} 
\caption{Gradient of \emph{RPReLU}+\emph{Clip}}
\label{fig:drprelu}
\endminipage\hfill
\end{figure}

(ii) \emph{RPReLU}, where a parameterized \emph{PReLU} is placed ahead of sign to be more flexible, see Eq.~(\ref{eq:rprelu}).

\begin{equation}
  f(x_i) = \begin{cases}
    x_i-\gamma_i + \zeta_i & x\ge \gamma_{i} \\
    \beta_i(x_i-\gamma_i) + \zeta_i & x< \gamma_{i}  \\
  \end{cases}       \\
  \label{eq:rprelu}
\end{equation}
\vskip 2mm

where $\gamma_i$ and $\zeta_i$ are learnable shifts, and $\beta_i$ is a learnable slope for the negative portion of the function. We plot this new activation function (\emph{RPReLU} + \emph{Clip}) and its gradient in Fig.~\ref{fig:rprelu} and \ref{fig:drprelu}.

\noindent\textbf{Motivation 1:} Motivated by the observation that a parameterized variation on the distribution of the input feature map can gain better performance, we propose to \emph{systematically adding a complementary function prior to sign.} Their combination serves as the new activation function, where the binary inputs can capture more information contained in the real-valued inputs. The challenge then is how to find a decent complementary function.  

\subsection{Genetic Algorithm}
\noindent
The \emph{Genetic Algorithm} (GA) is a widely-adopted optimization algorithm based on population metaheuristics. GA belongs to the larger class of evolutionary algorithm (EA). Proposed by John Holland and popularized by David Goldberg \cite{goldberg1988genetic}, GA shows great success in various optimization problems, especially when aiming at global search in a discrete space.   

GA simulates the process of natural selection, namely the Darwinian Theory of \emph{Survival of the Fittest}. Specifically, GA is realized as an iterative process, where a population of candidate solutions are evolved towards a better solution. Before the iteration starts, two definitions are required: (i) a genetic representation of the solution domain that determines the search space; (ii) a fitness function that estimates the degree of how fit the candidate solution is, usually being the objective function of the optimization problem.

GA typically starts from a population of randomly generated individuals, where the fitness of each individual is evaluated. Then, the parent individuals are stochastically selected from the current population, where the individuals with better fitness scores have a better chance to be selected for reproduction. Based on the selected parent individuals, an offspring is generated by operations such as crossover, mutation. The new offspring is then evaluated. The best individuals among the population alongside the new offspring form the new population, which will be used for the next iteration. Usually the population size is predefined based on the problem settings. The iterative process stops when the population has converged (not producing any offspring that are significantly different from the previous generation) or a satisfactory solution is found.

GA is an appealing approach to obtain high-quality solutions for an optimization problem thanks to its ability to deliver a good solution fast-enough. Please refer to \cite{katoch2020review} for a detailed review about GA. 

\noindent\textbf{Motivation 2:} Our purpose is to search a good complementary function, according to an architecture template with its operators selected from a list of candidate functions, the search space is discrete. Since the combination of good operators tends to produce a better solution, it better matches the process of GA. Additionally because of its simplicity and efficiency, we adopt GA in this work for searching the optimal activation function of BNNs.

\section{Methodology}
\label{s:Methodology}
\noindent
We first introduce the proposed encoding strategy and fitness function and then describe our GA-based searching approach in detail.

\subsection{Encoding Strategy and Search Space}
\noindent
The gene encoding strategy plays a key role in GA, which essentially dictates the search space. The search space needs to be sufficiently large such that a strong solution can be generated, but not too vast and complex that the majority of the solution space cannot be traversed. The definition of the search space is a trade-off between the capacity (i.e., more likely to incorporate a good solution in the space) and the complexity (i.e., more difficult to find a good solution in the space).

\begin{figure}[!htb]
\minipage{0.48\columnwidth}
\centering
\includegraphics[width=0.59\columnwidth]{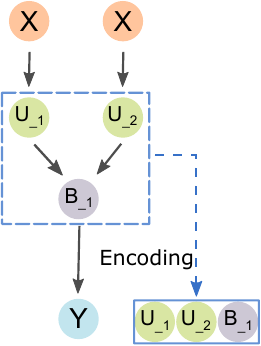} 
\caption{Type-I encoding (1st-order). $X$ is the input, $Y$ is the output. $U$ stands for unary operator, which has 22 candidate operators. $B$ stands for binary operator, which has 11 candidate operators.}
\label{fig:enc_1}
\endminipage\hfill
\minipage{0.480\columnwidth}
\centering
\includegraphics[width=0.73\columnwidth]{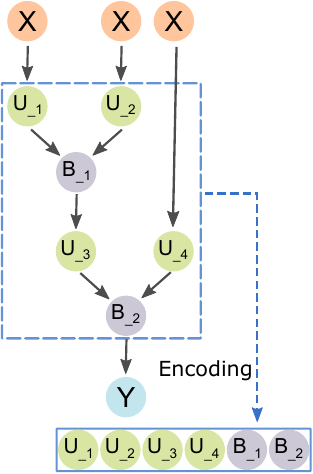} 
\caption{Type-II encoding (2nd-order).}
\label{fig:enc_2}
\endminipage\hfill
\end{figure}

In this paper, we represent the activation function as a graph that transforms an input tensor to an output tensor with the same shape (i.e., one-to-one mapping). To simplify the search process and constraint the size of the search space to a reasonable and computationally feasible size, we fix the graph structure in two patterns: \textbf{Type-I}, representing the first-order architecture, is shown in Figures~\ref{fig:enc_1}; \textbf{Type-II}, representing the second-order architecture, as shown in Fig.~\ref{fig:enc_2}. The order here refers to the number of combination operators as discussed later. Based on our observation, higher-order architectures show marginal benefits with substantially larger search space. 

In Fig.~\ref{fig:enc_1} and \ref{fig:enc_2}, $X$ is the input tensor of the complementary activation function. $Y$ is the output. Note that $X$ is duplicated for extracting and integrating information from different paths. $U$ represents a unary operator, which takes one tensor as input and one tensor as the output. The input and output tensors are in the same shape. $B$ represents a binary operator, which takes two tensors with the same size as the input, and one tensor as output also with the same size. Overall, the mapping from $X$ to $Y$ respects to one-to-one mapping.

For Type-I encoding shown in Fig.~\ref{fig:enc_1}, the replicated $X$ tensors pass through independent unary operators, and then combined through a binary operator to generate the output tensor $Y$. As for Type-II encoding in Fig.~\ref{fig:enc_2}, the three replicated $X$ tensors pass through their independent unary operators respectively, and then merge gradually through two binary operators to generate the final output tensor $Y$.
 
With the fixed graph structure, each candidate solution can be encoded as a fixed length string of values which represent the combination of different operators. Each value in the encoded string represents the index of a corresponding operator. For Type-I, the encoding output is a string with length of 3, representing 2 unary operators and 1 binary operator. The Type-II encoding, in contrast, has 6 values which include 4 unary operators and 2 binary operators. Regarding the format of the string, all unary operators stay on the left and binary operators stay on the right, which essentially yields benefits in the following processing which will be discussed later.

A complete list of unary and binary operators are summarized as follows:

\begin{itemize}
    \item  \textbf{Unary Operator:} $x$, $|x|$, $-x$, $0$, $x^2$, $x^3$, $sign(x)*\sqrt{|x|}$, $log(|x|)$, $1.0/(1+e^{-x})$, $e^{(-|x|)}$, $e^{-x^2}$, $sin(x)$, $cos(x)$, $tan(x)$, $atan(x)$, $erf(x)$, $erfc(x)$, $max(x,0)$, $min(x,0)$, $\alpha$, $\alpha*x$, $\alpha+x$.
    \item  \textbf{Binary operator:} $x+y$, $x-y$, $x*y$, $x/y$, $x/(x+y)$, $max(x,y)$,  $min(x,y)$, $x/(1+e^{-y})$, $e^{(-|x-y|)}$, $e^{(-(x-y)^2)}$, $\beta*x+(1-\beta)*y$. 
\end{itemize}
where $\alpha$ and $\beta$ are learnable channel-wise parameters, which can be jointly optimized with other parameters in the neural networks through back-propagation. $0$ in Unary operators means the output is always 0, which enables the pruning of redundant operators. 

Overall, there are 22 candidate unary operators and 11 candidate binary operators. Therefore, the value of gene-$U$ thus can go from 0 to 21, where each value corresponds to an operator in the list. The same condition applies to gene-$B$. Those operators are scalar to scalar which are easy to implement and lower the computation complexity. Using such a fixed graph structure, combinations of different unary and binary operators can form distinct candidates and constitute the search space. Although the search space is drastically shrunk with the fixed graph structure (i.e., architecture), it is still massive: for Type-I, the search space is 22$\times$22$\times$11=5.3K; for Type-II, the search space is: $22^4$$\times$$11^2$=28.3M.

\subsection{Fitness Metric}
\noindent
The fitness metric in GA is to assign a fitness value for each individual in the population, which indicates how good a candidate solution is. The fitness score also determines the ability of an individual to compete with other individuals during selection. The fitness function is designed according to the problem, which is usually the value returned by the objective function of the optimization problem.

In this paper, we focus on searching for promising generalized complementary activation functions for BNNs. The image classification is used to evaluate the fitness, the top-1 validation accuracy will be the fitness value. Since not all well-known real-valued models can be seamlessly applied to BNNs, and given the widely adoption of ResNet structure in BNNs (the shortcuts are essential for rich information flow tackling the information loss from binarization and gradient vanishing/explosion for the BNN latent variables), we use the classical ResNet18 model to perform the evaluation. As each trial assessment of the fitness evaluation requires the training of a BNN model, to be efficient, the BNN model is trained on CIFAR10 dataset. In particular, the top-1 validation accuracy after the BNN version of Resnet18 trained on CIFAR10 for 15 epoches will be assigned to each candidate individual as the fitness value.

\vspace{4pt}\noindent\textbf{Early Rejection:} Since the candidate activation function can be generated pseudo-randomly, some of them may result in non-convergence during training, we adopt the early rejection strategy. We set a rejection threshold $T_1$, whereby when the validation accuracy is less than $T_1$ after the first epoch of training, the candidate individual will be discarded without further training, and we move-on to the next step. Otherwise, the training is continued up to 15 epochs. With this early rejection strategy, the search process can be much more efficient.

\subsection{GA Implementation}
\begin{figure}[!t]
\centering
\includegraphics[width=0.7\columnwidth]{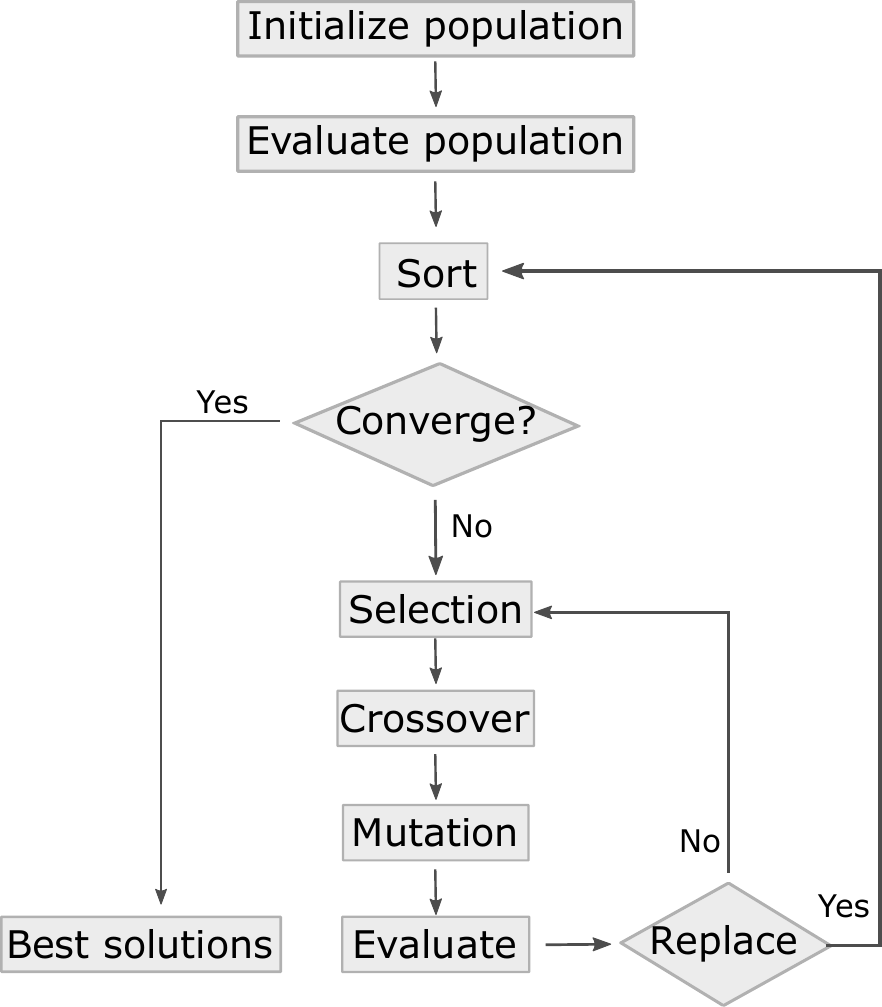} 
\caption{The workflow of the proposed genetic algorithm.}
\label{fig:flow}
\end{figure}

\noindent
Fig.~\ref{fig:flow} illustrates the implementation workflow of GA. By defining the encoding strategy and fitness function, the initial population of size $S$ can be generated. Each individual of the population is evaluated and assigned with a fitness value. The population is then sorted according to the fitness values from high to low. Therefore, the first individual appears to show the best fit. From here, the evolution iteration starts. In each iteration, we first choose two parent individuals and then a new offspring is generated by combining the two parent individuals with crossover and mutation. The new offspring is evaluated to obtain its fitness value. 

This fitness value is then compared with the fitness value of the last individual in the population: \textbf{(i)} if the the new offspring shows a higher fitness value, the last individual is replaced by the new offspring, then the new population is sorted in descending order based on its fitness values. The next iterator then begins using the new population. \textbf{(ii)} Otherwise, the next iteration starts using the old population. During the evolution process, the total population size keeps unchanged ($S$). Once the population has converged meaning that no more satisfactory solutions can be generated, the iterative search terminates and a series of optimal solutions are reported.

We describe each process of Fig.~\ref{fig:flow} in detail.

\vspace{4pt}\noindent\textbf{Initialization:}
To ensure the diversity of the population, individuals in the initial population are generated randomly. For Type-I (Type-II) encoding, the first two (four) genes are randomly sampled from the 22 unary operators; the last one (two) gene is randomly sampled from the 11 binary operators. The fitness value is then evaluated for the generated individual. This process of generation continues until the size of population reaches to $S$. Then the population is sorted in descending order based on the fitness values.

\vspace{4pt}\noindent\textbf{Selection:}
Selection is a critical step in GA. The convergence rate of GA largely depends on the selection pressure. Selection involves choosing the parent individuals from the population which will be used for reproduction of the new offspring. The selection process is mostly based on fitness. For diversity, each individual should have the chance to be selected. Nevertheless, the theory of "Survival of the Fittest" of GA gives preference to the individuals with higher fitness values; those individuals have a higher chance of being selected for reproduction. 

There are several well-known selection techniques, such as elitism, roulette wheel, rank, tournament, Boltzmann, stochastic universal sampling, etc. Each selection technique has its advantages and drawbacks. Taking into consideration the variability and efficiency of the algorithm, three selection techniques -- \emph{elitism selection}, \emph{tournament selection}, and \emph{proportionate selection} -- are adopted here. In each iteration, one of the three selection techniques is randomly picked for selecting the two parent individuals. We discuss each selection technique:

\textbf{\emph{1) Elitism selection}} chooses the two best individuals based on their fitness values. It always chooses the best individuals in the population for reproduction, which leads to a better chance of reproducing a better individual. However, Elitism may result in early convergence due to being trapped in a local maximum. 

\textbf{\emph{2) Tournament selection}} chooses individuals according to their fitness values from a stochastic roulette wheel in pairs. The first individual is randomly selected from the population; the second individual is randomly picked from the right part of the first one in the population. Therefore, the second individual has a lower fitness than the first one. Both the two individuals are chosen at random, which ensures each individual has a chance to be selected.

\textbf{\emph{3) Proportionate selection}} is implemented by describing the fitness value as a probability distribution over the population. Then two random individuals are selected from the distribution. Due to the transformation, the chance of each individual being selected is proportional to its fitness value.

Elitism selection prefers the best individuals whereas tournament selection and proportionate selection  stochastically choose individuals. The probabilistic combination of those three selection techniques can take the advantages of each technique, while avoiding their drawbacks, accelerating the convergence speed.


\vspace{4pt}\noindent\textbf{Crossover:} The crossover step generates new offspring by combining different portions of two or more selected parent individuals. The well-known crossover operators include \emph{single-point}, \emph{two-point}, \emph{k-point}, \emph{uniform}, \emph{partially matched}, \emph{ordered}, \emph{precedence preserving}, \emph{shuffle}, etc.
In this work, the single-point crossover is adopted due to its simplicity and the relatively simple settings of our GA problem. 

\begin{figure}[!t]
\centering
\includegraphics[width=0.6\columnwidth]{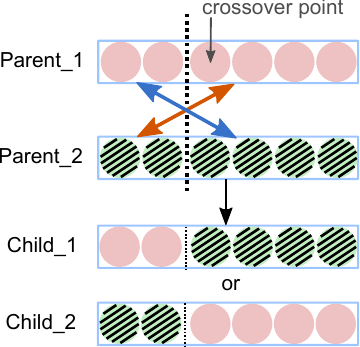} 
\caption{Single-point crossover.}
\label{fig:cross}
\end{figure}

Fig.~\ref{fig:cross} illustrates how crossover is performed between the two parent individuals in this paper. The crossover point is randomly picked. According to its position, both of the parent individuals are split into a left part and a right part. The new offspring is then formed by combining the left part of $Parent_1$ with the right part of $Parent_2$, or by combining the left part of $Parent_2$ with the right part of $Parent_1$. The new offspring has the same size as its parents.

\vspace{4pt}\noindent\textbf{Mutation:} Mutation is essential for GA. The key idea is to insert random genes in offspring to maintain the diversity of the population for avoiding premature convergence. The well-known mutation operators include \emph{displacement}, \emph{simple inversion}, and \emph{scramble mutation}. 
Displacement mutation displaces a substring of a given individual within itself. Simple inversion mutation reverses the substring between any two specified locations of an individual solution. Scramble mutation places the elements in a specified range of an individual solution in a random order. 

\begin{figure}[!t]
\centering
\includegraphics[width=0.6\columnwidth]{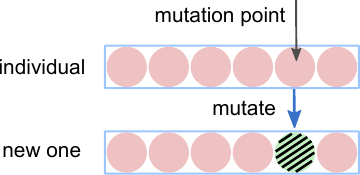} 
\caption{Single-point mutation.}
\label{fig:mut}
\end{figure}

In our GA design, we use single point mutation, which is one variant of displacement mutation, as illustrated in Fig.~\ref{fig:mut}. First, a single mutation point is randomly selected from the individual. Then the chosen gene is then replaced by a new randomly generated gene. Based on the mutation point, we figure out whether the mutation refers to operator $U$ or $B$. If $U$, a replacement gene is randomly generated from the 22 unary operators; if $B$, the replacement gene is randomly generated from the 11 binary operators.

Note that the mutation operator is capable of maximizing the variance of the newly generated individuals, which is important in finding the global optimal solution.

\section{Experiments}
\label{s:EXperiments}
\subsection{GA Setting}
\noindent
We perform the GA-based searching for optimal BNN complementary activation functions. The BNN version of ResNet18 with candidate complementary activation function is trained on CIFAR10. We use the standard pre-processing, the batch size is 128. The Adam optimizer with $betas=(0.9,0.999)$ is  adopted with a constant learning rate 5e-3. After 15 epochs of training, the top-1 validation accuracy is assigned to the candidate individual as its fitness value.

Like most BNNs, the first and last layer of BNN models used in this paper should always be left in full-precision. This is because the first layer extracts the initial information from the input images, whereas the last layer immediately dictates the final prediction -- both require a higher capacity in order to show good performance.

The early rejection threshold $T_1$ is empirically set as 11\% in our fitness evaluation. With the early rejection strategy, the compute resources can be allocated focusing on the most promising candidates. To further enhance searching efficiency, the threshold can be slowly increased as the search continues, so that it can early reject more unqualified candidate individuals and preserve promising solutions. This is based on the intuition that as time goes on, the descendants of the GA tend to be more well-behaved than in the early stage of the search. Increasing the early rejection threshold thus improves the search efficiency. As the searching goes on, the threshold $T_1$ increases to be 25\%, 35\%, 40\%. The size of population $S$ is set to be 30.

\subsection{Activation Functions found by GA Search}
\noindent

\begin{table}[htbp]
  \centering
  \begin{tabular}{  | c| c|c|}
    \hline
    \textbf{Type} & \textbf{Function}  & \textbf{Graph} \\ [3pt]
	\hline
	RSign\cite{liu2020reactnet} & $  \begin{cases}
    +1 & x\geq \alpha \\
    -1 & x \textless \alpha  \\ 
  \end{cases}     $ & \begin{minipage}[b]{0.12\columnwidth}
		\centering
		\raisebox{-.7\height}{\includegraphics[width=\linewidth]{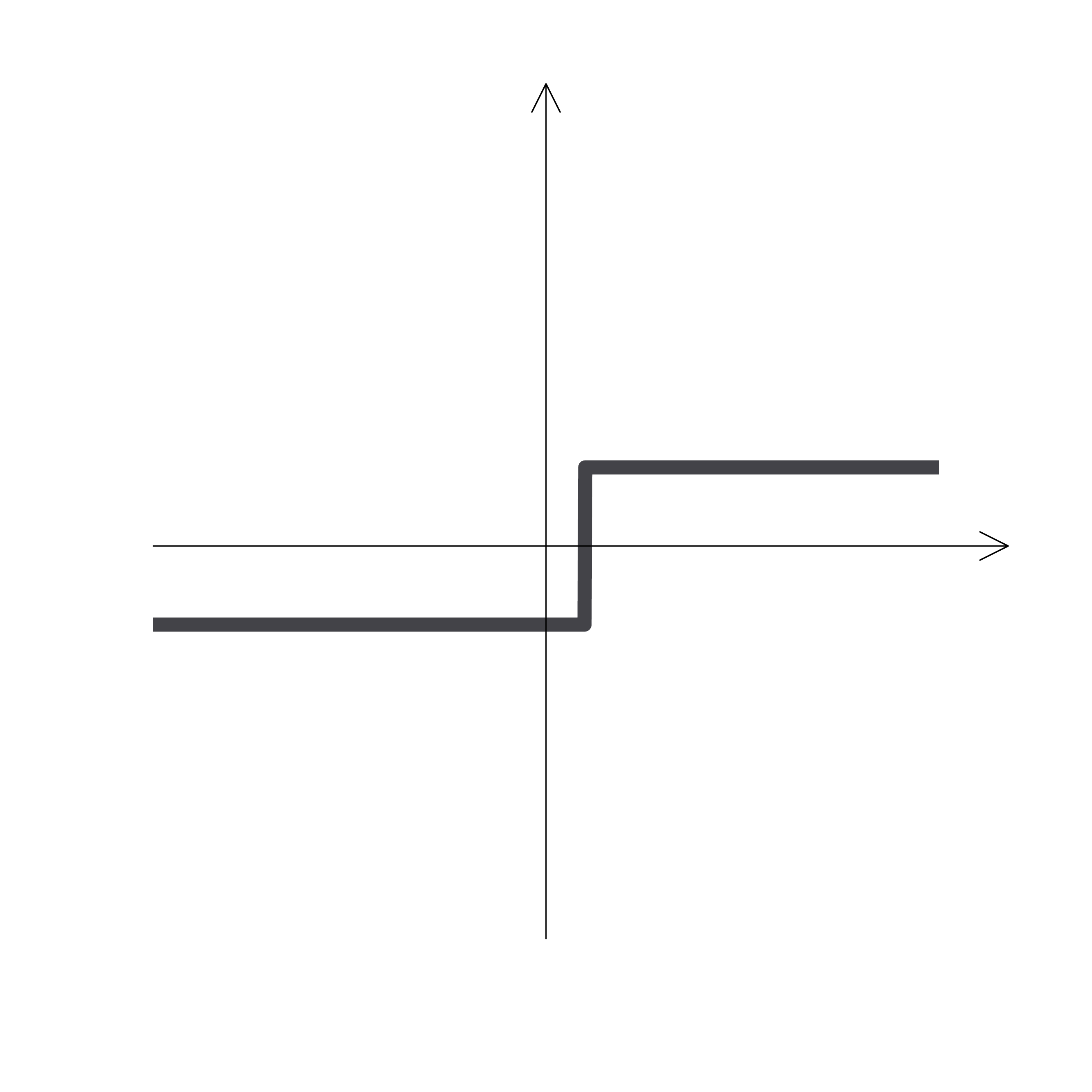}}
	\end{minipage} \\   [20pt]
	\hline
	RPReLU\cite{liu2020reactnet} & $ \begin{cases}
    x-\gamma+\zeta & x\geq \gamma \\
    \beta*(x-\gamma)+\zeta & x \textless \gamma  \\
  \end{cases}  $ & \begin{minipage}[b]{0.12\columnwidth}
		\centering

		\raisebox{-.7\height}{\includegraphics[width=\linewidth]{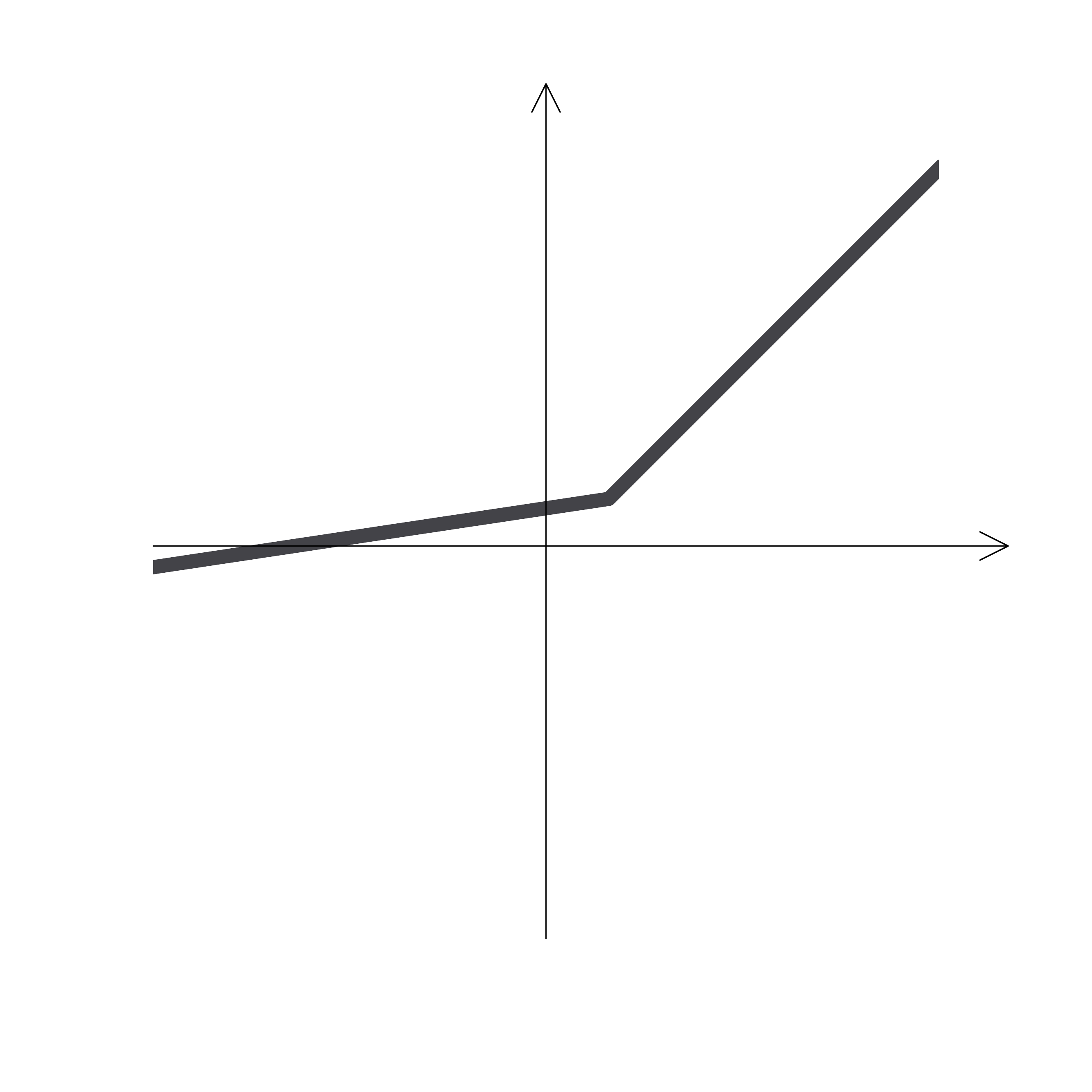}}
	\end{minipage} \\  [20pt]
    \hline
    AF1 & $ sin(x)-cos(x) $ & \begin{minipage}[b]{0.12\columnwidth}
		\centering
		\raisebox{-.7\height}{\includegraphics[width=\linewidth]{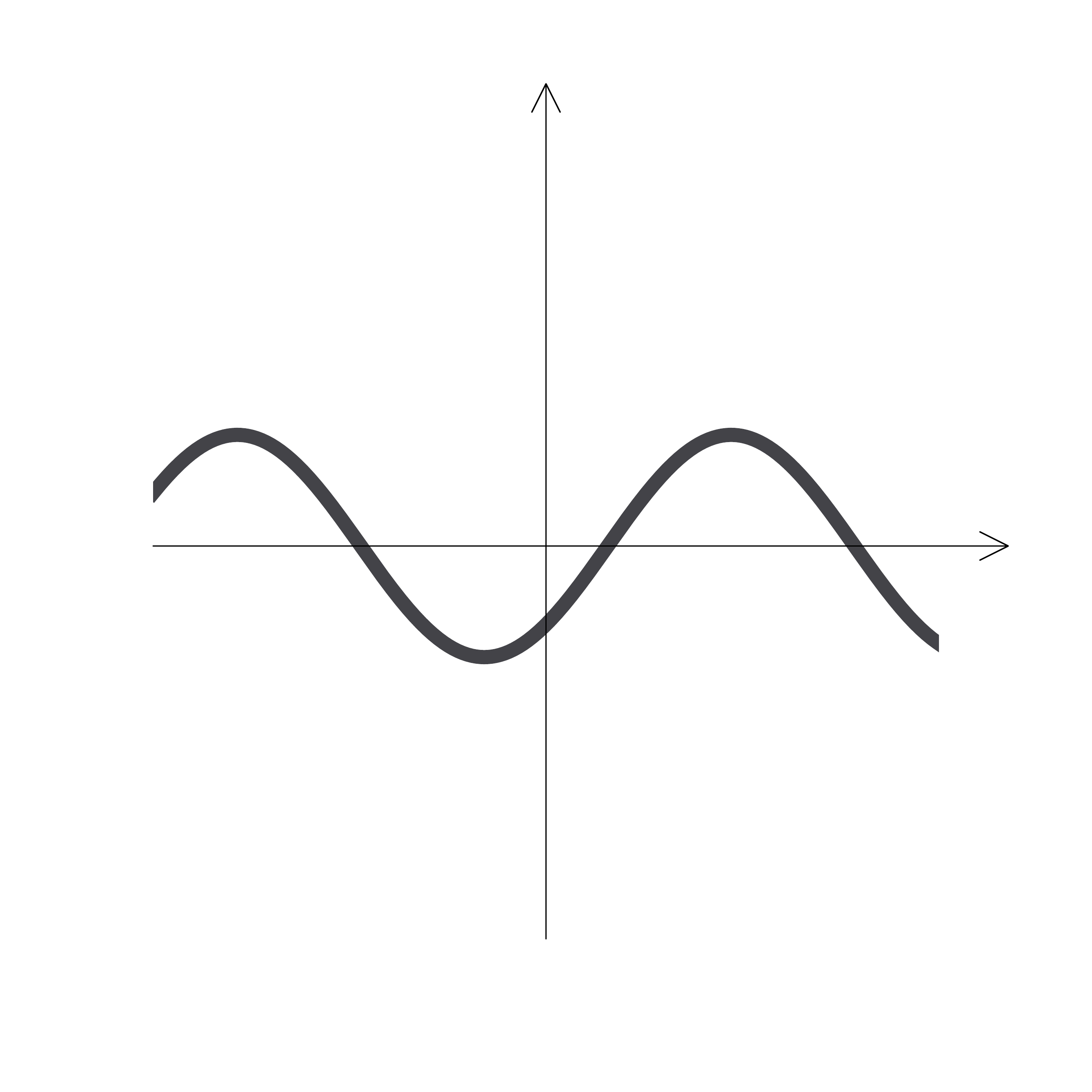}}
	\end{minipage} \\   [20pt]
	\hline
  AF2  & $ sin(x)+cos(x) $  & \begin{minipage}[b]{0.12\columnwidth}
		\centering
		\raisebox{-.7\height}{\includegraphics[width=\linewidth]{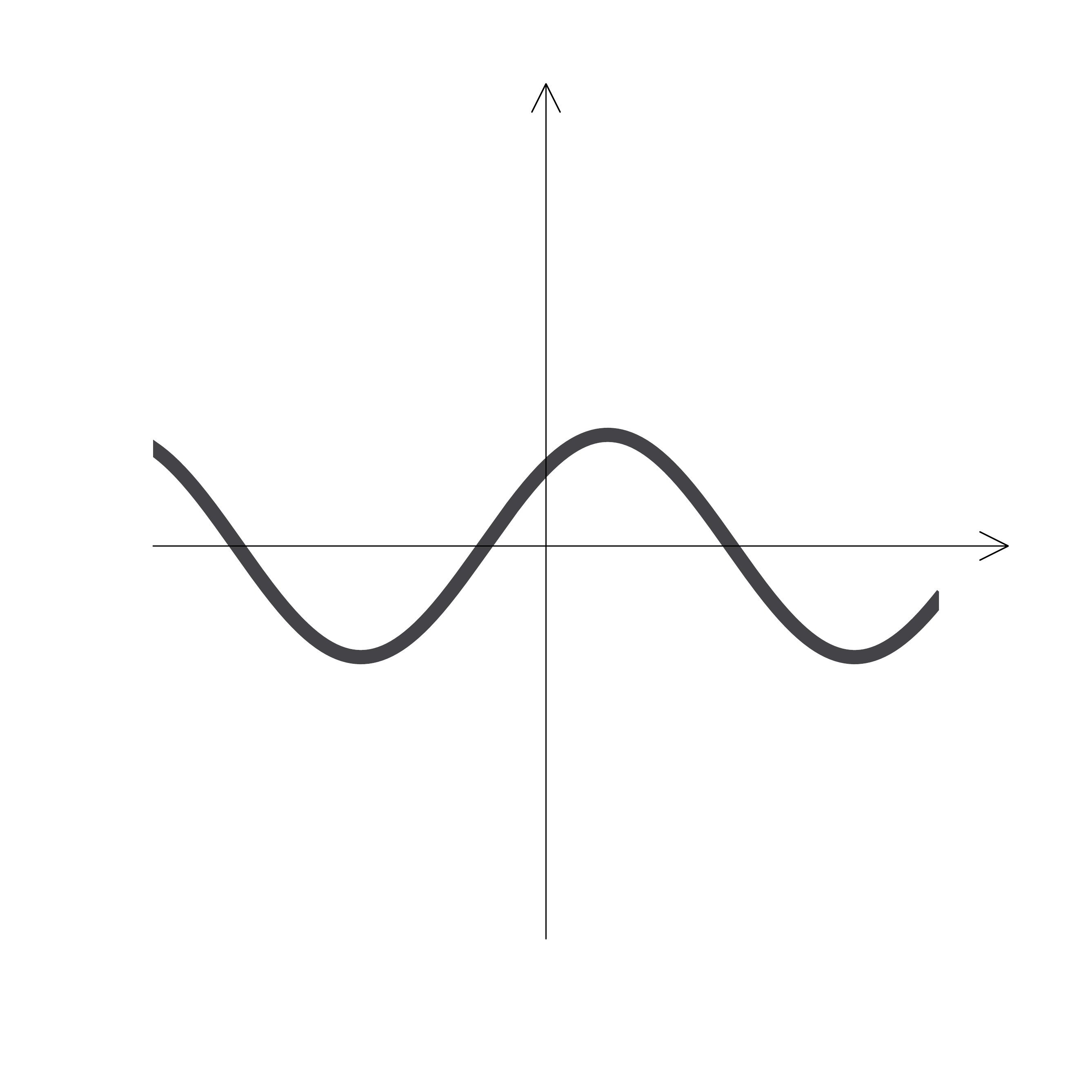}}
	\end{minipage}  \\ [20pt]
  \hline
  AF3 & $ max(x,0) + sin(x) $ &  \begin{minipage}[b]{0.12\columnwidth}
		\centering
		\raisebox{-.7\height}{\includegraphics[width=\linewidth]{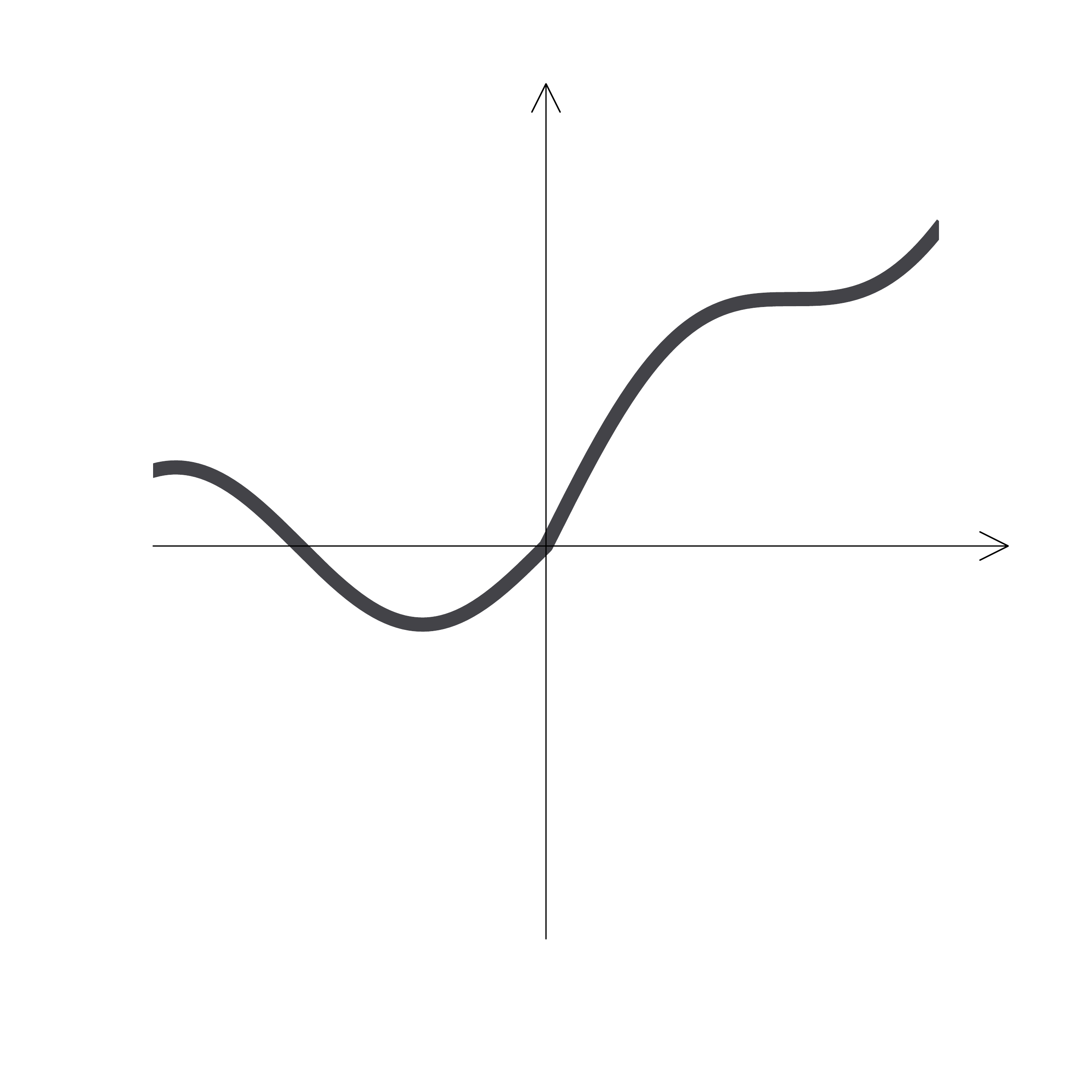}}
	\end{minipage} \\ [20pt]
  \hline
AF4 & $ \beta*cos(x) + (1-\beta)*x $ & \begin{minipage}[b]{0.12\columnwidth}
		\centering
		\raisebox{-.7\height}{\includegraphics[width=\linewidth]{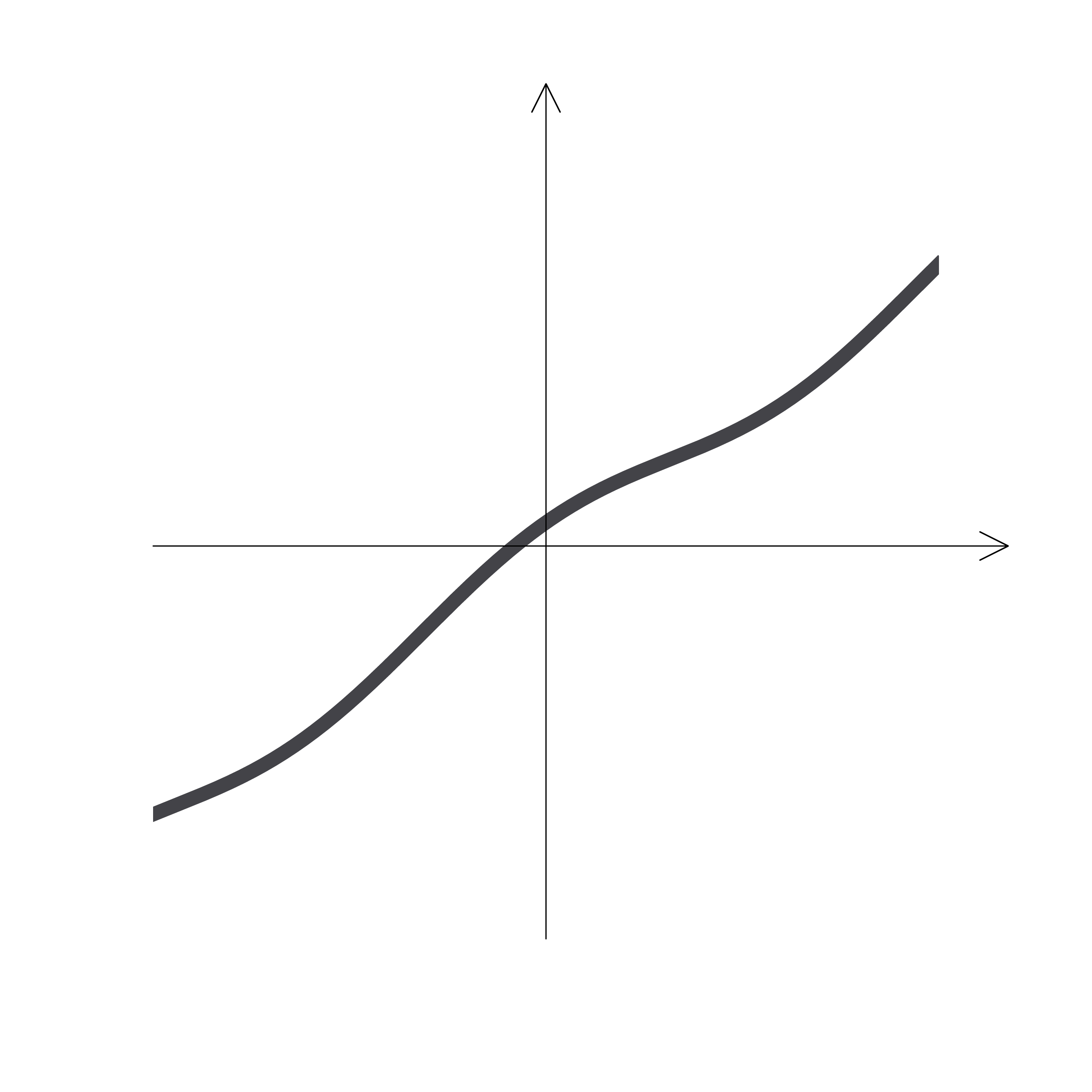}}
	\end{minipage}  \\ [20pt]
  \hline
  AF5 & $ min(x,0) + sin(x) $ &  \begin{minipage}[b]{0.12\columnwidth}
		\centering
		\raisebox{-.7\height}{\includegraphics[width=\linewidth]{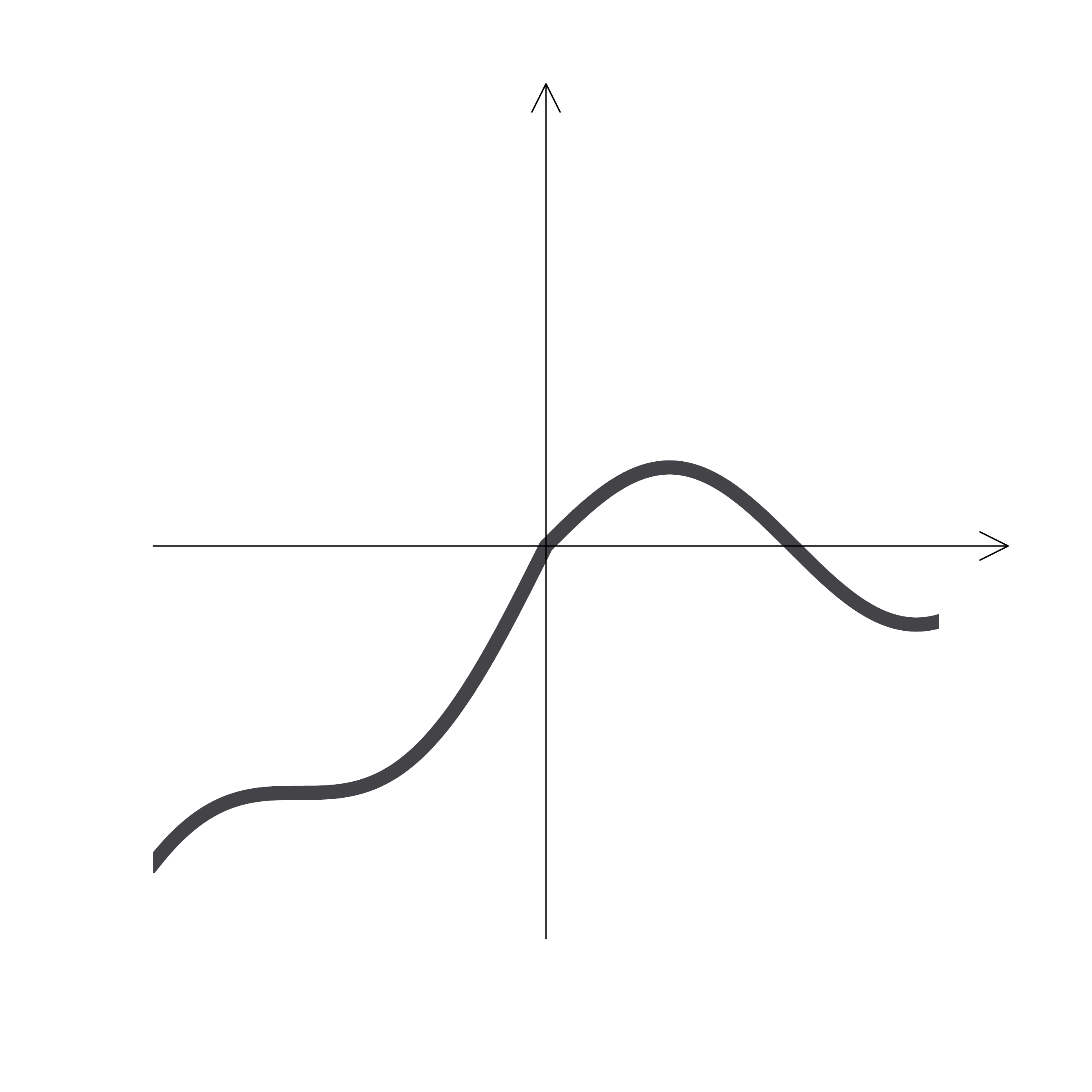}}
	\end{minipage} \\  [20pt]
  \hline
  AF6 & $ \beta*erf(x) + (1-\beta)*max(x,0) $ &   \begin{minipage}[b]{0.12\columnwidth}
		\centering
		\raisebox{-.7\height}{\includegraphics[width=\linewidth]{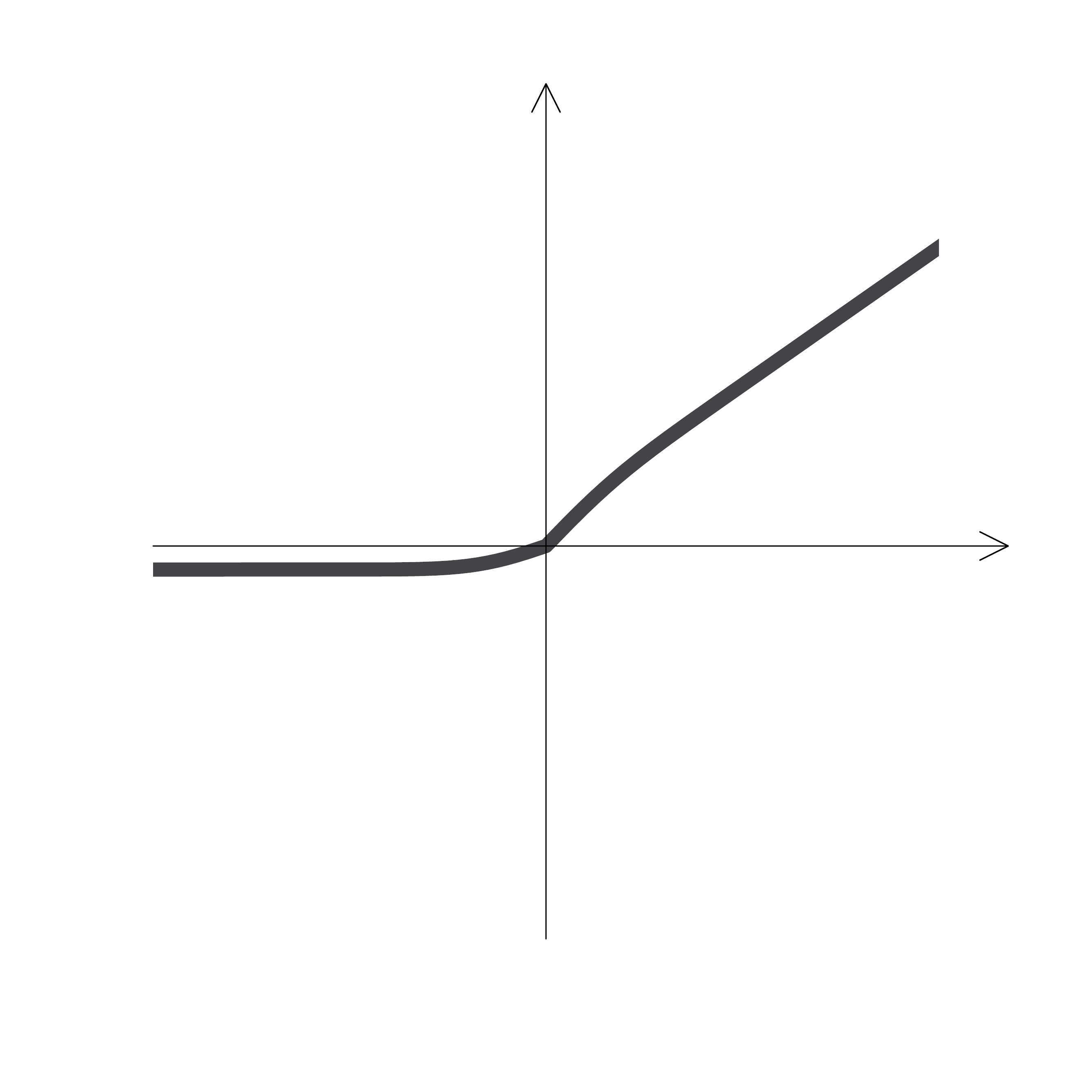}}
	\end{minipage} \\  [20pt]
  \hline
 AF7 & $ exp(-x^2) - sin(x) $ &   \begin{minipage}[b]{0.12\columnwidth}
		\centering
		\raisebox{-.7\height}{\includegraphics[width=\linewidth]{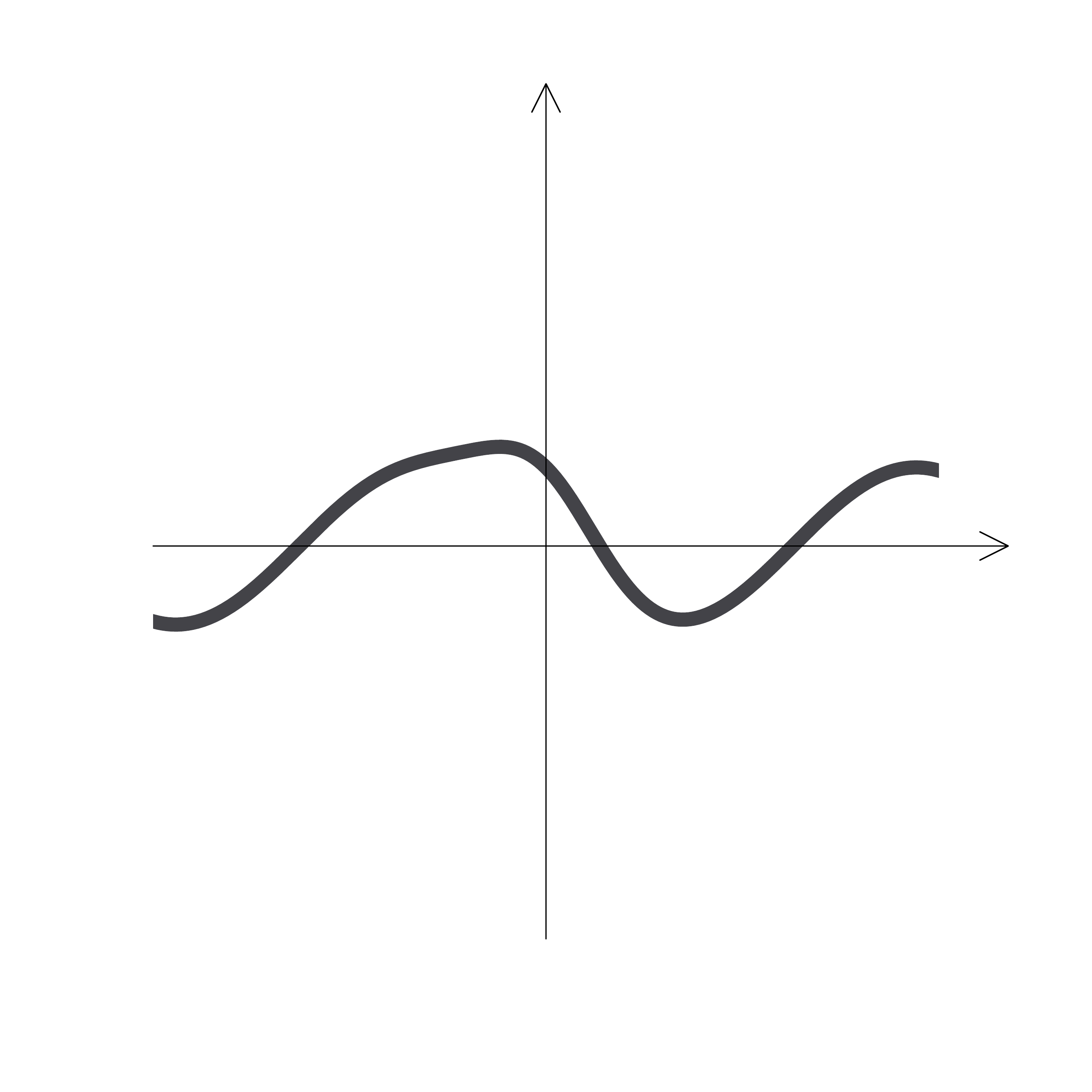}}
	\end{minipage} \\  [20pt]
  \hline
  AF8 & $ cos(x) + atan(x) $ &  \begin{minipage}[b]{0.12\columnwidth}
		\centering
		\raisebox{-.7\height}{\includegraphics[width=\linewidth]{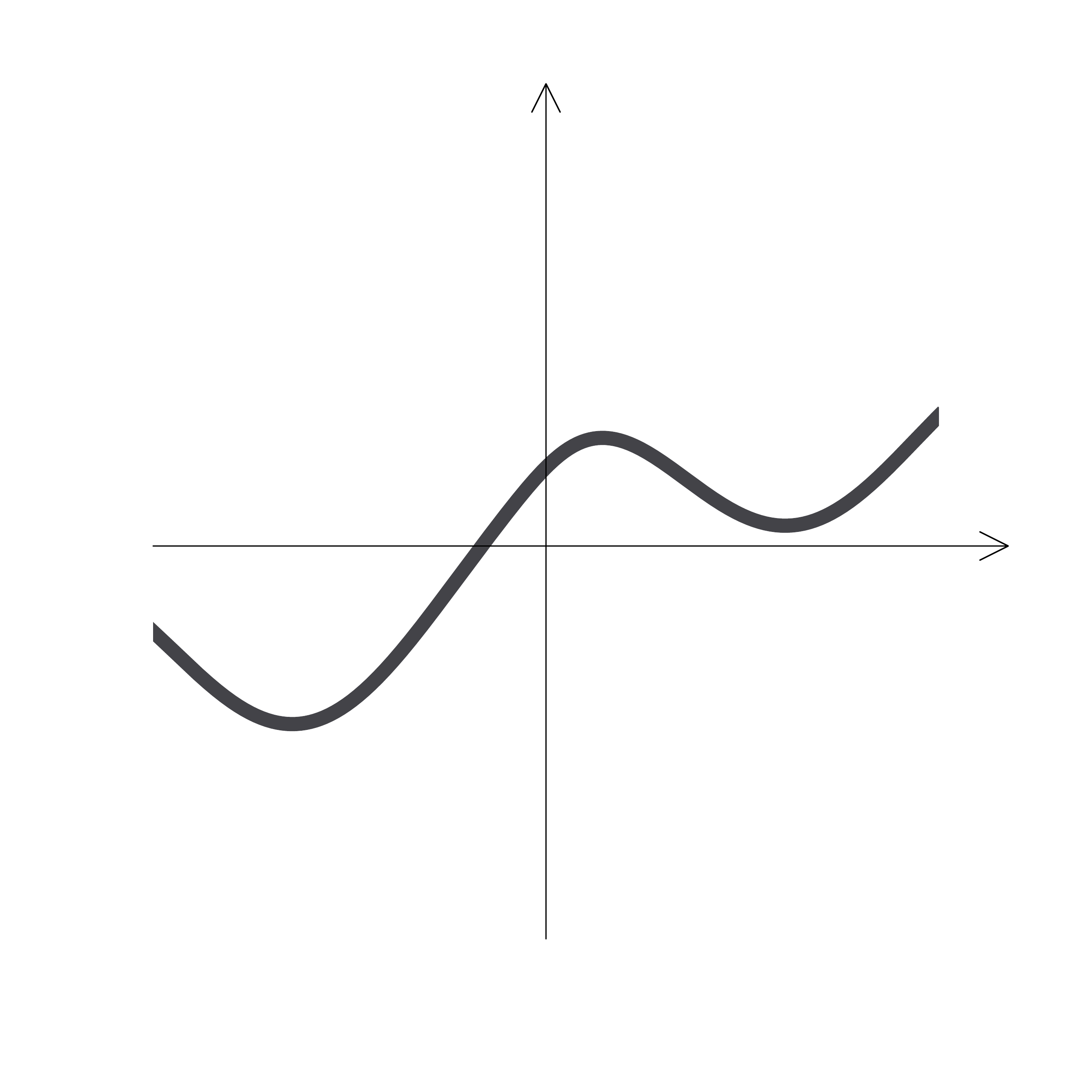}}
	\end{minipage}  \\  [20pt]
  \hline
  AF9 & $ \beta*cos(x) + (1-\beta)*atan(x) $ &  \begin{minipage}[b]{0.12\columnwidth}
		\centering
		\raisebox{-.7\height}{\includegraphics[width=\linewidth]{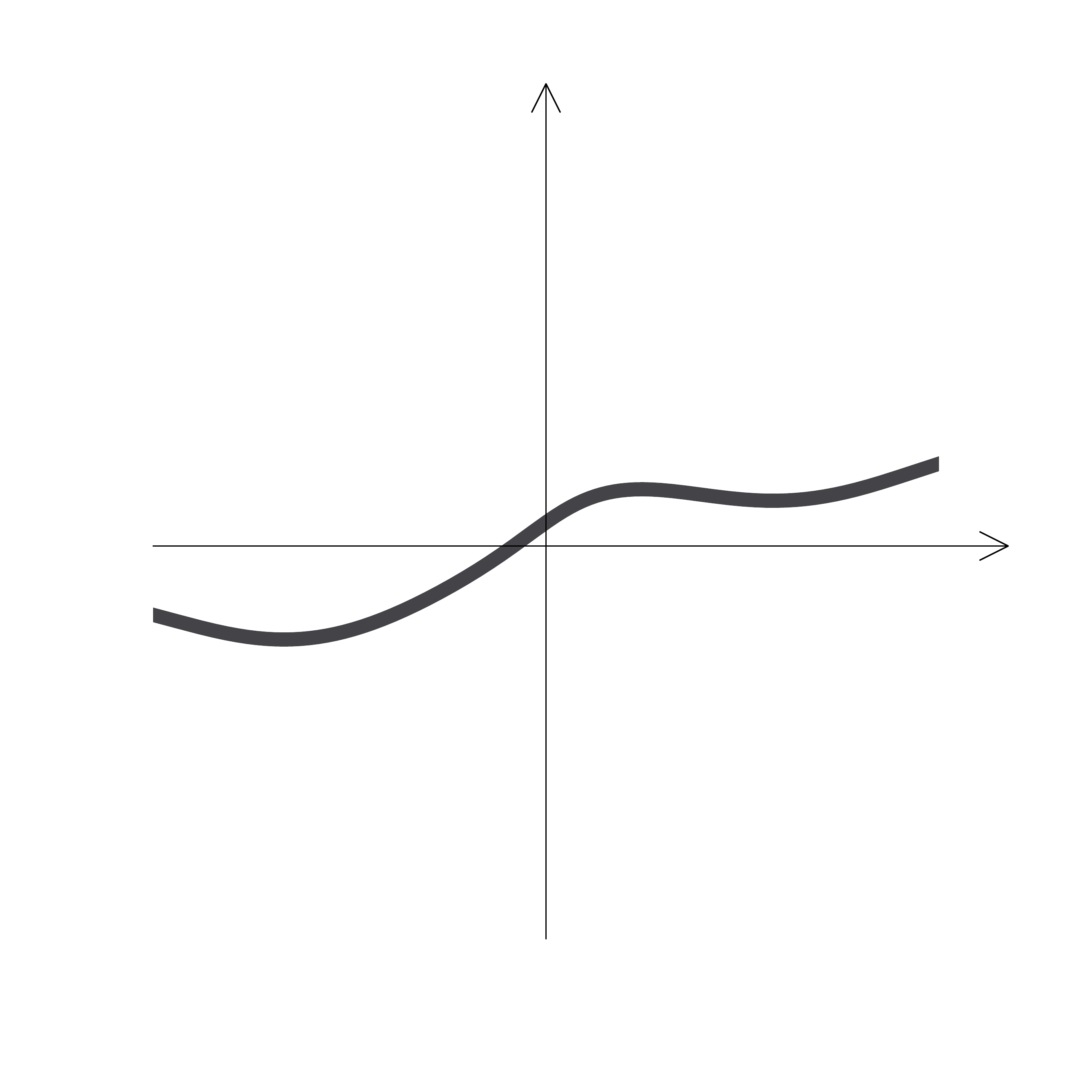}}
	\end{minipage}   \\  [20pt]
  \hline
  AF10 & $ cos(x) - atan(x) $ &  \begin{minipage}[b]{0.12\columnwidth}
		\centering
		\raisebox{-.7\height}{\includegraphics[width=\linewidth]{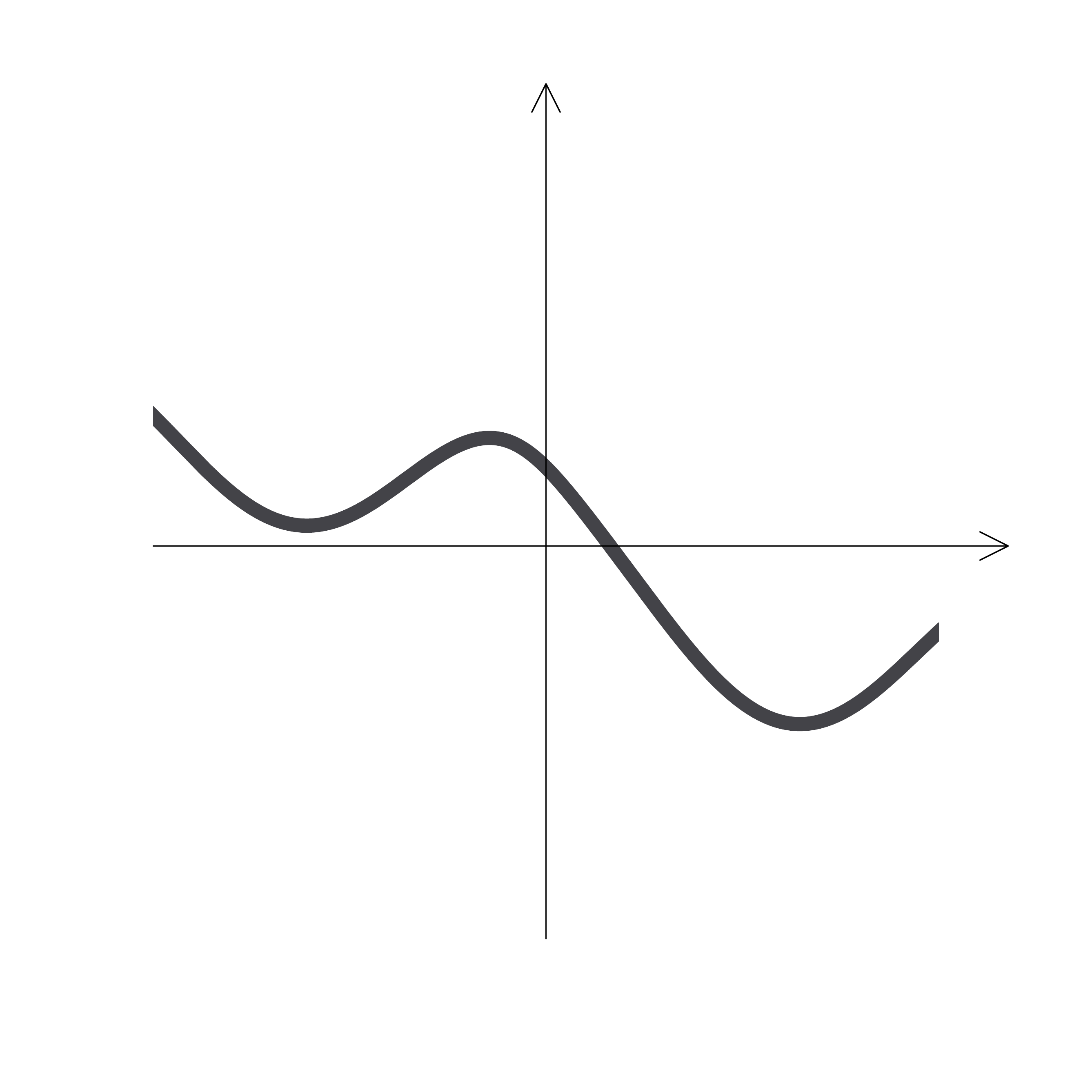}}
	\end{minipage}   \\  [20pt]
  \hline
  AF11 & $ cos(atan(x)/2) + x $  &  \begin{minipage}[b]{0.12\columnwidth}
		\centering
		\raisebox{-.7\height}{\includegraphics[width=\linewidth]{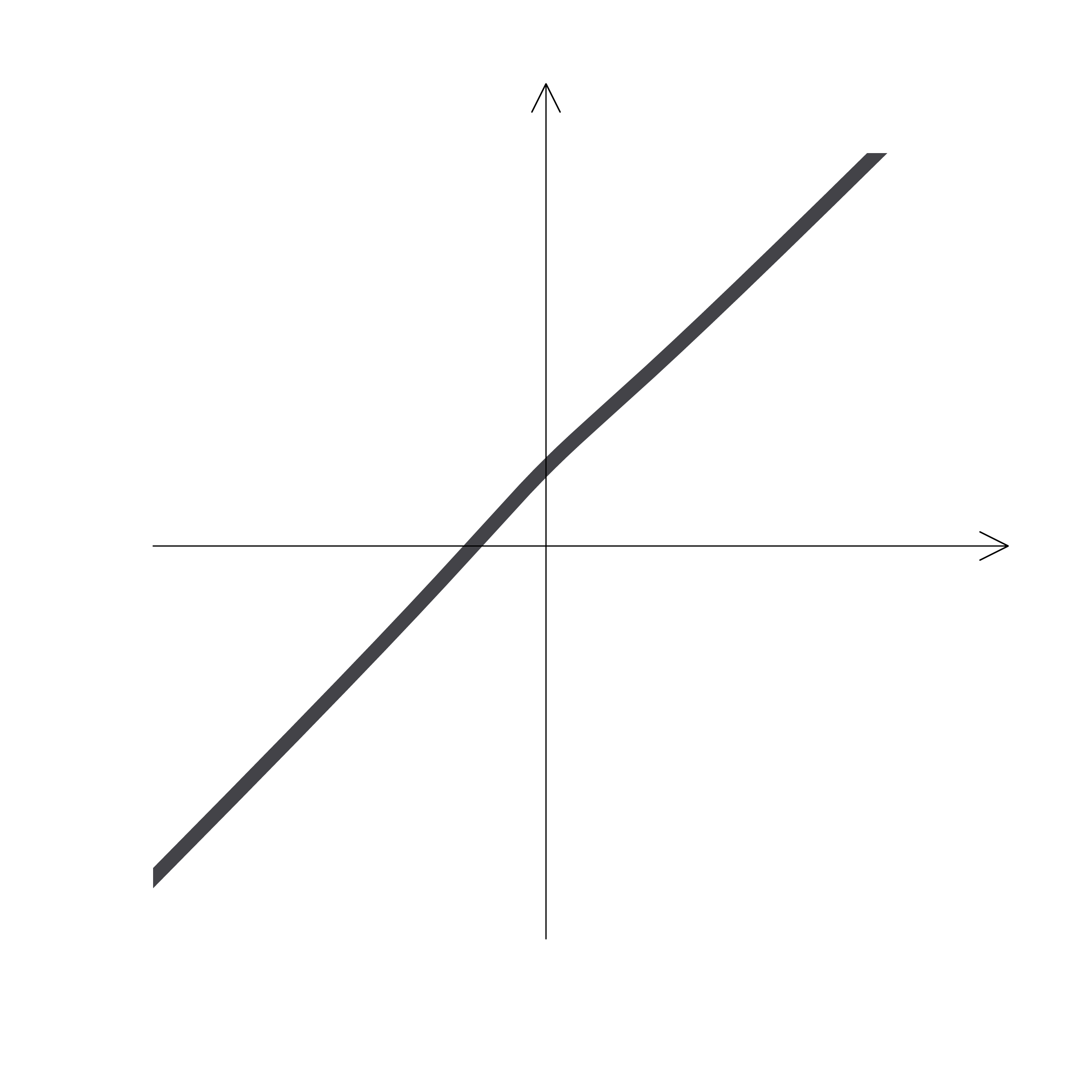}}
	\end{minipage}  \\  [20pt]
  \hline
  AF12 & $ \beta* cos(x + \alpha) + (1-\beta)*x $ &  \begin{minipage}[b]{0.12\columnwidth}
		\centering
		\raisebox{-.7\height}{\includegraphics[width=\linewidth]{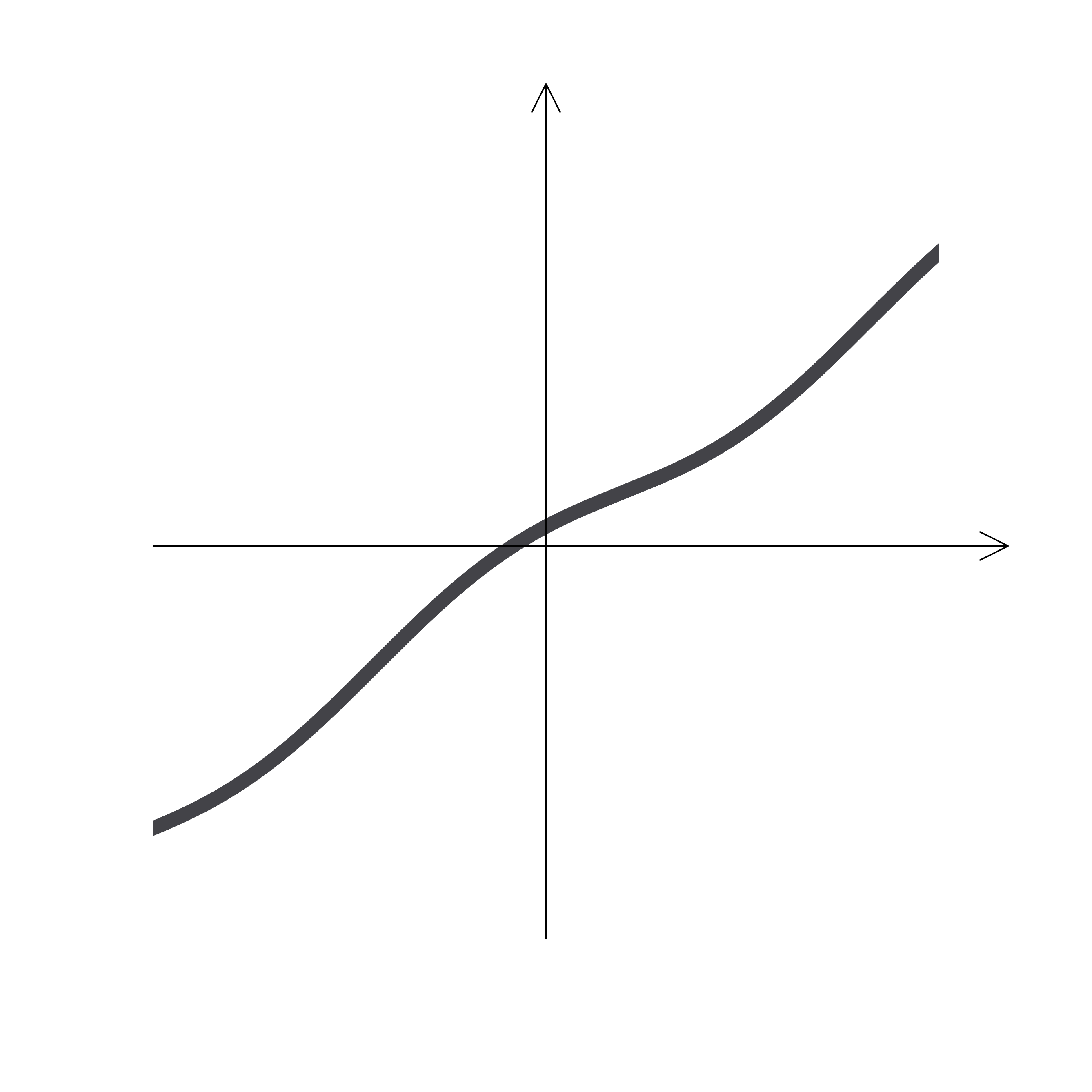}}
	\end{minipage}  \\  [20pt]
  \hline
 AF13 & $ cos(atan(x)) + x $ &  \begin{minipage}[b]{0.12\columnwidth}
		\centering
		\raisebox{-.7\height}{\includegraphics[width=\linewidth]{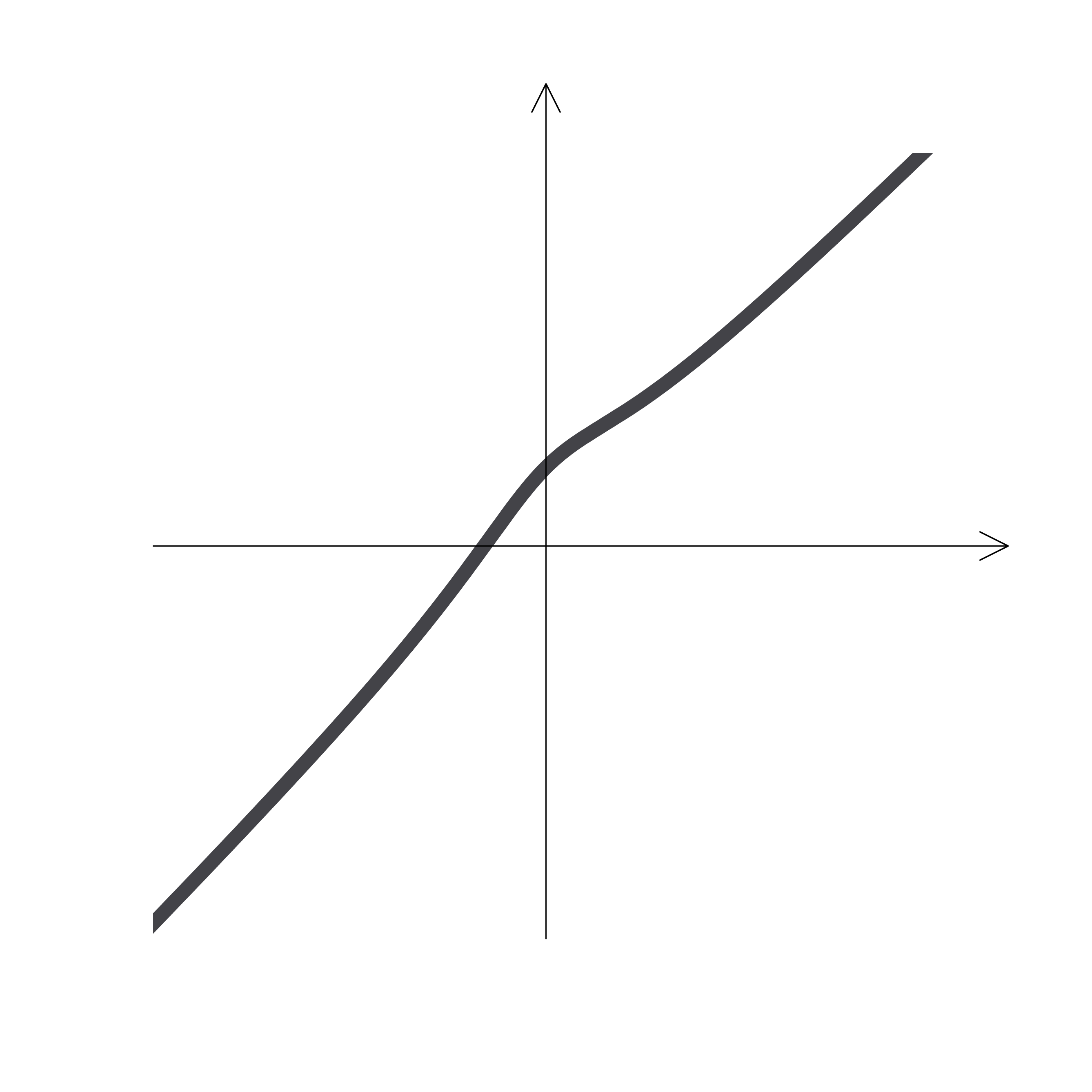}}
	\end{minipage}  \\   [20pt]
  \hline
  AF14 & $ cos(erf(x)) -x $ & \begin{minipage}[b]{0.12\columnwidth}
		\centering
		\raisebox{-.7\height}{\includegraphics[width=\linewidth]{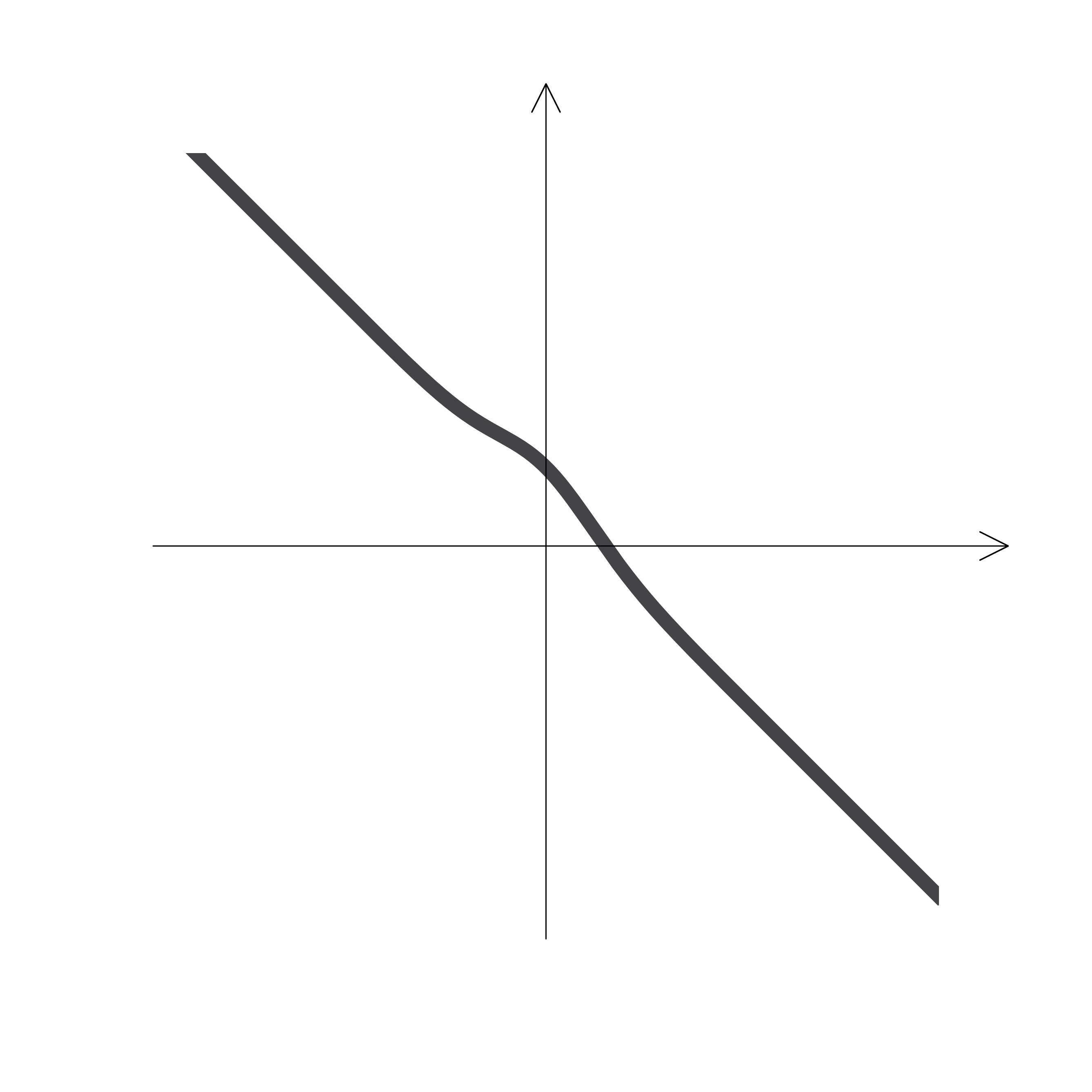}}
	\end{minipage}  \\  [20pt]
  \hline
  AF15 &$ cos(-x) + x $ &  \begin{minipage}[b]{0.12\columnwidth}
		\centering
		\raisebox{-.7\height}{\includegraphics[width=\linewidth]{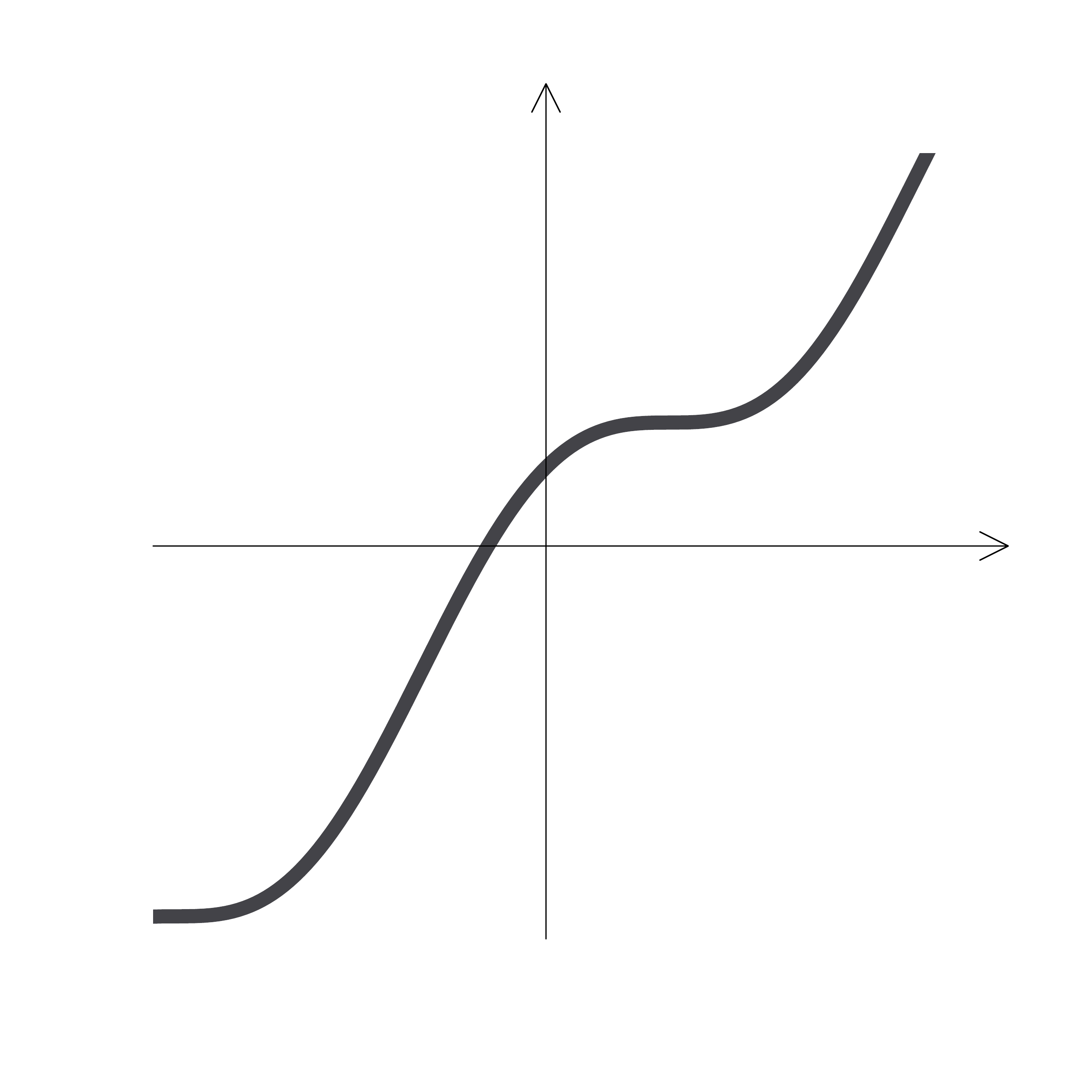}}
	\end{minipage}  \\   [20pt]
  \hline
  
  \end{tabular}
  \caption{Top discovered activation functions and graphs}
  \label{tab:fun and graph}
\end{table}

Through the GA-based searching, we identified multiple new complementary activation functions for BNNs, as listed in Table~\ref{tab:fun and graph}. Specifically, the first two are the hand-crafted \emph{RSign} and \emph{RPReLU} from \cite{liu2020reactnet} discussed in Section~2.1, shown for comparison purposes. The 10 activation functions from AF1 to AF10 are obtained in the GA search with Type-I encoding. The last 5 AFs from AF11 to AF15 are obtained in the GA search with Type-II encoding. We show the math expression and the graph in Table~\ref{tab:fun and graph} as well. Looking at the table, we draw the following observations:
\begin{itemize}
    \item These obtained top performing AFs are all relatively simple in their expressions. This is potentially because complicated function increases difficulties in optimization.  
    \item Counterintuitively, trigonometric functions such as $\sin$ and $\cos$ exist in almost all the AFs (except AF6). This observation indicates that incorporating periodic functions can extract more information from the inputs and enhances the representativeness of the activation function. 
    \item For the binary operator $B$, the linear combination such as $x+y$, $x-y$, $\beta*x+(1-\beta)*y$ exist more often than other complex binary operators. This might because alternative binary operators may lead to instant and significant jitters/pulses, or discrete gradients, which jeopardizes the stability of the training.  
    \item For Type-II encoding (Fig.~\ref{fig:enc_2}), all the searched AFs (AF11-AF15) are having their $U_4$ gene being $x$ or $-x$. This may imply that although the unary operator can help in extracting information, the raw input $x$ contains the most useful information and should always be kept as the residual in the combination.
\end{itemize}

\subsection{Testing the Searched Activation Functions}
\noindent
To showcase the performance and generalization of the obtained novel complementary activation functions, we test them on CIFAR10 and ImageNet with different BNN models. As mentioned before, the first and last layer of BNN models are left in full-precision. For each model, we evaluate each AF in Table~\ref{tab:fun and graph}. The original BNN is used as the baseline. The two hand-crafted AFs (\emph{RSign} and \emph{RPReLU}) from \cite{liu2020reactnet} are used for comparison. All models are trained using the same hyperparameters except the activation functions. The Adam optimizer \cite{kingma2014adam} with $betas=(0.9, 0.999)$ is adopted for the training.

\vspace{4pt}\noindent\textbf{CIFAR10:} We tested all the AFs in Table~\ref{tab:fun and graph} on CIFAR10 with 3 widely used BNN models: \emph{ResNet18}, \emph{NIN-Net} and \emph{ResNet34}. We use standard pre-processing. All models are trained until 300 epochs. The initial learning rate is 5e-3, and is shrunk by a factor of 0.2 at the 80th, 150th, 200th, 240th, and 270th epoch respectively. The batch size is 128. The final validation accuracy is shown in Table~\ref{Acc_cifar10},

\begin{table}[htbp]
\centering
\begin{tabular}{c c c c} 
\hline
\hline
\textbf{AF} & \textbf{ResNet18} & \textbf{NIN-Net} & \textbf{ResNet34} \\
\hline
 BNN baseline \cite{courbariaux2016binarized}    & 90.69 &  86.03 & 91.00\\
\hline
RSign\cite{liu2020reactnet} & 90.64 & 86.36 & 91.10\\
\hline
RPReLU \cite{liu2020reactnet} & 90.79 & 86.34 & 90.66\\
\hline
\hline
AF1 & 91.20 &  86.87  &  91.45 \\
\hline
AF2 &91.02 & 86.77 & 91.39 \\
\hline
AF3 & \textbf{91.29}  & 86.55 & 91.08 \\
\hline
AF4 & 91.16 & \textbf{87.04} &  \textbf{91.68} \\
\hline
AF5 & \textbf{91.28} &  86.36 &  91.26 \\
\hline
AF6 &  \textbf{91.32} & \textbf{87.48}  &  91.39  \\
\hline
AF7 &  90.75 & 86.09 &  90.98 \\
\hline
AF8 & 91.06 & 87.03 &  91.16 \\
\hline
AF9 & 90.93 & 86.96 &  \textbf{91.75} \\
\hline
AF10 & 91.24 & \textbf{87.42} &  91.48 \\
\hline
AF11 & \textbf{91.32} & 85.46   &  \textbf{92.08} \\
\hline
AF12 & \textbf{91.40} & \textbf{87.25}  & 91.93  \\
\hline
AF13 & 90.90   &  86.56 & \textbf{92.20}  \\
\hline
AF14 & 91.15 & \textbf{87.22}  &  \textbf{91.93} \\
\hline
AF15 & 91.14 & 86.15 &  91.58 \\
\hline
\hline
\end{tabular}
\caption{Validation Accuracy on CIFAR10.}
\label{Acc_cifar10}
\end{table}

As shown in Table~\ref{Acc_cifar10}, all the 15 AFs obtained by the GA-based search attain performance improvement over the BNN baseline on ResNet18. Specifically, the top 5 functions (AF12, AF6, AF11, AF3, AF5) improve accuracy by 0.71\%, 0.63\%, 0.63\%, 0.6\%, 0.59\%, respectively. For NIN-Net, 14 out of the 15 AFs show performance improvement. The top 5 functions (AF6, AF10, AF12, AF14, AF4) improve performance by 1.45\%, 1.39\%, 1.22\%, 1.19\%, 1.01\%. Finally, for ResNet34, 14 out of the 15 functions outperform the original BNN architecture. The top 5 functions (AF13, AF11, AF14, AF9, AF4) improve accuracy by 1.2\%, 1.08\%, 0.93\%, 0.75\%, 0.68\% respectively. The best improvement over the baseline BNN on CIFAR10 is attained by AF6 on NIN-Net, showcasing 1.45\% accuracy enhancement. Overall for CIFAR10, most of the GA-searched AFs (except AF3, AF7, AF11 and AF15) match or outperform the baseline BNN, RSign and RPReLU on all of the three network models (ResNet18, NIN-Net and ResNet34). We did not observe significant performance improvement using \emph{RSign} and \emph{RPReLU}.

\vspace{4pt}\noindent\textbf{ImageNet:} For ImageNet, two different network models are utilized: \emph{ResNet18} and \emph{NIN-E}. NIN-E is an enhanced implementation of NIN-Net proposed in \cite{tang2017train}, which enlarges the kernel size of the first four $mlpconv$ layers from $1\times 1$ to $3\times 3$, and increases the number of output channels of the first two $mlpconv$ layers from 96 to 128.

We use the standard pre-processing technique: all images are resized to $256 \times 256$ and randomly cropped to $224 \times 224$ for training. The validation dataset uses center crop only. We set the initial learning rate as 5e-3, which is gradually decreased during training by a factor of 5 at the 25th, 35th, 40th, and 45th epoch. Each model is trained over 50 epochs. The batch size is 64. The validation accuracy is listed in Table~\ref{Acc_imagenet}.

\begin{table}[htbp]
\centering
\begin{tabular}{c c c c c} 
\hline
\hline
\textbf{AF} & \textbf{ResNet18 (Top-1/5)} & \textbf{NIN-E (Top-1/5)} \\
\hline
 BNN baseline \cite{courbariaux2016binarized}     & 54.284 / 77.670  &  49.730 / 74.144 \\
\hline
RSign \cite{liu2020reactnet} & 54.292 / 77.618 & 49.980 / 74.514 \\
\hline
RPReLU \cite{liu2020reactnet} & 53.628 / 76.850 & 50.266 / 74.252 \\
\hline
\hline
AF1 & \textbf{55.398} / 78.558 & \textbf{51.738} / 75.290 \\
\hline
AF2 & \textbf{55.268} / 78.394 & 51.510 / 75.320 \\
\hline
AF3 & \textbf{56.066} / 79.056  & \textbf{51.680} / 75.572 \\
\hline
AF4 & 55.064 / 78.106 & \textbf{52.232} / 76.186 \\
\hline
AF5 & \textbf{55.712} / 78.454 & 50.210 / 74.494 \\
\hline
AF6 & 54.354 / 77.636 & \textbf{51.862} / 75.518 \\
\hline
AF7 & 54.624 / 77.752 & 50.502 / 74.514 \\
\hline
AF8 & 54.194 / 77.640 & 49.918 / 74.272 \\
\hline
AF9 & 53.414 / 76.960 & \textbf{52.270} / 75.822 \\
\hline
AF10 & 53.004 / 76.454 & 50.834 / 75.216 \\
\hline
AF11 & 54.026 / 77.386 & 50.722 / 74.768 \\
\hline
AF12 & \textbf{55.514} / 78.556 & 51.558 / 75.308 \\
\hline
AF13 & 54.390 / 77.728 & 50.714 / 74.742 \\
\hline
AF14 & 54.488 / 77.976 & 50.324 / 74.466 \\
\hline
AF15 & 55.092 / 78.246 & 50.374 / 74.500 \\
\hline
\hline
\end{tabular}
\caption{Validation Accuracy on ImageNet.}
\label{Acc_imagenet}
\end{table}

For ResNet18, 11 out of the 15 discovered AFs improve the top-1 accuracy. The best behaving functions (AF3, AF5, AF12, AF1, AF2) harvest 1.782\%, 1.428\%, 1.23\%, 1.114\%, 0.984\% performance improvement respectively, over the BNN baseline.
For NIN-E, all the 15 AFs improve the validation accuracy. The promising functions (AF9, AF4, AF6, AF1, AF3) improve the top-1 accuracy by 2.54\%, 2.502\%, 2.132\%, 2.008\%, 1.95\%, respectively. The most significant improvement is attained by AF9 on NIN-E, showcasing as much as 2.54\% accuracy improvement over the BNN baseline.

Overall, except AF5, AF8, AF9, AF10 and AF11, all the remaining AFs (10 out of 15) outperform the BNN baseline, RSign, and RPReLU approaches. Again, we did not observe performance improvements by using RSign and RPReLU on ResNet18. For NIN-E, RSign and RPReLU bring nearly 0.25\% and 0.54\% top-1 accuracy improvement.

\section{Discussion}
\label{s:Discussion}
\noindent
Based on our evaluation with the three distinct network models on CIFAR10 and ImageNet, seven (i.e., $AF1$, $AF2$, $AF4$, $AF6$, $AF12$, $AF13$, $AF14$) out of the 15 investigated activation functions stand out showing performance improvement and successfully generalize such improvement to a different dataset. These seven searched activation functions are listed below.
Compared to STE, the gradients of these functions are much smoother, rendering a more stable training process, which partially explains the better performance.

\begin{figure*}[!htb]
\minipage{0.28\columnwidth}
\includegraphics[width=1\columnwidth]{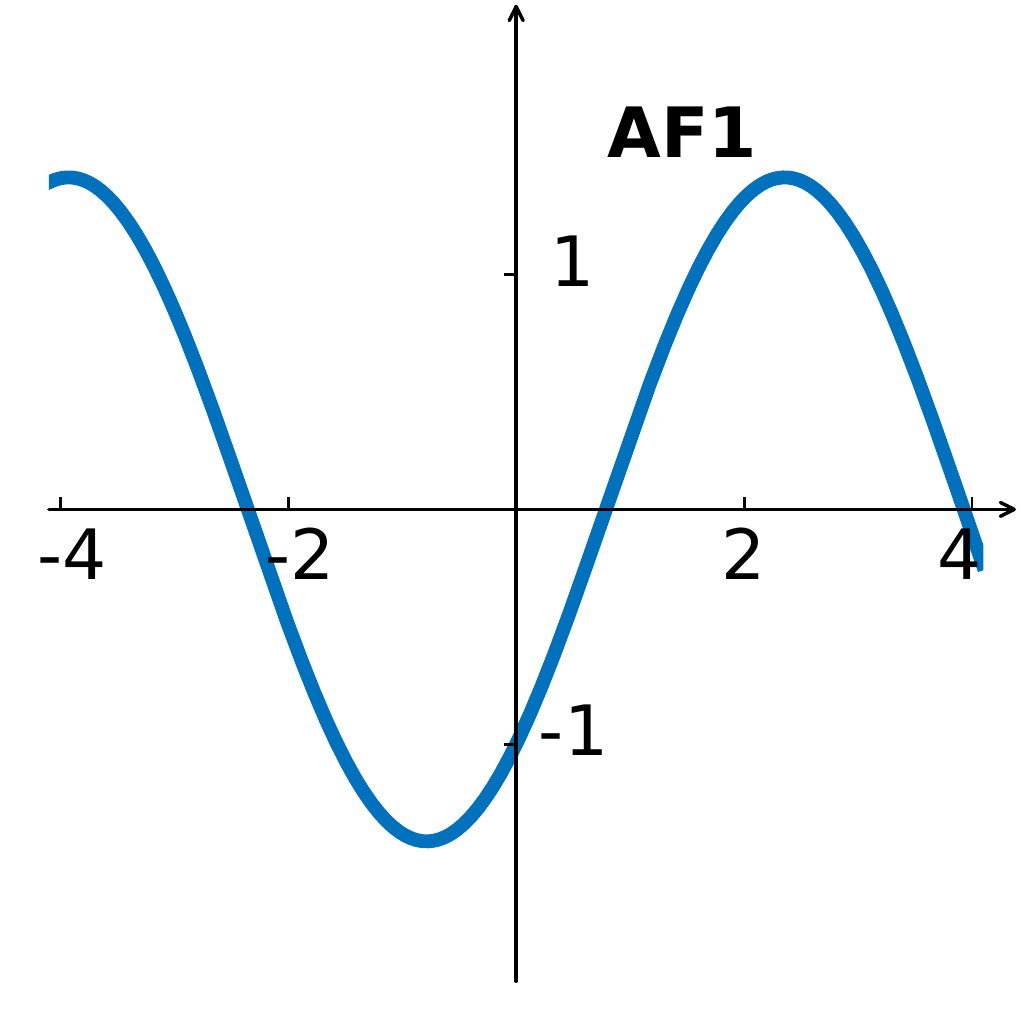} 
\label{fig:F1}
\endminipage\hfill
\minipage{0.28\columnwidth}
\includegraphics[width=1\columnwidth]{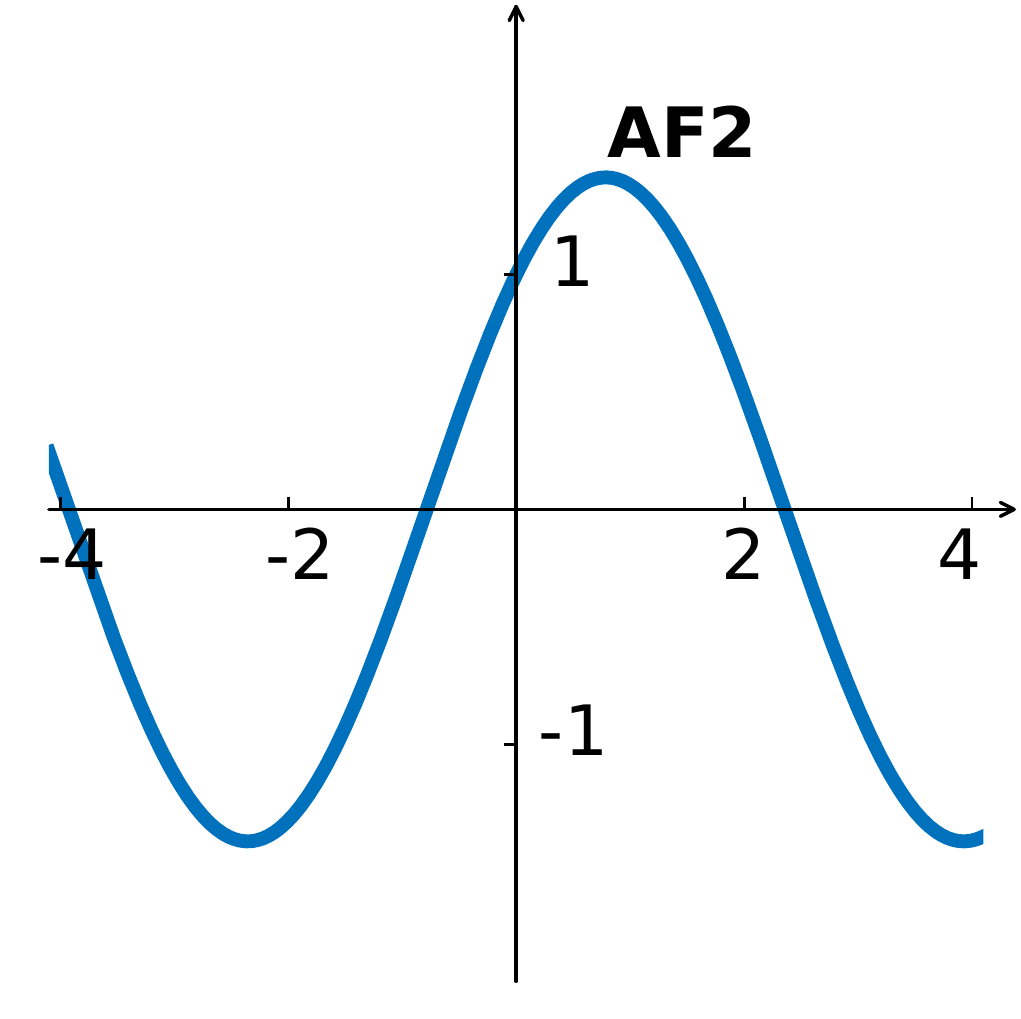} 
\label{fig:F2}
\endminipage\hfill
\minipage{0.28\columnwidth}
\includegraphics[width=1\columnwidth]{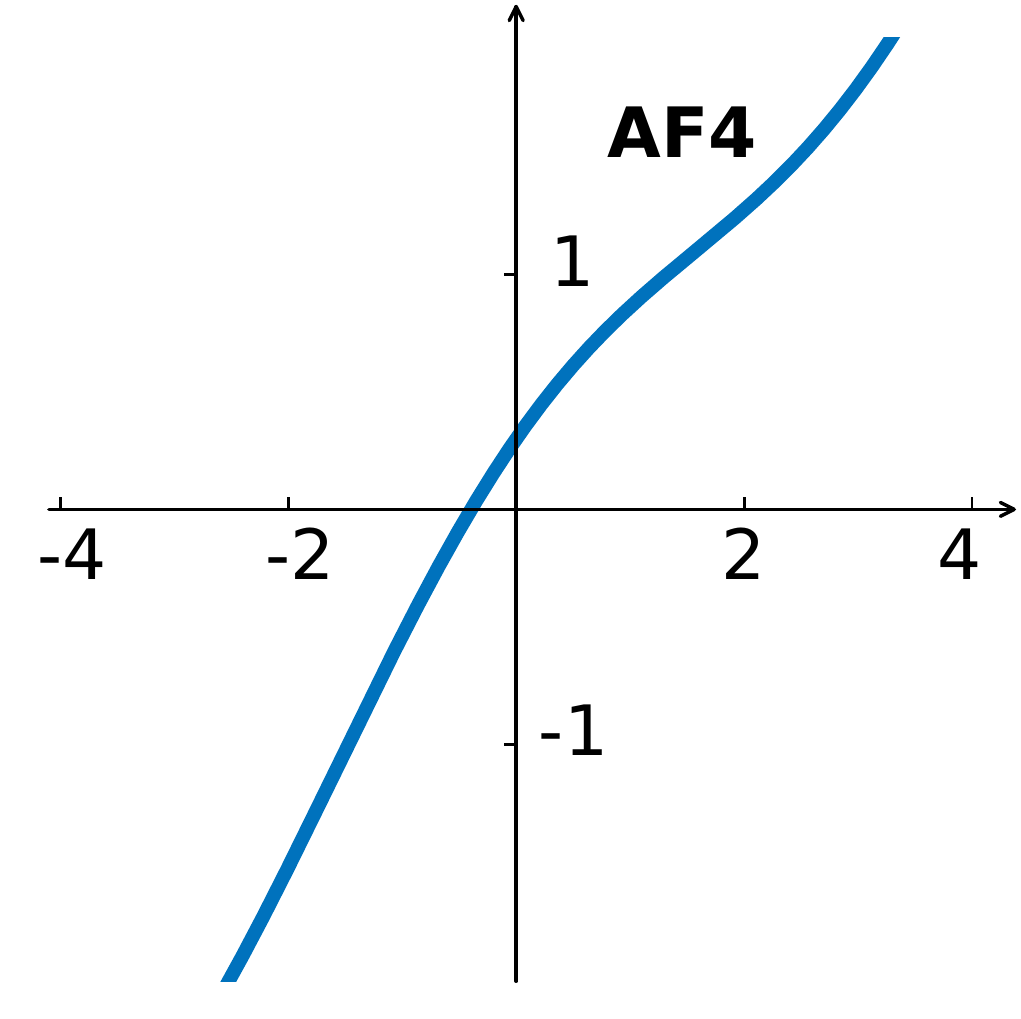} 
\label{fig:F3}
\endminipage\hfill
\minipage{0.28\columnwidth}
\includegraphics[width=1\columnwidth]{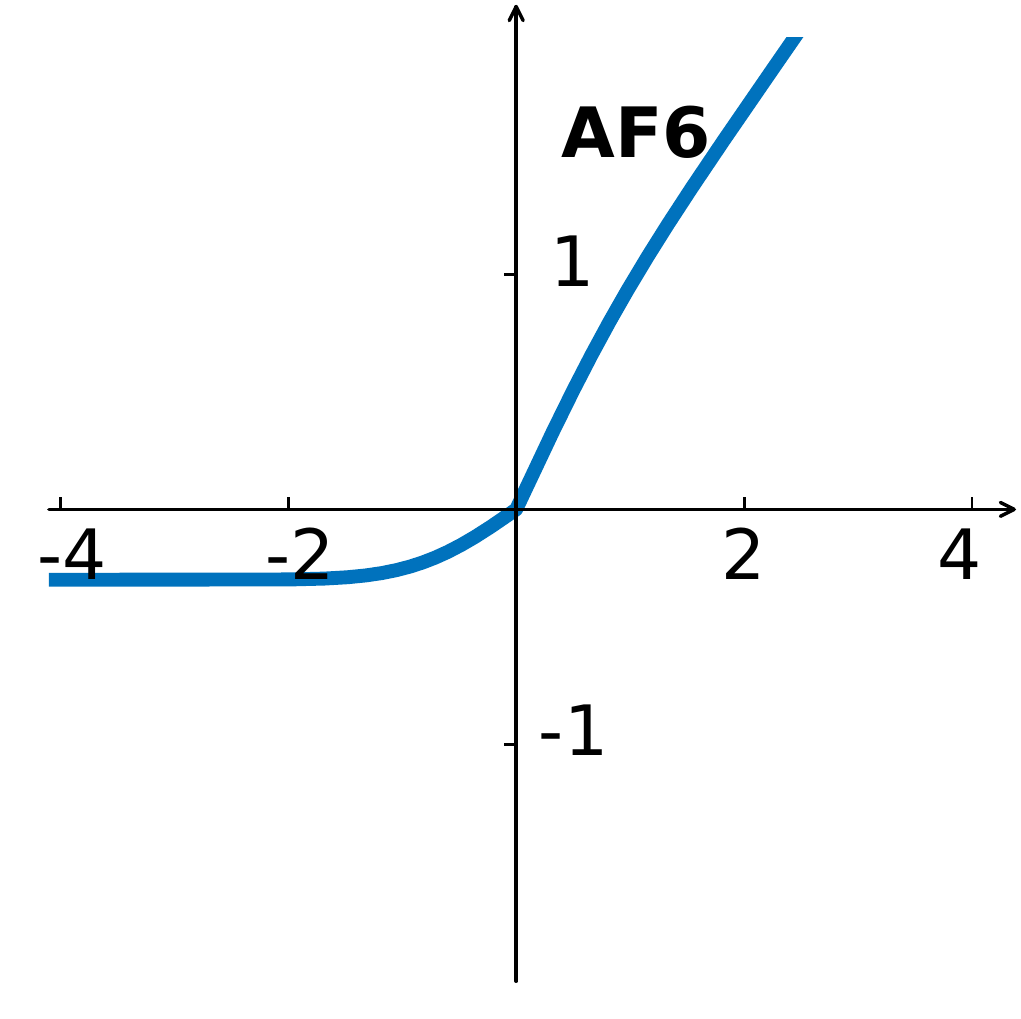} 
\label{fig:F4}
\endminipage\hfill
\minipage{0.28\columnwidth}
\includegraphics[width=1\columnwidth]{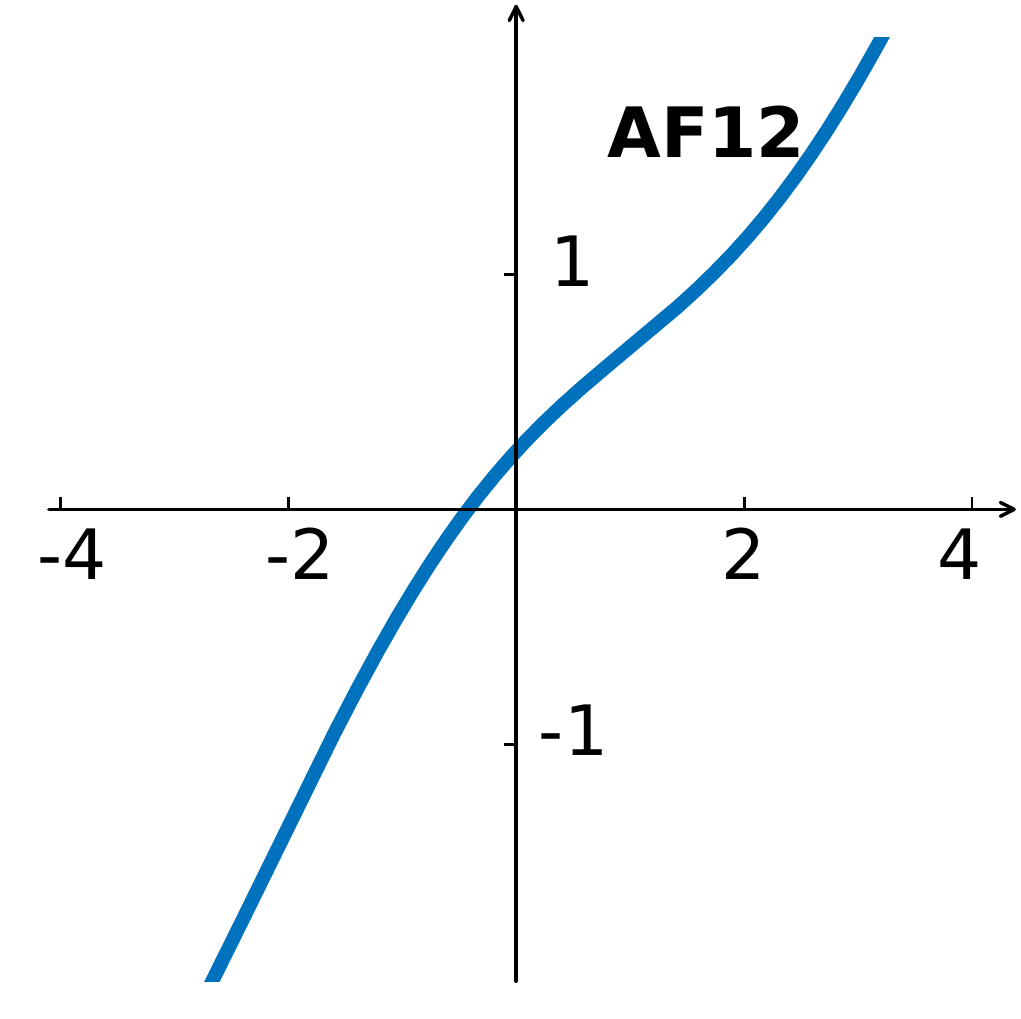} 
\label{fig:F5}
\endminipage\hfill
\minipage{0.28\columnwidth}
\includegraphics[width=1\columnwidth]{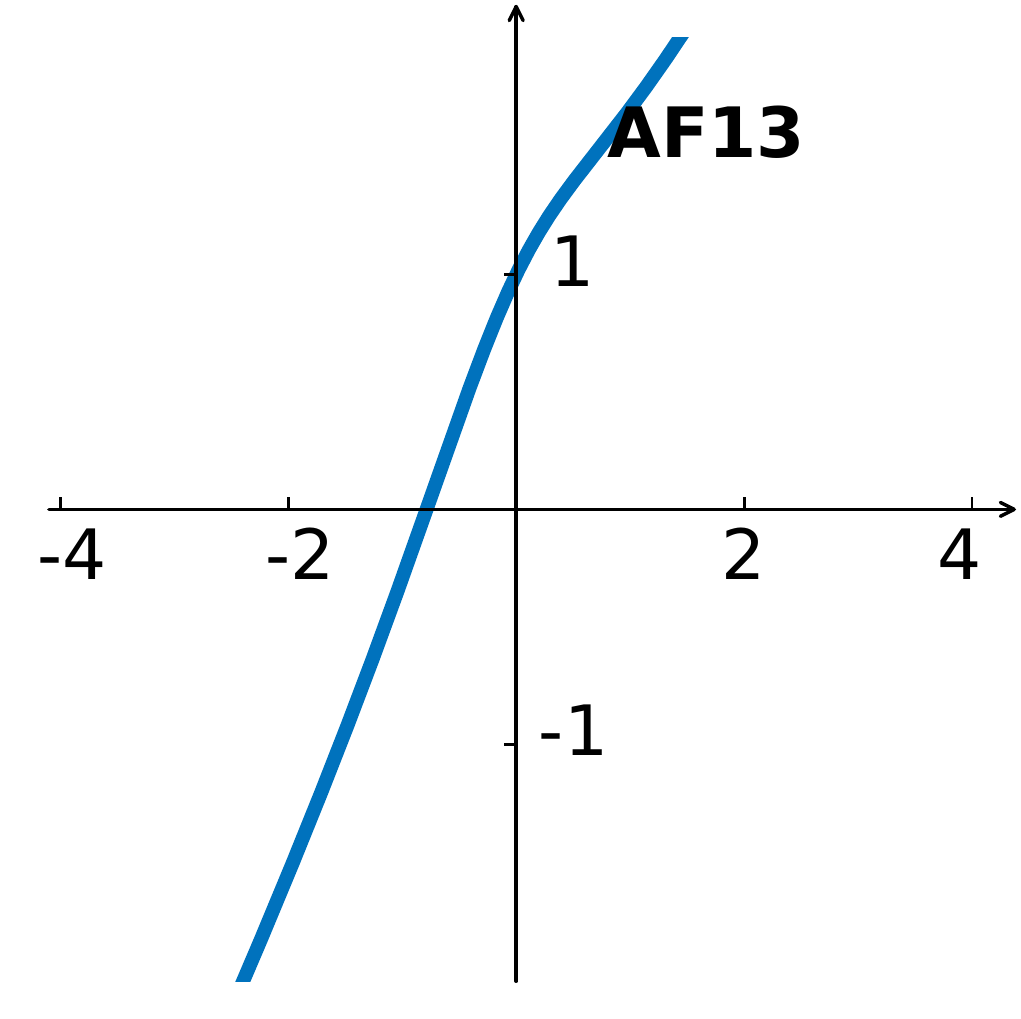} 
\label{fig:F6}
\endminipage\hfill
\minipage{0.28\columnwidth}
\includegraphics[width=1\columnwidth]{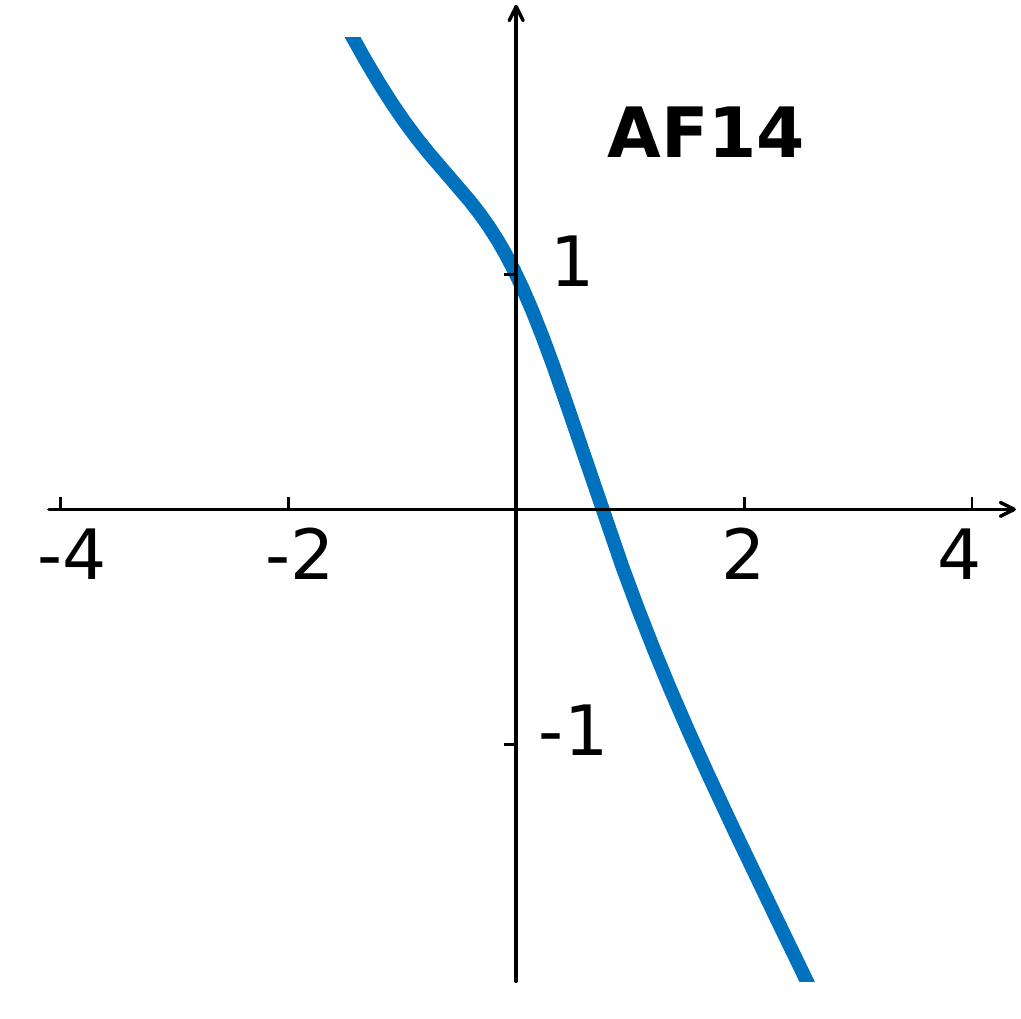} 
\label{fig:F7}
\endminipage\hfill
\caption{Discovered promising Activation Functions}
\label{fig:AFun}
\end{figure*}

\begin{figure*}[!htb]
\minipage{0.28\columnwidth}
\includegraphics[width=1\columnwidth]{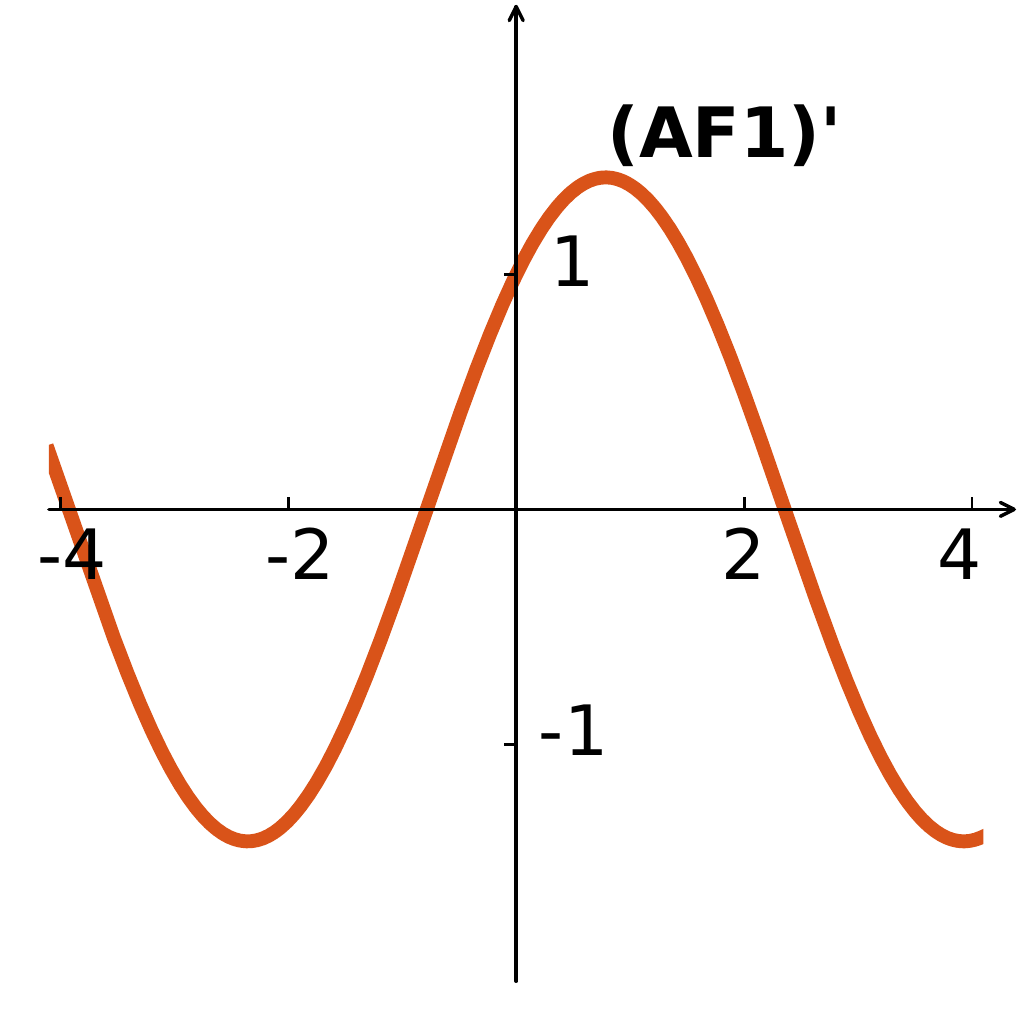} 
\label{fig:F1}
\endminipage\hfill
\minipage{0.28\columnwidth}
\includegraphics[width=1\columnwidth]{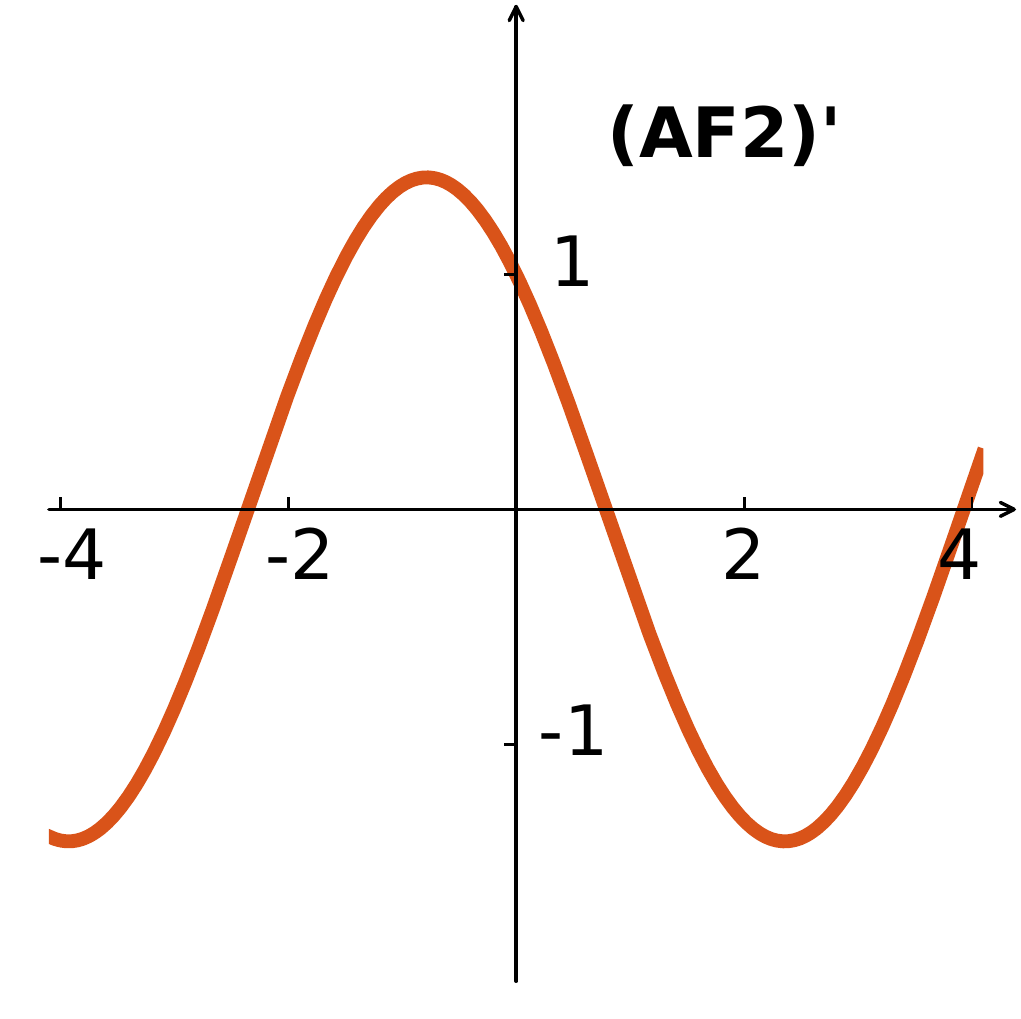} 
\label{fig:F2}
\endminipage\hfill
\minipage{0.28\columnwidth}
\includegraphics[width=1\columnwidth]{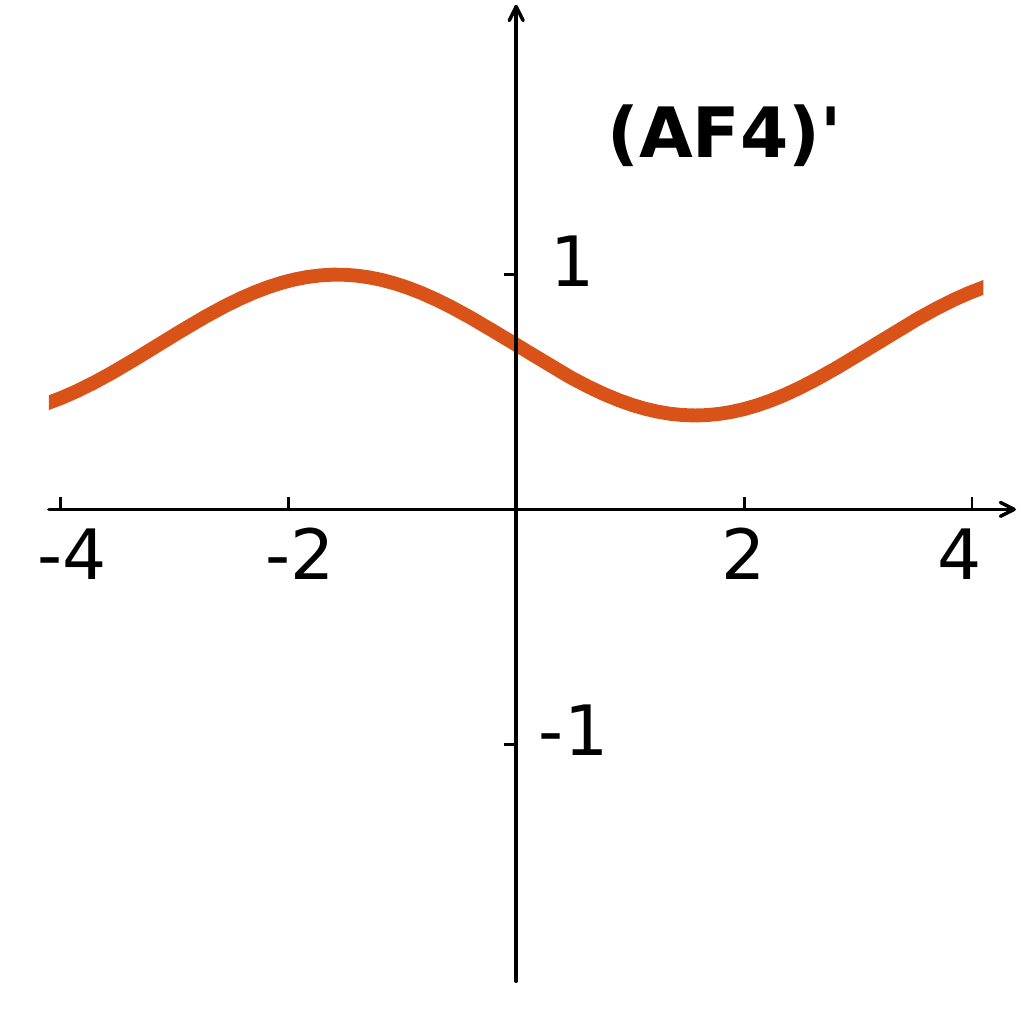} 
\label{fig:F3}
\endminipage\hfill
\minipage{0.28\columnwidth}
\includegraphics[width=1\columnwidth]{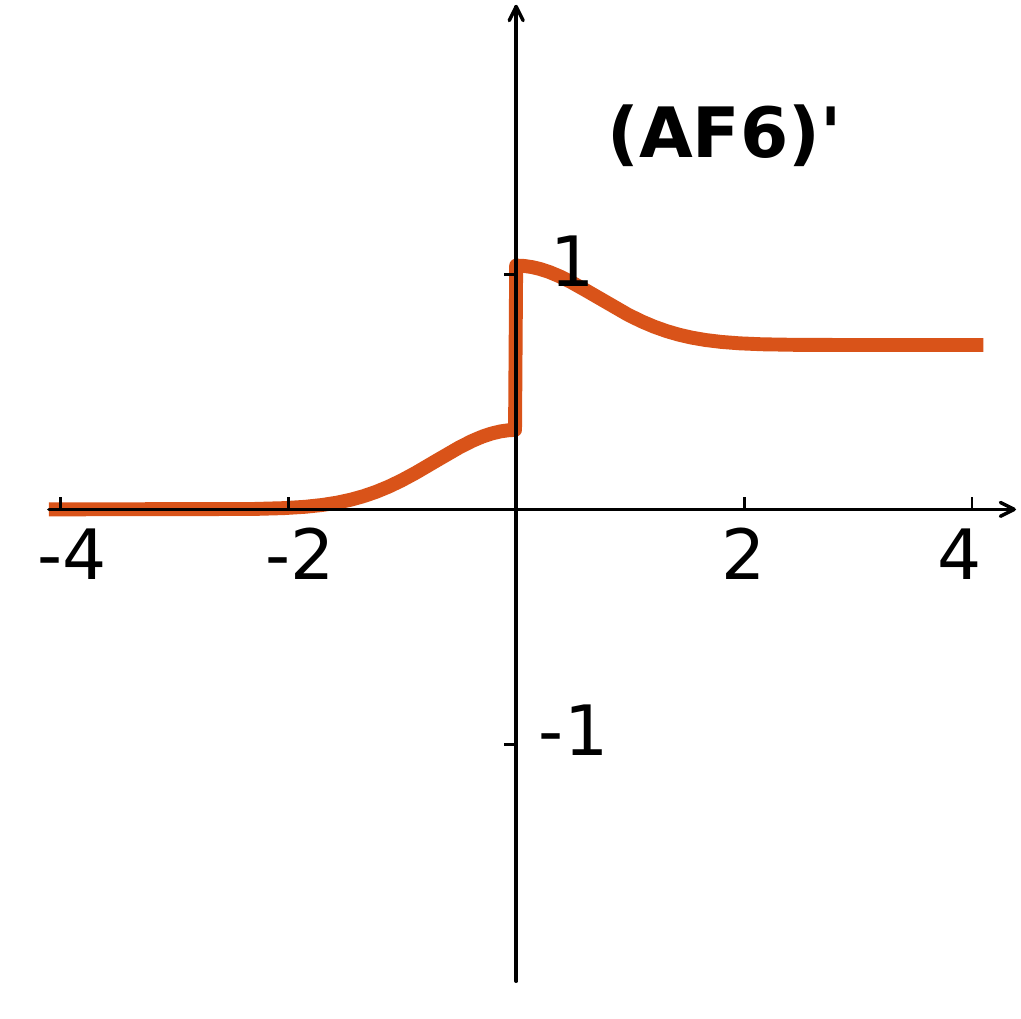} 
\label{fig:F4}
\endminipage\hfill
\minipage{0.28\columnwidth}
\includegraphics[width=1\columnwidth]{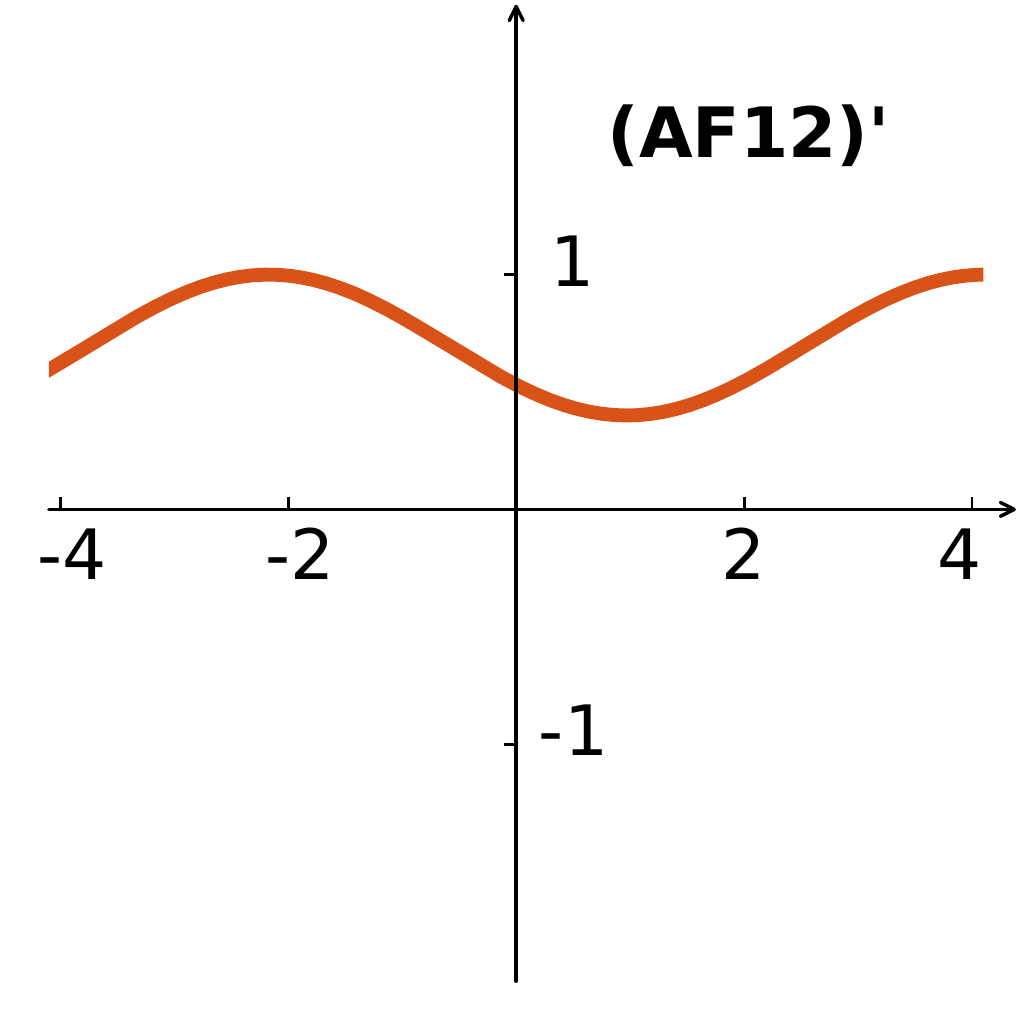} 
\label{fig:F5}
\endminipage\hfill
\minipage{0.28\columnwidth}
\includegraphics[width=1\columnwidth]{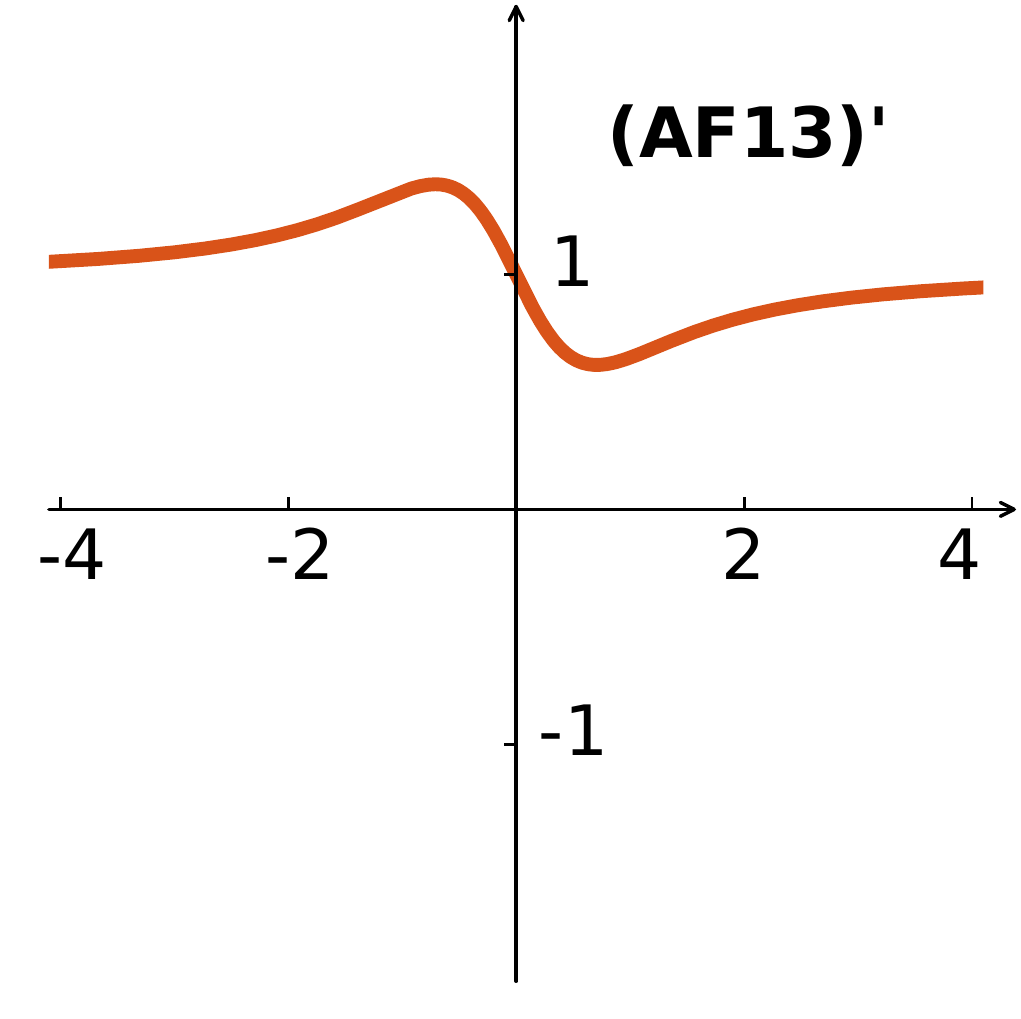} 
\label{fig:F6}
\endminipage\hfill
\minipage{0.28\columnwidth}
\includegraphics[width=1\columnwidth]{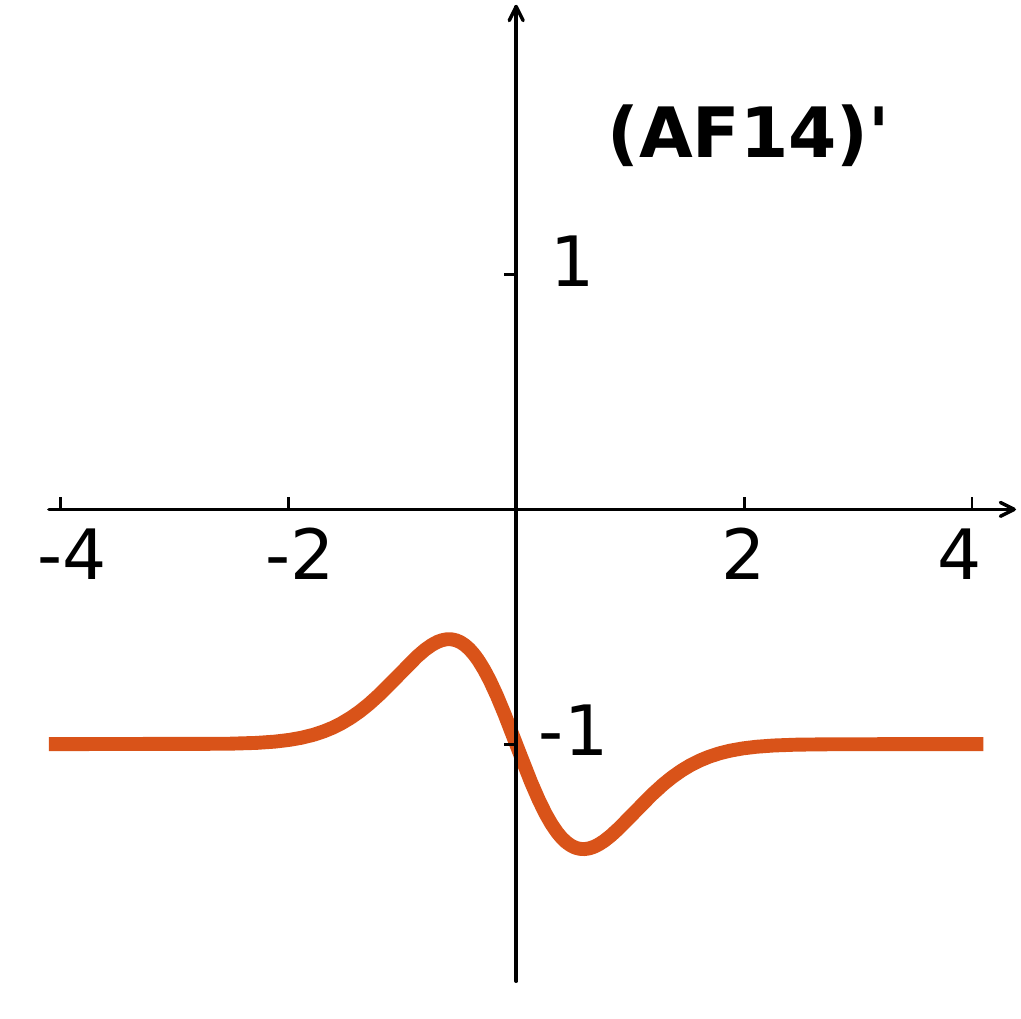} 
\label{fig:F7}
\endminipage\hfill
\caption{The gradient of discovered promising Activation Functions}
\label{fig: grad of AF}
\end{figure*}

\begin{figure*}[!htb]
\minipage{0.28\columnwidth}
\includegraphics[width=1\columnwidth]{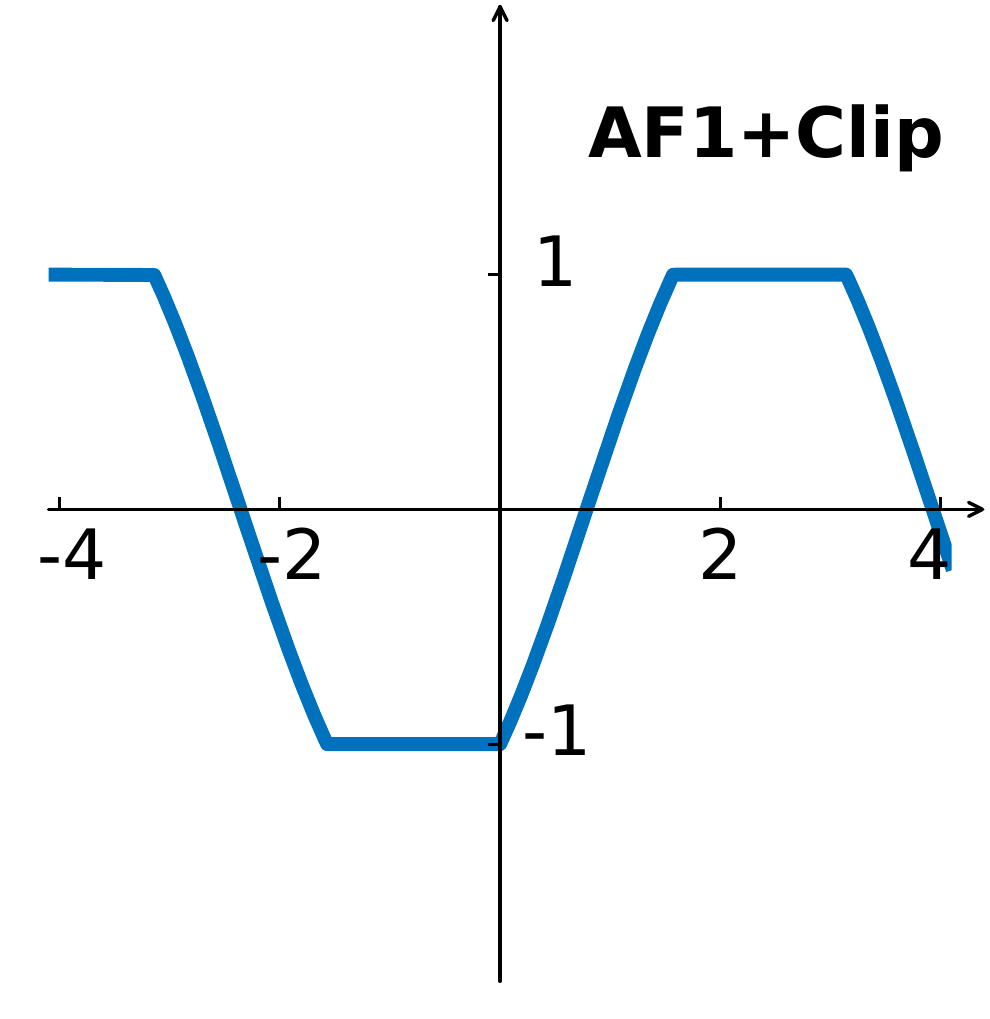} 
\label{fig:F1}
\endminipage\hfill
\minipage{0.28\columnwidth}
\includegraphics[width=1\columnwidth]{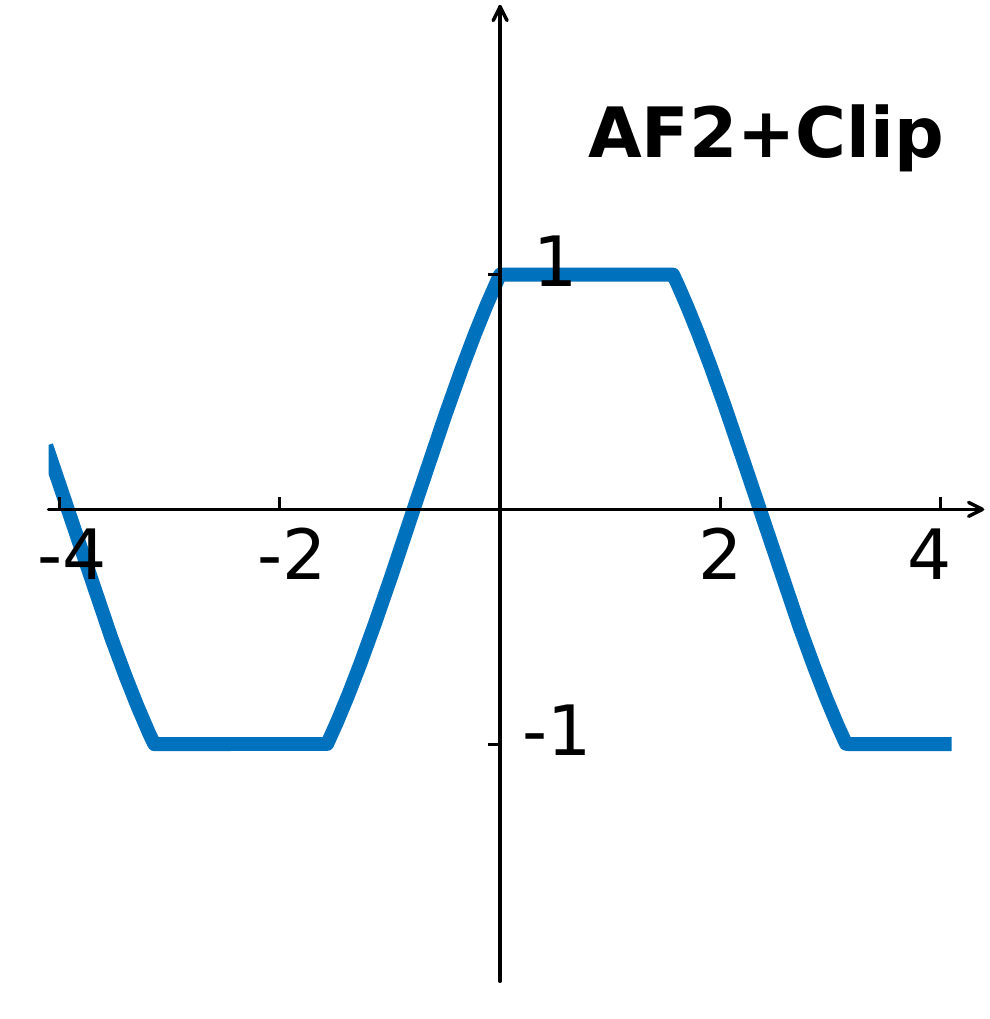} 
\label{fig:F2}
\endminipage\hfill
\minipage{0.28\columnwidth}
\includegraphics[width=1\columnwidth]{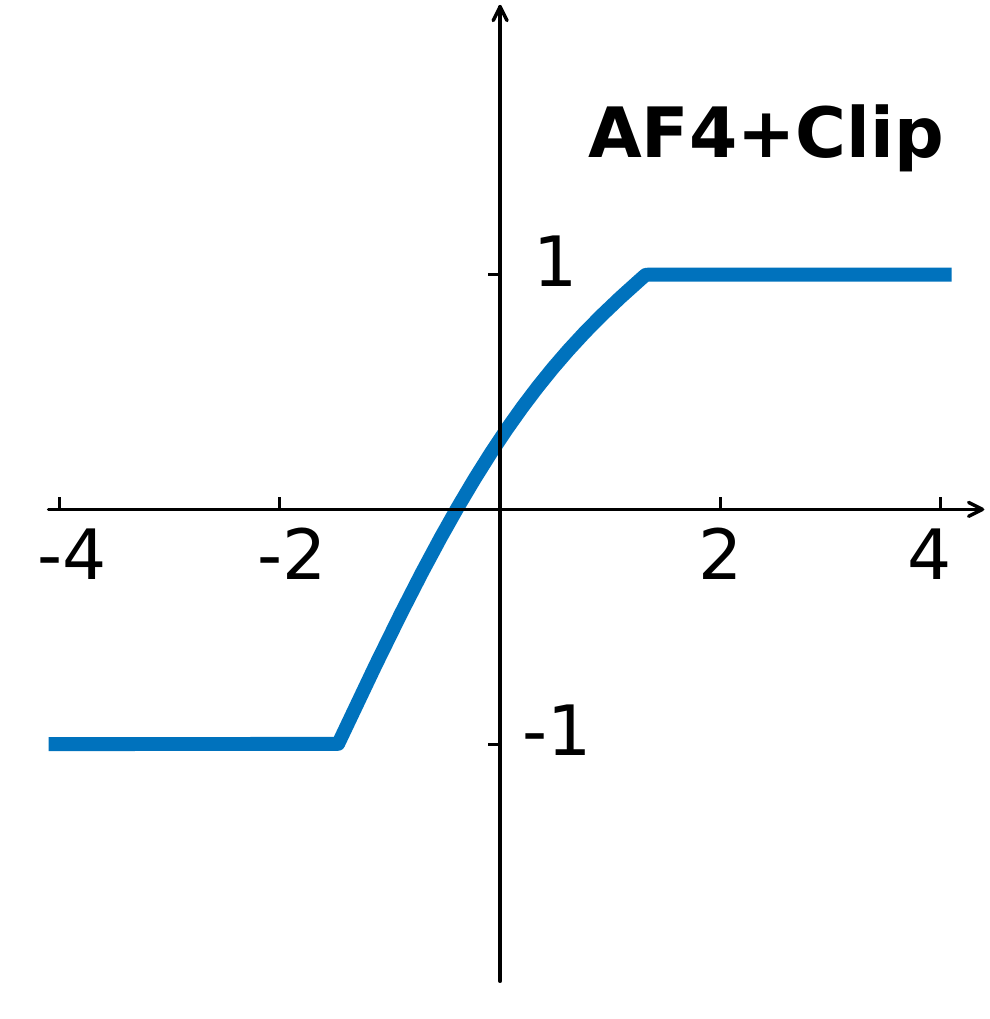} 
\label{fig:F3}
\endminipage\hfill
\minipage{0.28\columnwidth}
\includegraphics[width=1\columnwidth]{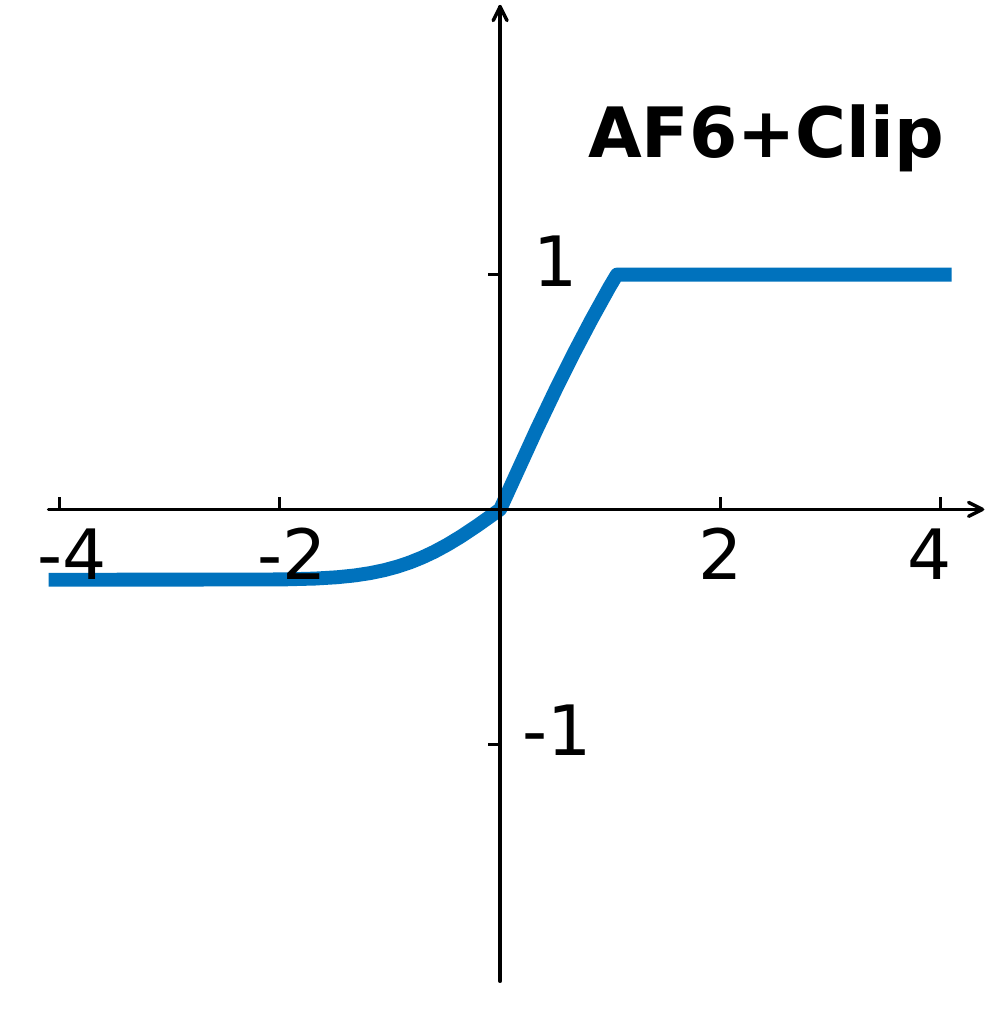} 
\label{fig:F4}
\endminipage\hfill
\minipage{0.28\columnwidth}
\includegraphics[width=1\columnwidth]{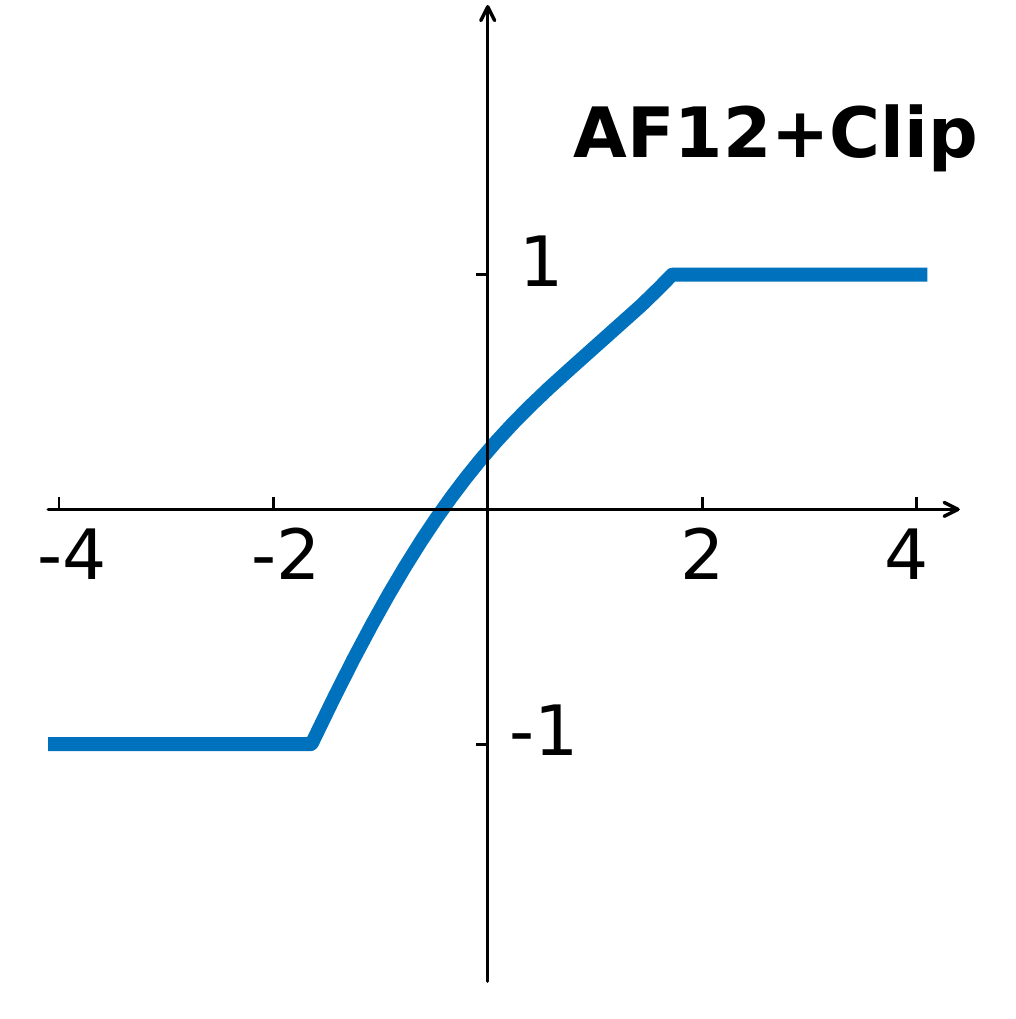} 
\label{fig:F5}
\endminipage\hfill
\minipage{0.28\columnwidth}
\includegraphics[width=1\columnwidth]{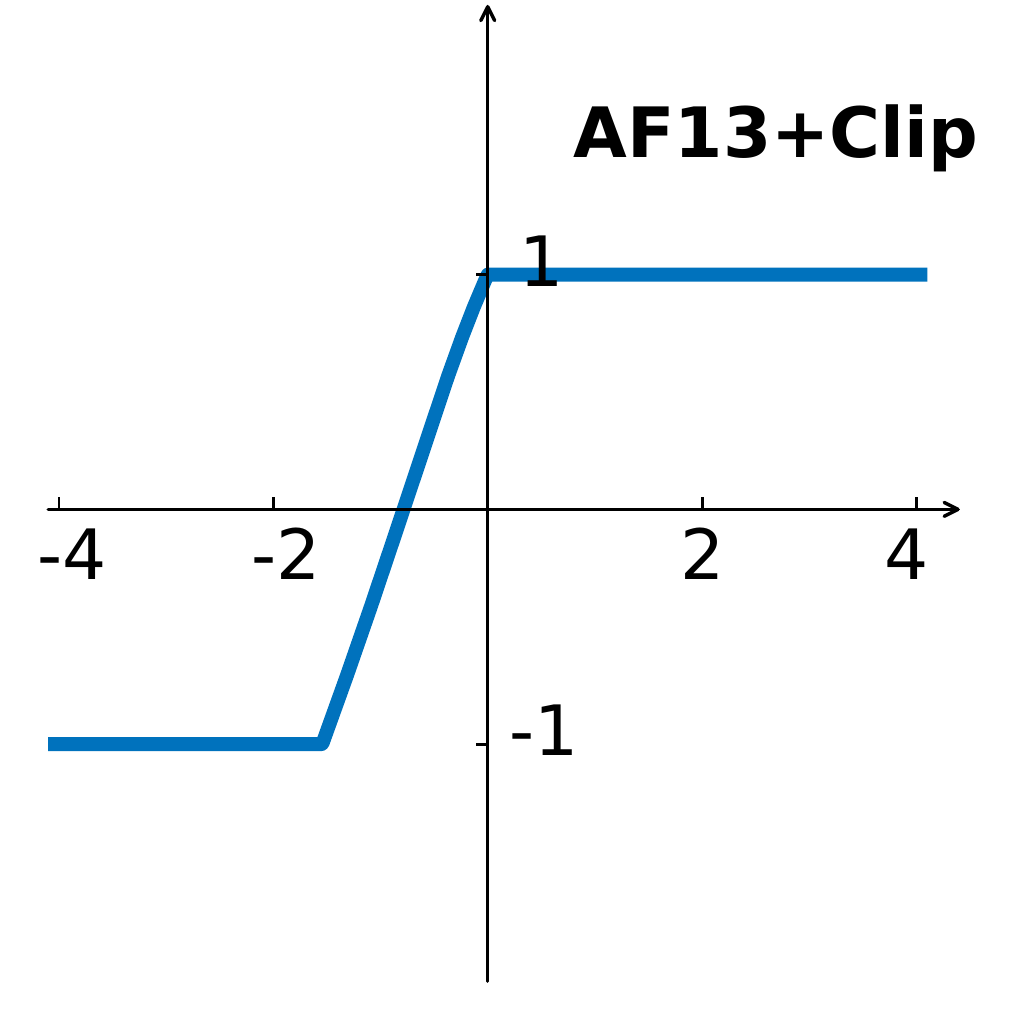} 
\label{fig:F6}
\endminipage\hfill
\minipage{0.28\columnwidth}
\includegraphics[width=1\columnwidth]{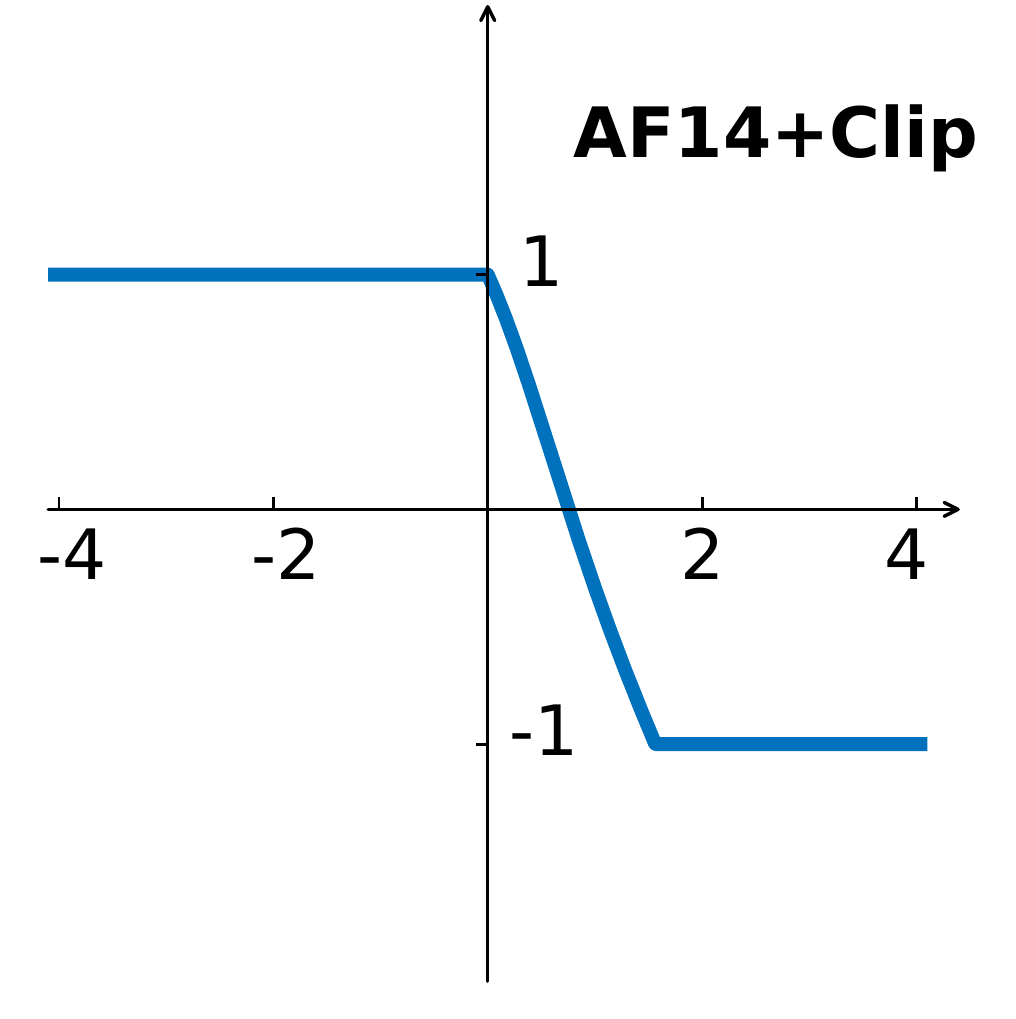} 
\label{fig:F7}
\endminipage\hfill
\caption{Discovered promising Activation Functions + Clip}
\label{fig:AFClip}
\end{figure*}

\begin{figure*}[!htb]
\minipage{0.28\columnwidth}
\includegraphics[width=1\columnwidth]{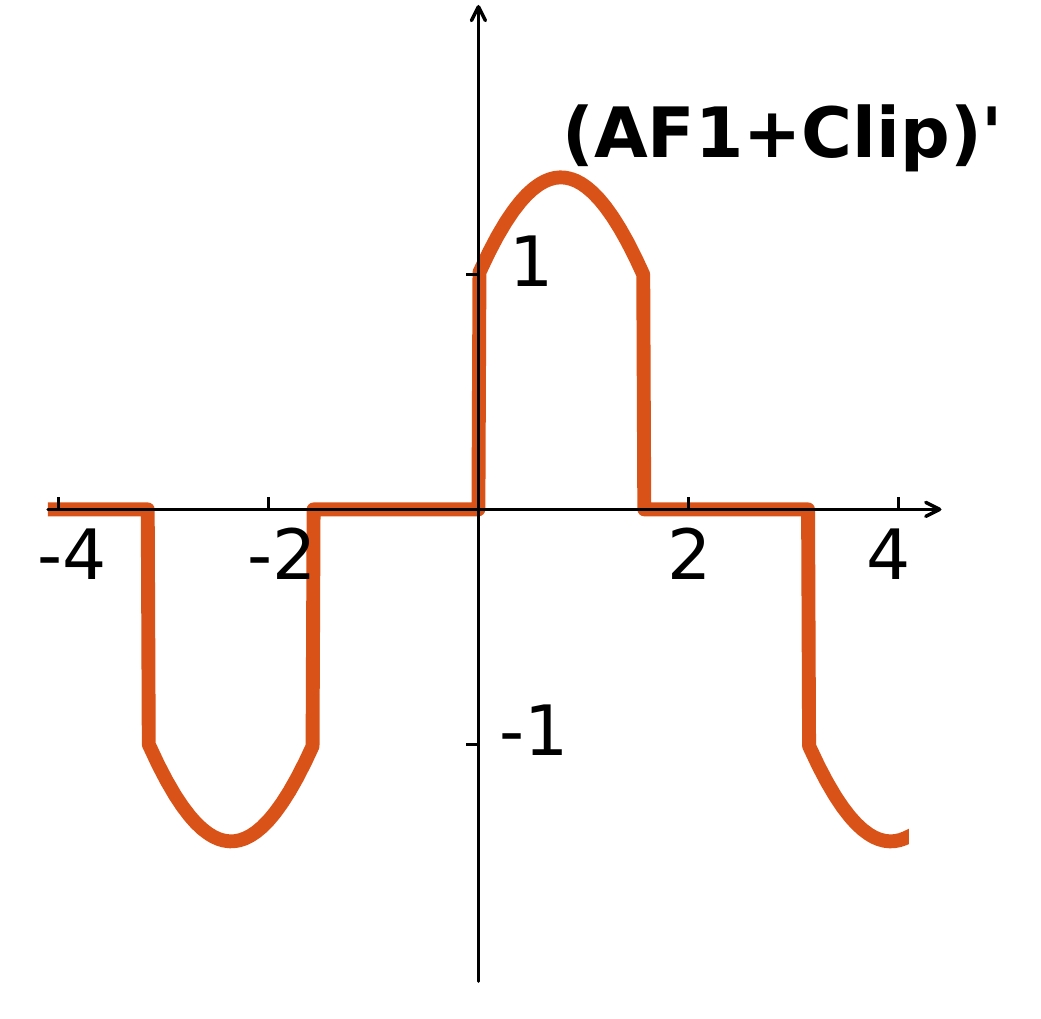} 
\label{fig:F1}
\endminipage\hfill
\minipage{0.28\columnwidth}
\includegraphics[width=1\columnwidth]{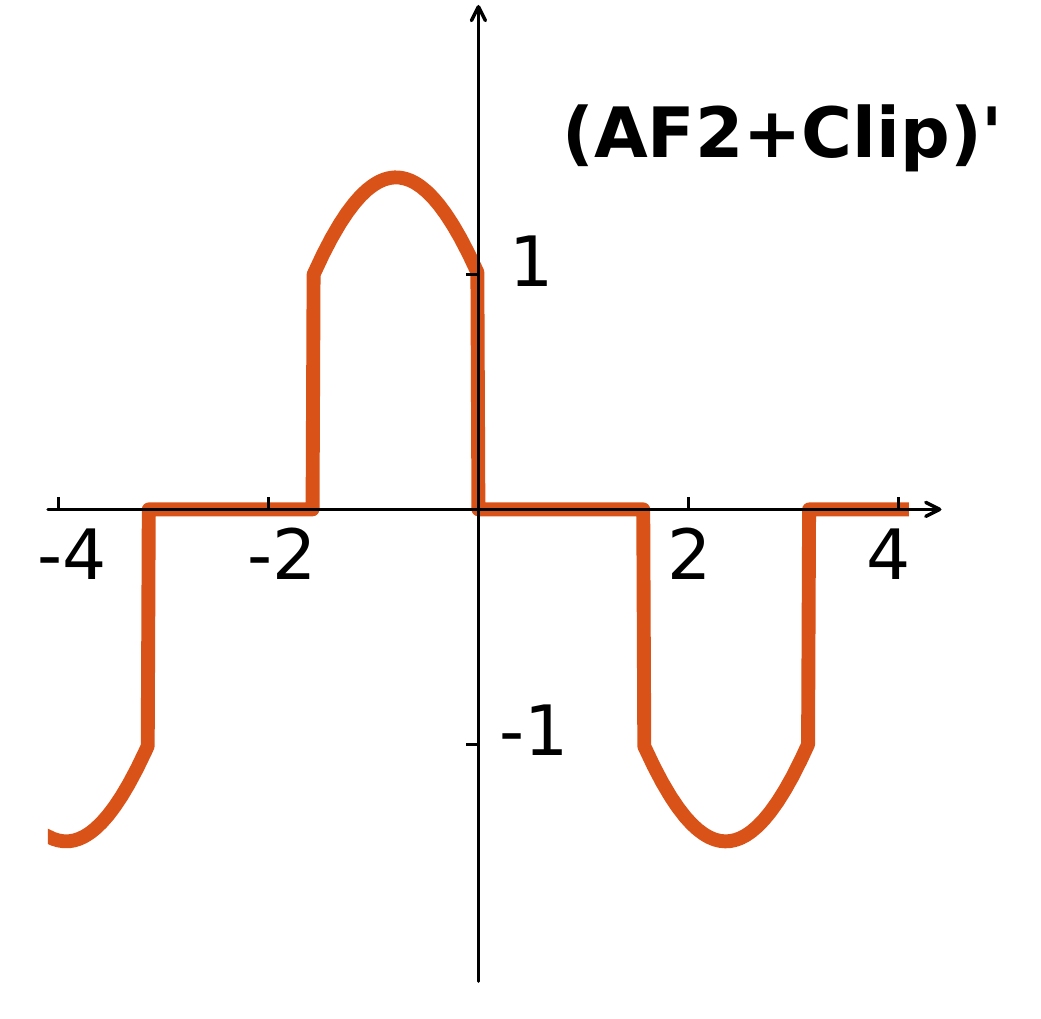} 
\label{fig:F2}
\endminipage\hfill
\minipage{0.28\columnwidth}
\includegraphics[width=1\columnwidth]{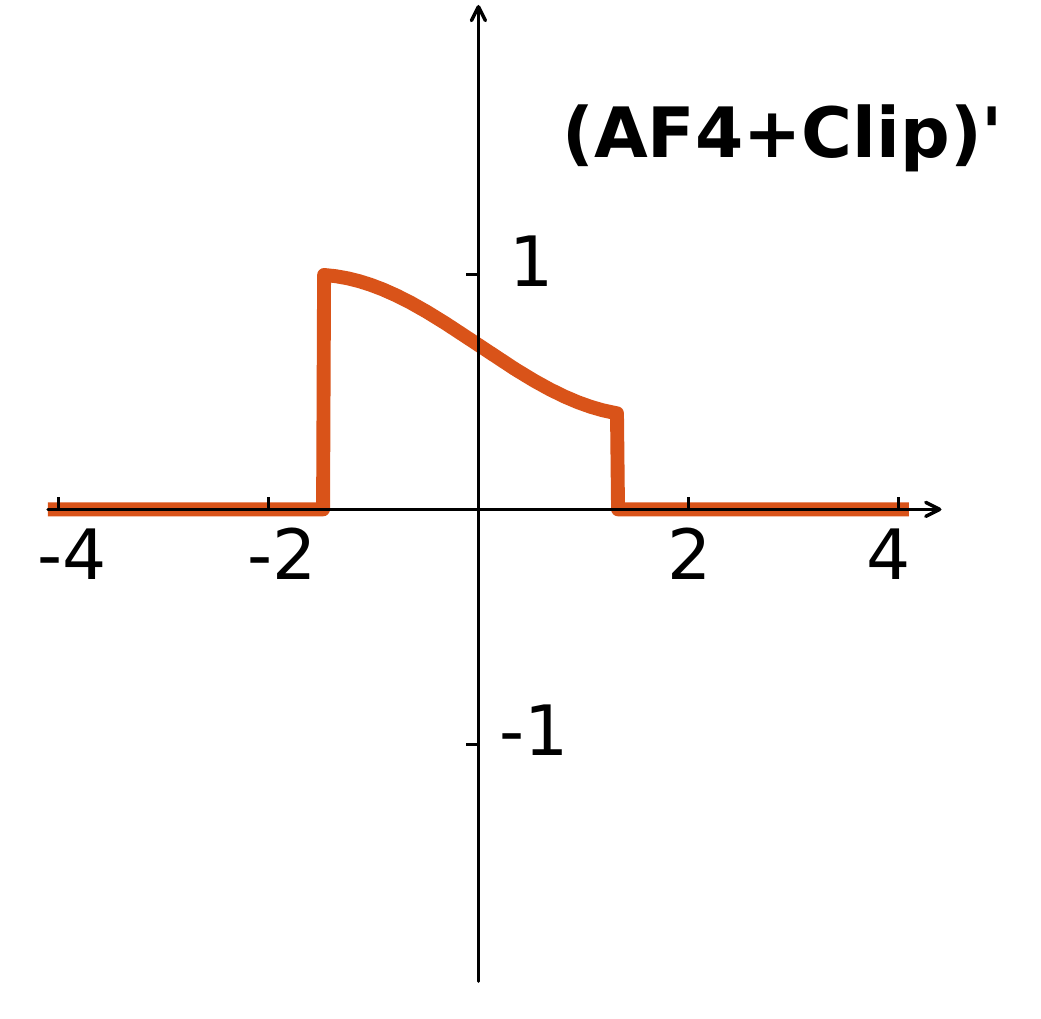} 
\label{fig:F3}
\endminipage\hfill
\minipage{0.28\columnwidth}
\includegraphics[width=1\columnwidth]{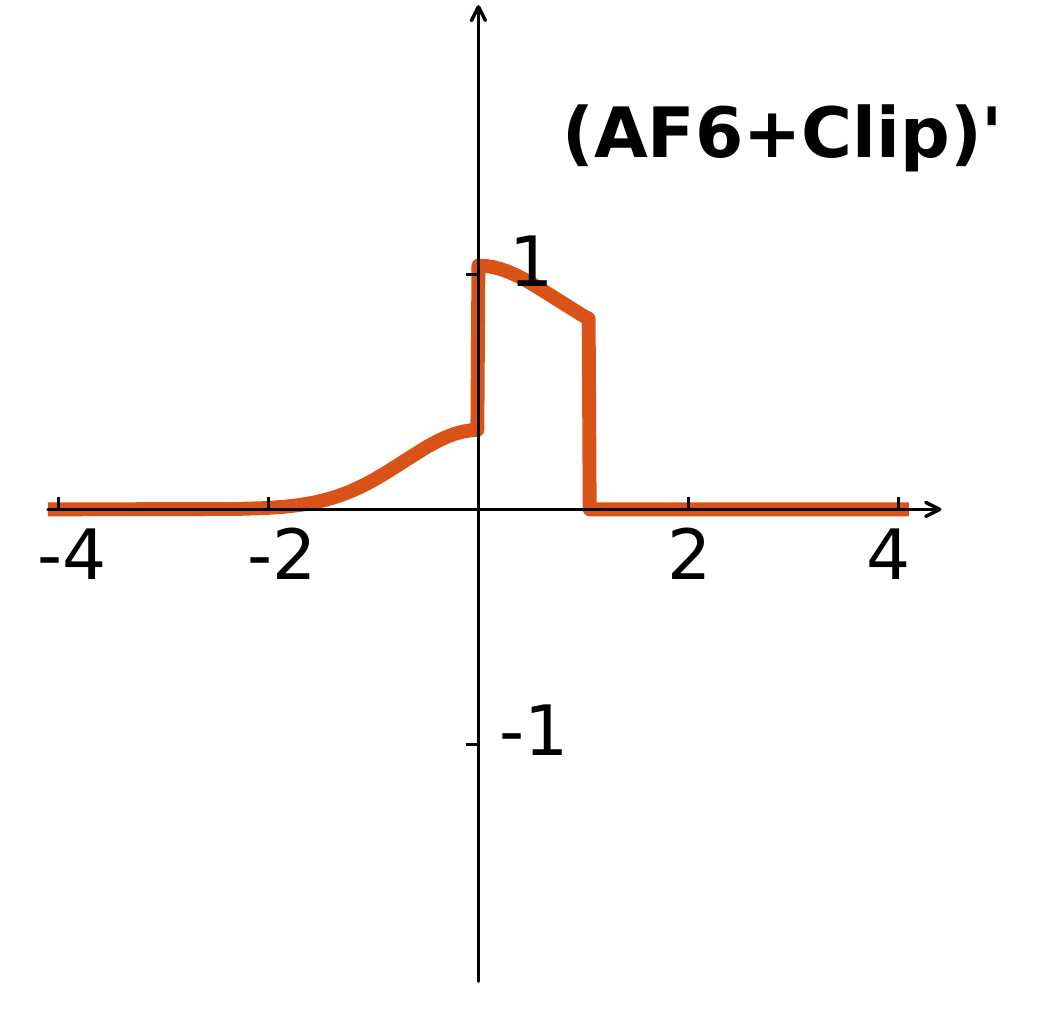} 
\label{fig:F4}
\endminipage\hfill
\minipage{0.28\columnwidth}
\includegraphics[width=1\columnwidth]{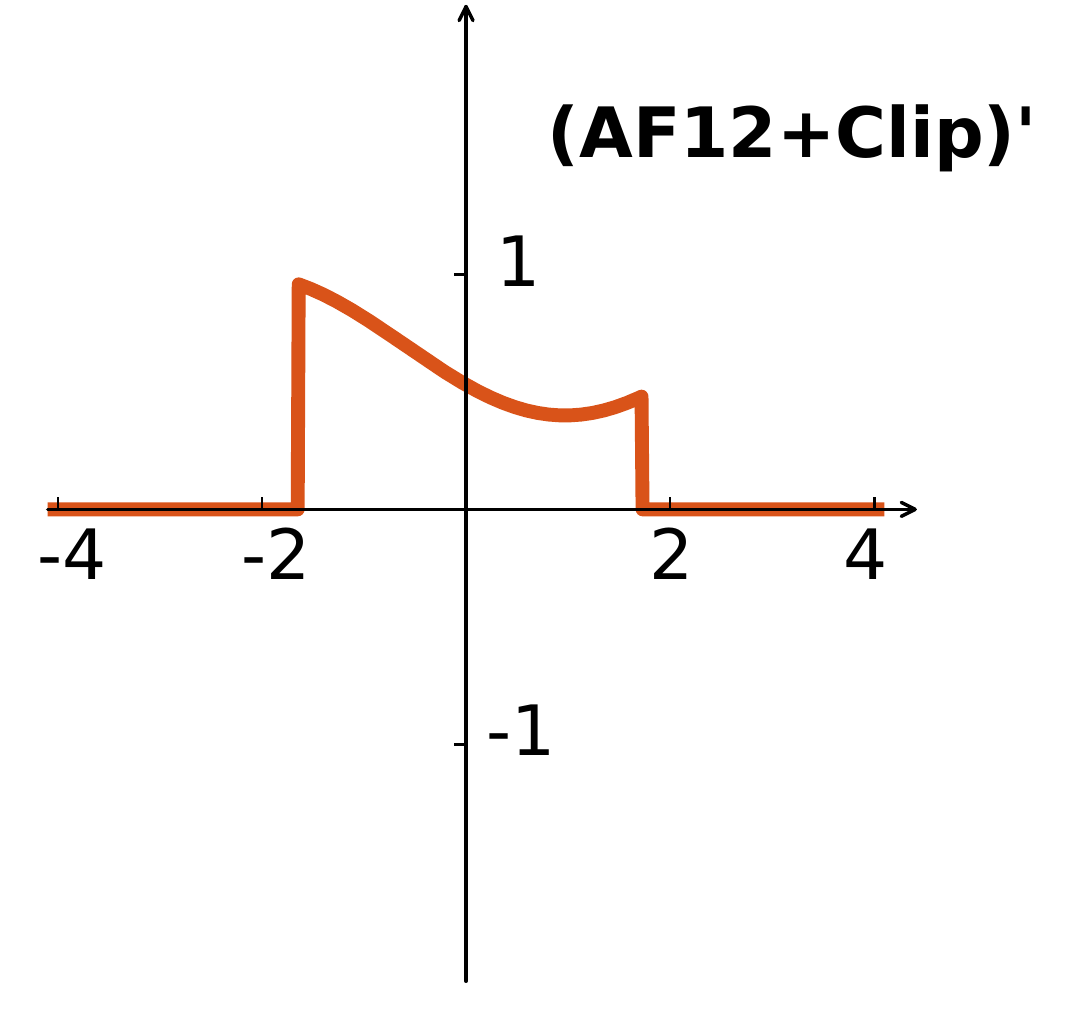} 
\label{fig:F5}
\endminipage\hfill
\minipage{0.28\columnwidth}
\includegraphics[width=1\columnwidth]{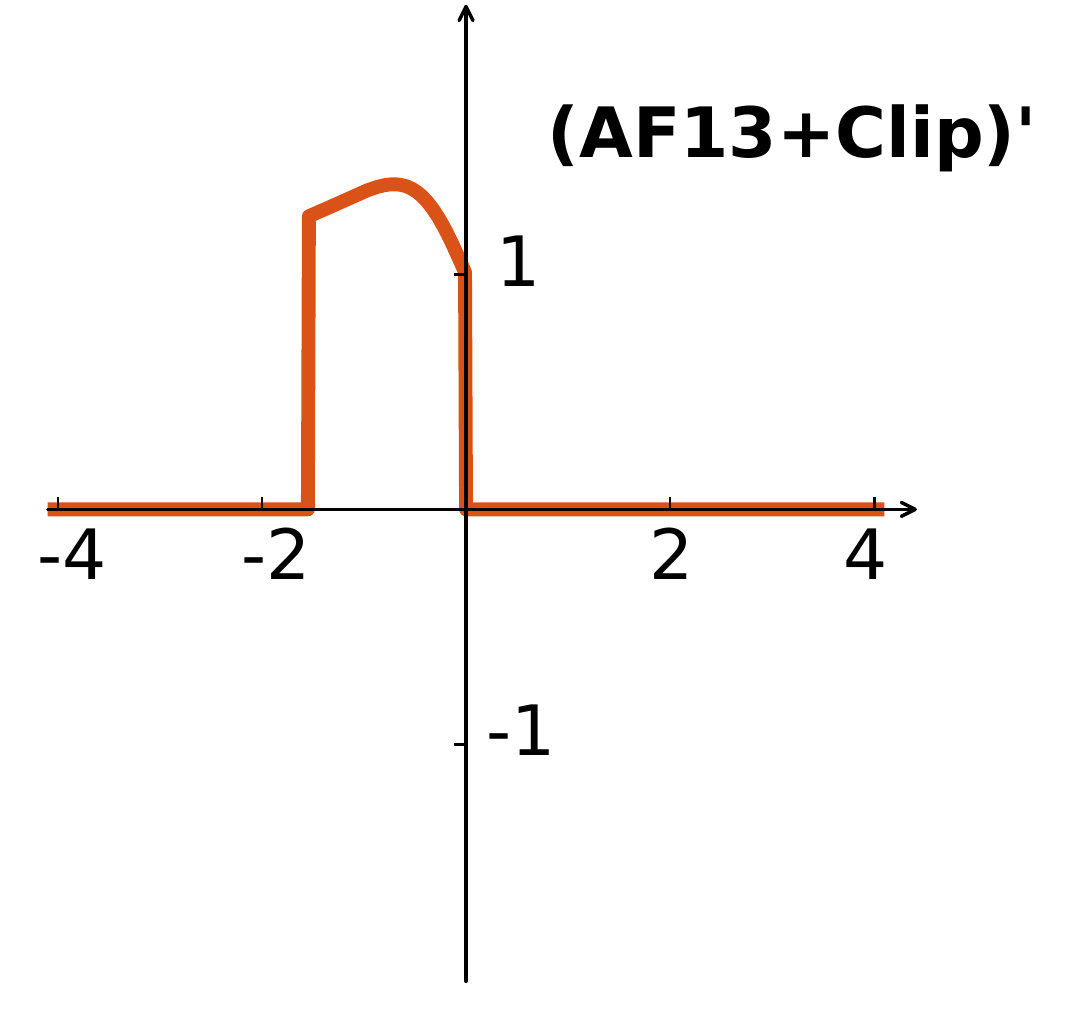} 
\label{fig:F6}
\endminipage\hfill
\minipage{0.28\columnwidth}
\includegraphics[width=1\columnwidth]{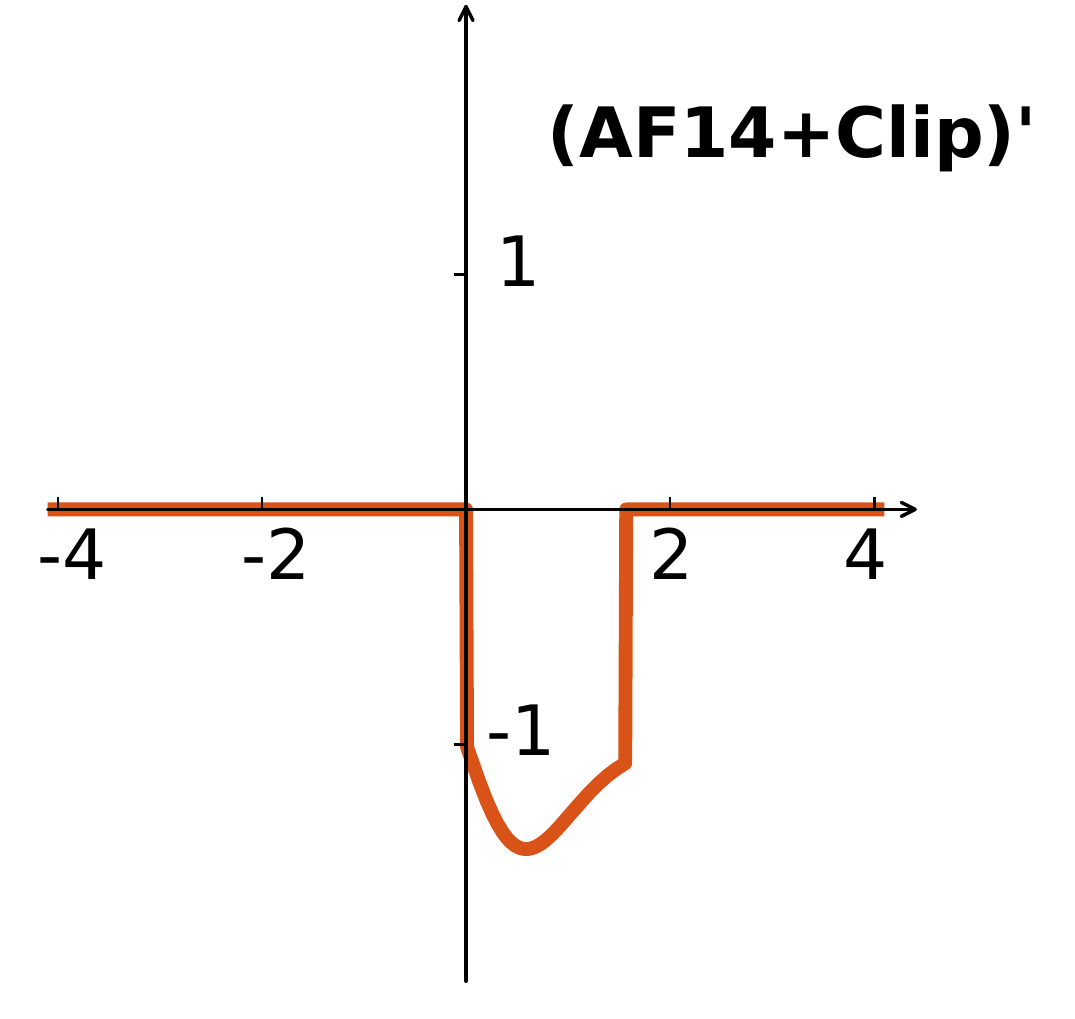} 
\label{fig:F7}
\endminipage\hfill
\caption{The gradient of Discovered promising Activation Functions + Clip}
\label{fig: grad AFClip}
\end{figure*}

\begin{itemize}
    \item AF1: $ sin(x)-cos(x) $,
    \item AF2: $ sin(x)+cos(x) $,
    \item AF4: $ \beta*cos(x) + (1-\beta)*x $,
    \item AF6: $ \beta*erf(x) +  (1-\beta)*max(x,0) $,
    \item AF12: $ \beta* cos(x + \alpha) + (1-\beta)*x $,
    \item AF13: $ cos(atan(x)) + x $,
    \item AF14: $ cos(erf(x)) -x $,
\end{itemize}
The $\alpha$ and $\beta$ here are channel-wise parameters learned during the training. Fig.~\ref{fig:AFun} illustrates these complementary activation functions and Fig.~\ref{fig: grad of AF} visualizes their gradients. Additionally, Fig.~\ref{fig:AFClip} shows the complete activation functions (i.e., AF + Clip(-1,1)). Fig.~\ref{fig: grad AFClip} further depicts their gradients. 

Regarding the impact of computation efficiency, we argue these searched activation functions are very lightweight scalar functions. Additionally, existing works \cite{li2019bstc, geng2019lp} have already shown that the $batchnorm$ layer can be integrated with the $Sign$ layer to form a very efficient threshold-comparison layer during BNN inference, which can greatly simply hardware design and eliminate runtime cost. We argue that this valuable feature is still conserved with our searched complementary activation functions. This is obvious with the monotonic functions (AF3-AF15):

\begin{equation*}
\begin{split}
  ASign(x) & = \begin{cases}
    +1 & AF(x)\geq 0 \\
    -1 & AF(x) \textless 0  \\
  \end{cases}       \\
     & =  \begin{cases}
    +1 & x\geq \tau \\
    -1 & x \textless \tau  \\
  \end{cases}       \\
\end{split}
\end{equation*}
\vskip 2mm

For the two periodic functions (AF1 and AF2), the integration of batchnorm, activation, and sign can be achieved by piece-wise threshold operation.

Regarding the extra memory space cost, there are no additional parameters in AF1, AF2, AF13 and AF14. Other AFs may contain one or two parameters, but only in  channel-wise, which is almost negligible comparing to the size of the weights.

Please note that although in this work we adopt GA-based searching to automatically find new complementary functions for BNNs, this approach can be applied to other network models such as quantized neural networks, transformers, CNNs, or for domain-specific architectural search.

\section{Related Work}
\label{s:Related Work}
\subsection{Binary Neural Networks}
\noindent
BNNs demonstrate substantially advanced memory and computational efficiency at the cost of certain  accuracy degradation. Since the proposal of BNNs \cite{courbariaux2016binarized, hubara2016binarized}, there have been significant efforts committed to closing the accuracy gap between BNNs and the full-precision DNNs. These include adding learnable scaling factors (also known as gain term) \cite{zhou2016dorefa,tang2017train,lin2017towards,darabi2018bnn}, adopting multiple basis \cite{zhou2016dorefa,tang2017train,lin2017towards, ghasemzadeh2018rebnet}, designing BNN-oriented network structure \cite{bethge2021meliusnet, liu2018bi, liu2020reactnet, bethge2019binarydensenet}, and reforming BNN training methodologies\cite{lahoud2019self, liu2020reactnet, darabi2018bnn, alizadeh2018empirical,hou2016loss,helwegen2019latent}. More detailed information can be found in the BNN surveys \cite{simons2019review, qin2020binary}.

\subsection{Automatic Search}
\noindent
Hand-crafted activation functions heavily rely on human expertise and prior network architecture design experience. Due to such limitations, the use of automatic searching algorithms has received more attention in recent years. Various searching strategies have been proposed, including Bayesian optimization \cite{swersky2014raiders, kandasamy2018neural}, evolutionary approaches \cite{real2017large,elsken2018efficient,real2019aging}, reinforcement learning (RL)\cite{baker2016designing,zhong2018practical,zoph2016neural}, and gradient-based methodology \cite{liu2018darts, xie2018snas, cai2018proxylessnas}, etc. 

Particularly for activation function search, Ramachandran et al. \cite{ramachandran2017searching} identified a new activation function called Switch through a combination of exhaustive search and RL, showcasing improved performance than widely used ReLU function. In \cite{liu2020evolving}, researchers co-design the normalization and activation functions as a single computation graph. After evolving the structure from low-level primitives, a novel structure called EvoNorm is identified that can work better on image classification, instance segmentation, image synthesis, etc.

\section{Conclusion}
\label{s:Conclusion}
\noindent
In this paper we focus on closing the accuracy gap between BNNs and their DNN counterparts by augmenting a complementary activation function prior to the sign-based binarization. In order to efficiently search in the large design space, we propose to rely on genetic algorithm for identifying novel AFs presenting improved and generalizable performance. Evaluations on different network models and datasets show that the auto-searched AFs can harvest considerable accuracy improvement (up to 2.54\% on ImageNet).

\renewcommand\refname{\large\textbf{References}}

\bibliographystyle{unsrt}

\bibliography{References}

\begin{thebibliography}{10}

\bibitem{courbariaux2016binarized}
Matthieu Courbariaux, Itay Hubara, Daniel Soudry, Ran El-Yaniv, and Yoshua
  Bengio.
\newblock Binarized neural networks: Training deep neural networks with weights
  and activations constrained to+ 1 or-1.
\newblock {\em arXiv preprint arXiv:1602.02830}, 2016.

\bibitem{hubara2016binarized}
Itay Hubara, Matthieu Courbariaux, Daniel Soudry, Ran El-Yaniv, and Yoshua
  Bengio.
\newblock Binarized neural networks.
\newblock In {\em Proceedings of the 30th international conference on neural
  information processing systems}, pages 4114--4122. Citeseer, 2016.

\bibitem{fasfous2021binarycop}
Nael Fasfous, Manoj-Rohit Vemparala, Alexander Frickenstein, Lukas
  Frickenstein, and Walter Stechele.
\newblock Binarycop: Binary neural network-based covid-19 face-mask wear and
  positioning predictor on edge devices.
\newblock {\em arXiv preprint arXiv:2102.03456}, 2021.

\bibitem{chen2020gpu}
Gang Chen, Haitao Meng, Yucheng Liang, and Kai Huang.
\newblock Gpu-accelerated real-time stereo estimation with binary neural
  network.
\newblock {\em IEEE Transactions on Parallel and Distributed Systems},
  31(12):2896--2907, 2020.

\bibitem{huang2021fpga}
Chun-Hsian Huang.
\newblock An fpga-based hardware/software design using binarized neural
  networks for agricultural applications: A case study.
\newblock {\em IEEE Access}, 9:26523--26531, 2021.

\bibitem{ma2019efficient}
Yinglan Ma, Hongyu Xiong, Zhe Hu, and Lizhuang Ma.
\newblock Efficient super resolution using binarized neural network.
\newblock In {\em Proceedings of the IEEE/CVF Conference on Computer Vision and
  Pattern Recognition Workshops}, pages 0--0, 2019.

\bibitem{ma2018binary}
Chao Ma, Yulan Guo, Yinjie Lei, and Wei An.
\newblock Binary volumetric convolutional neural networks for 3-d object
  recognition.
\newblock {\em IEEE Transactions on Instrumentation and Measurement},
  68(1):38--48, 2018.

\bibitem{li2019bstc}
Ang Li, Tong Geng, Tianqi Wang, Martin Herbordt, Shuaiwen~Leon Song, and Kevin
  Barker.
\newblock Bstc: A novel binarized-soft-tensor-core design for accelerating
  bit-based approximated neural nets.
\newblock In {\em Proceedings of the International Conference for High
  Performance Computing, Networking, Storage and Analysis}, pages 1--30, 2019.

\bibitem{galloway2017attacking}
Angus Galloway, Graham~W Taylor, and Medhat Moussa.
\newblock Attacking binarized neural networks.
\newblock {\em arXiv preprint arXiv:1711.00449}, 2017.

\bibitem{hu2018bitflow}
Yuwei Hu, Jidong Zhai, Dinghua Li, Yifan Gong, Yuhao Zhu, Wei Liu, Lei Su, and
  Jiangming Jin.
\newblock Bitflow: Exploiting vector parallelism for binary neural networks on
  cpu.
\newblock In {\em 2018 IEEE International Parallel and Distributed Processing
  Symposium (IPDPS)}, pages 244--253. IEEE, 2018.

\bibitem{li2020accelerating}
Ang Li and Simon Su.
\newblock Accelerating binarized neural networks via bit-tensor-cores in turing
  gpus.
\newblock {\em IEEE Transactions on Parallel and Distributed Systems},
  32(7):1878--1891, 2020.

\bibitem{geng2019o3bnn}
Tong Geng, Tianqi Wang, Chunshu Wu, Chen Yang, Wei Wu, Ang Li, and Martin~C
  Herbordt.
\newblock O3bnn: An out-of-order architecture for high-performance binarized
  neural network inference with fine-grained pruning.
\newblock In {\em Proceedings of the ACM International Conference on
  Supercomputing}, pages 461--472, 2019.

\bibitem{geng2019lp}
Tong Geng, Tianqi Wang, Chunshu Wu, Chen Yang, Shuaiwen~Leon Song, Ang Li, and
  Martin Herbordt.
\newblock Lp-bnn: Ultra-low-latency bnn inference with layer parallelism.
\newblock In {\em 2019 IEEE 30th International Conference on
  Application-specific Systems, Architectures and Processors (ASAP)}, volume
  2160, pages 9--16. IEEE, 2019.

\bibitem{geng2020o3bnn}
Tong Geng, Ang Li, Tianqi Wang, Chunshu Wu, Yanfei Li, Runbin Shi, Wei Wu, and
  Martin Herbordt.
\newblock O3bnn-r: An out-of-order architecture for high-performance and
  regularized bnn inference.
\newblock {\em IEEE Transactions on parallel and distributed systems},
  32(1):199--213, 2020.

\bibitem{anderson2017high}
Alexander~G Anderson and Cory~P Berg.
\newblock The high-dimensional geometry of binary neural networks.
\newblock {\em arXiv preprint arXiv:1705.07199}, 2017.

\bibitem{bethge2021meliusnet}
Joseph Bethge, Christian Bartz, Haojin Yang, Ying Chen, and Christoph Meinel.
\newblock Meliusnet: An improved network architecture for binary neural
  networks.
\newblock In {\em Proceedings of the IEEE/CVF Winter Conference on Applications
  of Computer Vision}, pages 1439--1448, 2021.

\bibitem{bengio2013estimating}
Yoshua Bengio, Nicholas L{\'e}onard, and Aaron Courville.
\newblock Estimating or propagating gradients through stochastic neurons for
  conditional computation.
\newblock {\em arXiv preprint arXiv:1308.3432}, 2013.

\bibitem{liu2018bi}
Zechun Liu, Baoyuan Wu, Wenhan Luo, Xin Yang, Wei Liu, and Kwang-Ting Cheng.
\newblock Bi-real net: Enhancing the performance of 1-bit cnns with improved
  representational capability and advanced training algorithm.
\newblock In {\em Proceedings of the European conference on computer vision
  (ECCV)}, pages 722--737, 2018.

\bibitem{darabi2018bnn+}
Sajad Darabi, Mouloud Belbahri, Matthieu Courbariaux, and Vahid~Partovi Nia.
\newblock Bnn+: Improved binary network training.
\newblock 2018.

\bibitem{liu2019circulant}
Chunlei Liu, Wenrui Ding, Xin Xia, Baochang Zhang, Jiaxin Gu, Jianzhuang Liu,
  Rongrong Ji, and David Doermann.
\newblock Circulant binary convolutional networks: Enhancing the performance of
  1-bit dcnns with circulant back propagation.
\newblock In {\em Proceedings of the IEEE/CVF Conference on Computer Vision and
  Pattern Recognition}, pages 2691--2699, 2019.

\bibitem{qin2020forward}
Haotong Qin, Ruihao Gong, Xianglong Liu, Mingzhu Shen, Ziran Wei, Fengwei Yu,
  and Jingkuan Song.
\newblock Forward and backward information retention for accurate binary neural
  networks.
\newblock In {\em Proceedings of the IEEE/CVF Conference on Computer Vision and
  Pattern Recognition}, pages 2250--2259, 2020.

\bibitem{lahoud2019self}
Fayez Lahoud, Radhakrishna Achanta, Pablo M{\'a}rquez-Neila, and Sabine
  S{\"u}sstrunk.
\newblock Self-binarizing networks.
\newblock {\em arXiv preprint arXiv:1902.00730}, 2019.

\bibitem{liu2020reactnet}
Zechun Liu, Zhiqiang Shen, Marios Savvides, and Kwang-Ting Cheng.
\newblock Reactnet: Towards precise binary neural network with generalized
  activation functions.
\newblock In {\em European Conference on Computer Vision}, pages 143--159.
  Springer, 2020.

\bibitem{ramachandran2017searching}
Prajit Ramachandran, Barret Zoph, and Quoc~V Le.
\newblock Searching for activation functions.
\newblock {\em arXiv preprint arXiv:1710.05941}, 2017.

\bibitem{liu2020evolving}
Hanxiao Liu, Andrew Brock, Karen Simonyan, and Quoc~V Le.
\newblock Evolving normalization-activation layers.
\newblock {\em arXiv preprint arXiv:2004.02967}, 2020.

\bibitem{liu2018darts}
Hanxiao Liu, Karen Simonyan, and Yiming Yang.
\newblock Darts: Differentiable architecture search.
\newblock {\em arXiv preprint arXiv:1806.09055}, 2018.

\bibitem{han2015deep}
Song Han, Huizi Mao, and William~J Dally.
\newblock Deep compression: Compressing deep neural networks with pruning,
  trained quantization and huffman coding.
\newblock {\em arXiv preprint arXiv:1510.00149}, 2015.

\bibitem{blalock2020state}
Davis Blalock, Jose Javier~Gonzalez Ortiz, Jonathan Frankle, and John Guttag.
\newblock What is the state of neural network pruning?
\newblock {\em arXiv preprint arXiv:2003.03033}, 2020.

\bibitem{ioffe2015batch}
Sergey Ioffe and Christian Szegedy.
\newblock Batch normalization: Accelerating deep network training by reducing
  internal covariate shift.
\newblock In {\em International conference on machine learning}, pages
  448--456. PMLR, 2015.

\bibitem{goldberg1988genetic}
David~E Goldberg and John~Henry Holland.
\newblock Genetic algorithms and machine learning.
\newblock 1988.

\bibitem{katoch2020review}
Sourabh Katoch, Sumit~Singh Chauhan, and Vijay Kumar.
\newblock A review on genetic algorithm: past, present, and future.
\newblock {\em Multimedia Tools and Applications}, pages 1--36, 2020.

\bibitem{kingma2014adam}
Diederik~P Kingma and Jimmy Ba.
\newblock Adam: A method for stochastic optimization.
\newblock {\em arXiv preprint arXiv:1412.6980}, 2014.

\bibitem{tang2017train}
Wei Tang, Gang Hua, and Liang Wang.
\newblock How to train a compact binary neural network with high accuracy?
\newblock In {\em Thirty-First AAAI conference on artificial intelligence},
  2017.

\bibitem{zhou2016dorefa}
Shuchang Zhou, Yuxin Wu, Zekun Ni, Xinyu Zhou, He~Wen, and Yuheng Zou.
\newblock Dorefa-net: Training low bitwidth convolutional neural networks with
  low bitwidth gradients.
\newblock {\em arXiv preprint arXiv:1606.06160}, 2016.

\bibitem{lin2017towards}
Xiaofan Lin, Cong Zhao, and Wei Pan.
\newblock Towards accurate binary convolutional neural network.
\newblock In {\em Advances in Neural Information Processing Systems}, pages
  345--353, 2017.

\bibitem{darabi2018bnn}
Sajad Darabi, Mouloud Belbahri, Matthieu Courbariaux, and Vahid~Partovi Nia.
\newblock Bnn+: Improved binary network training.
\newblock {\em arXiv preprint arXiv:1812.11800}, 2018.

\bibitem{ghasemzadeh2018rebnet}
Mohammad Ghasemzadeh, Mohammad Samragh, and Farinaz Koushanfar.
\newblock Rebnet: Residual binarized neural network.
\newblock In {\em 2018 IEEE 26th Annual International Symposium on
  Field-Programmable Custom Computing Machines (FCCM)}, pages 57--64. IEEE,
  2018.

\bibitem{bethge2019binarydensenet}
Joseph Bethge, Haojin Yang, Marvin Bornstein, and Christoph Meinel.
\newblock Binarydensenet: developing an architecture for binary neural
  networks.
\newblock In {\em Proceedings of the IEEE/CVF International Conference on
  Computer Vision Workshops}, pages 0--0, 2019.

\bibitem{alizadeh2018empirical}
Milad Alizadeh, Javier Fern{\'a}ndez-Marqu{\'e}s, Nicholas~D Lane, and Yarin
  Gal.
\newblock An empirical study of binary neural networks' optimisation.
\newblock 2018.

\bibitem{hou2016loss}
Lu~Hou, Quanming Yao, and James~T Kwok.
\newblock Loss-aware binarization of deep networks.
\newblock {\em arXiv preprint arXiv:1611.01600}, 2016.

\bibitem{helwegen2019latent}
Koen Helwegen, James Widdicombe, Lukas Geiger, Zechun Liu, Kwang-Ting Cheng,
  and Roeland Nusselder.
\newblock Latent weights do not exist: Rethinking binarized neural network
  optimization.
\newblock In {\em Advances in neural information processing systems}, pages
  7531--7542, 2019.

\bibitem{simons2019review}
Taylor Simons and Dah-Jye Lee.
\newblock A review of binarized neural networks.
\newblock {\em Electronics}, 8(6):661, 2019.

\bibitem{qin2020binary}
Haotong Qin, Ruihao Gong, Xianglong Liu, Xiao Bai, Jingkuan Song, and Nicu
  Sebe.
\newblock Binary neural networks: A survey.
\newblock {\em Pattern Recognition}, 105:107281, 2020.

\bibitem{swersky2014raiders}
Kevin Swersky, David Duvenaud, Jasper Snoek, Frank Hutter, and Michael~A
  Osborne.
\newblock Raiders of the lost architecture: Kernels for bayesian optimization
  in conditional parameter spaces.
\newblock {\em arXiv preprint arXiv:1409.4011}, 2014.

\bibitem{kandasamy2018neural}
Kirthevasan Kandasamy, Willie Neiswanger, Jeff Schneider, Barnabas Poczos, and
  Eric Xing.
\newblock Neural architecture search with bayesian optimisation and optimal
  transport.
\newblock {\em arXiv preprint arXiv:1802.07191}, 2018.

\bibitem{real2017large}
Esteban Real, Sherry Moore, Andrew Selle, Saurabh Saxena, Yutaka~Leon Suematsu,
  Jie Tan, Quoc~V Le, and Alexey Kurakin.
\newblock Large-scale evolution of image classifiers.
\newblock In {\em International Conference on Machine Learning}, pages
  2902--2911. PMLR, 2017.

\bibitem{elsken2018efficient}
Thomas Elsken, Jan~Hendrik Metzen, and Frank Hutter.
\newblock Efficient multi-objective neural architecture search via lamarckian
  evolution.
\newblock {\em arXiv preprint arXiv:1804.09081}, 2018.

\bibitem{real2019aging}
Esteban Real, Alok Aggarwal, Yanping Huang, and Quoc~V Le.
\newblock Aging evolution for image classifier architecture search.
\newblock In {\em AAAI Conference on Artificial Intelligence}, volume~2, 2019.

\bibitem{baker2016designing}
Bowen Baker, Otkrist Gupta, Nikhil Naik, and Ramesh Raskar.
\newblock Designing neural network architectures using reinforcement learning.
\newblock {\em arXiv preprint arXiv:1611.02167}, 2016.

\bibitem{zhong2018practical}
Zhao Zhong, Junjie Yan, Wei Wu, Jing Shao, and Cheng-Lin Liu.
\newblock Practical block-wise neural network architecture generation.
\newblock In {\em Proceedings of the IEEE conference on computer vision and
  pattern recognition}, pages 2423--2432, 2018.

\bibitem{zoph2016neural}
Barret Zoph and Quoc~V Le.
\newblock Neural architecture search with reinforcement learning.
\newblock {\em arXiv preprint arXiv:1611.01578}, 2016.

\bibitem{xie2018snas}
Sirui Xie, Hehui Zheng, Chunxiao Liu, and Liang Lin.
\newblock Snas: stochastic neural architecture search.
\newblock {\em arXiv preprint arXiv:1812.09926}, 2018.

\bibitem{cai2018proxylessnas}
Han Cai, Ligeng Zhu, and Song Han.
\newblock Proxylessnas: Direct neural architecture search on target task and
  hardware.
\newblock {\em arXiv preprint arXiv:1812.00332}, 2018.

\end{thebibliography}

\end{document}